\pdfoutput=1
\PassOptionsToPackage{table,dvipsnames}{xcolor}
\documentclass[11pt]{article}
\usepackage[final]{acl}
\usepackage{times}
\usepackage{latexsym}
\usepackage[T1]{fontenc}
\usepackage[utf8]{inputenc}
\usepackage{microtype}
\usepackage{inconsolata}
\usepackage{tabularx}
\usepackage{tcolorbox}
\usepackage{authblk}
\usepackage[section]{placeins}
\usepackage{amssymb}
\usepackage[table, dvipsnames]{xcolor}
\usepackage{booktabs}
\usepackage{arydshln}
\usepackage{graphicx}
\usepackage{multirow}
\usepackage{xspace}
\usepackage{etoolbox}
\usepackage{pifont}
\usepackage{csvsimple}
\usepackage{makecell}
\usepackage[colorinlistoftodos]{todonotes}
\usepackage{tikz}
\usepackage{amsmath}
\usepackage{amssymb}
\usepackage{scalerel}
\usepackage{stfloats}               
\newif\ifdraftmode
\draftmodetrue  

\newcommand{\papertitle}{\textsc{FLaME}\xspace}

\newcommand{\nummodels}{23\xspace}
\newcommand{\numtasks}{20\xspace}

\newcommand{\citeFLUE}{\citet{Shah2023-bh}\xspace}
\newcommand{\citepFLUE}{\citep{Shah2023-bh}\xspace}

\newcommand{\citepFLARE}{\citep{Xie2023-xf}\xspace}
\newcommand{\citeFinBen}{\citet{Xie2024-pn}\xspace}
\newcommand{\citepFinBen}{\citep{Xie2024-pn}\xspace}

\newcommand{\citepHELM}{\citep{Liang2022-ew}\xspace}

\newcommand{\citeTODO}{\textcolor{red}{[CITE?]}%
\gappto{\citetodolist}{\protect\noindent Line \the\inputlineno: TODO Citation Needed \thesection.\\}}
\newcommand{\citetodolist}{}

\newcommand{\cmark}{\ding{51}}
\newcommand{\xmark}{\ding{55}}

\newcommand\eg{e.g.\xspace}
\newcommand\ie{i.e.\xspace}

\newcommand\emdash{\xspace---\xspace}
\newcommand{\dash}{\emdash} 

\newcommand\refsec[1]{\hyperlink{#1}{§\ref{sec:#1}:~\textsc{#1}}}

\newcommand\reffig[1]{Figure~\ref{fig:#1}}

\newcommand\reftab[1]{Table~\ref{tab:#1}}
\newcommand\refapp[1]{Appendix~\ref{app:#1}}

\newcommand{\para}[1]{\noindent\textbf{#1}\quad}

\newcommand{\huggingface}{\includegraphics[height=1.7ex]{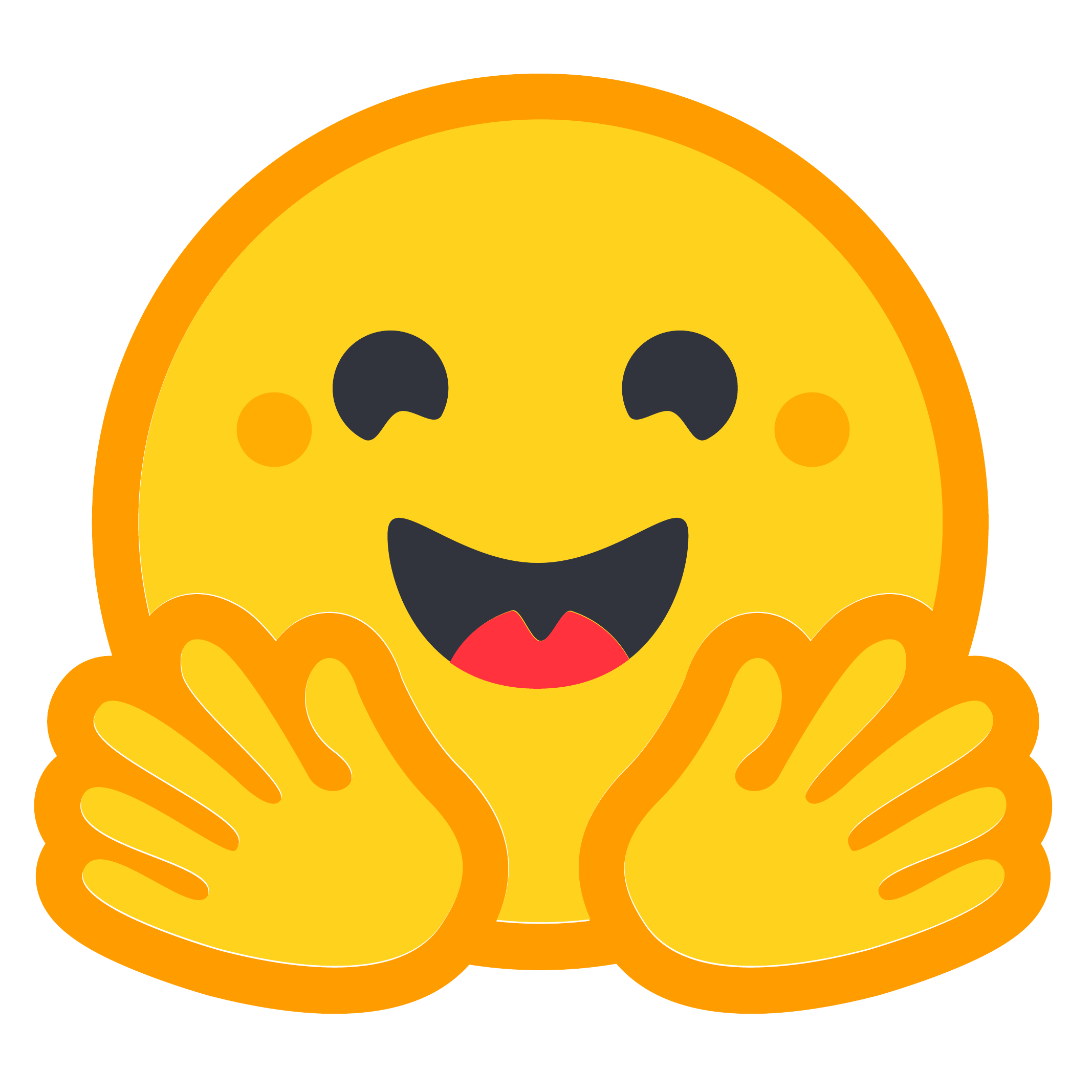}}

\newcommand{\github}{\includegraphics[height=1.7ex]{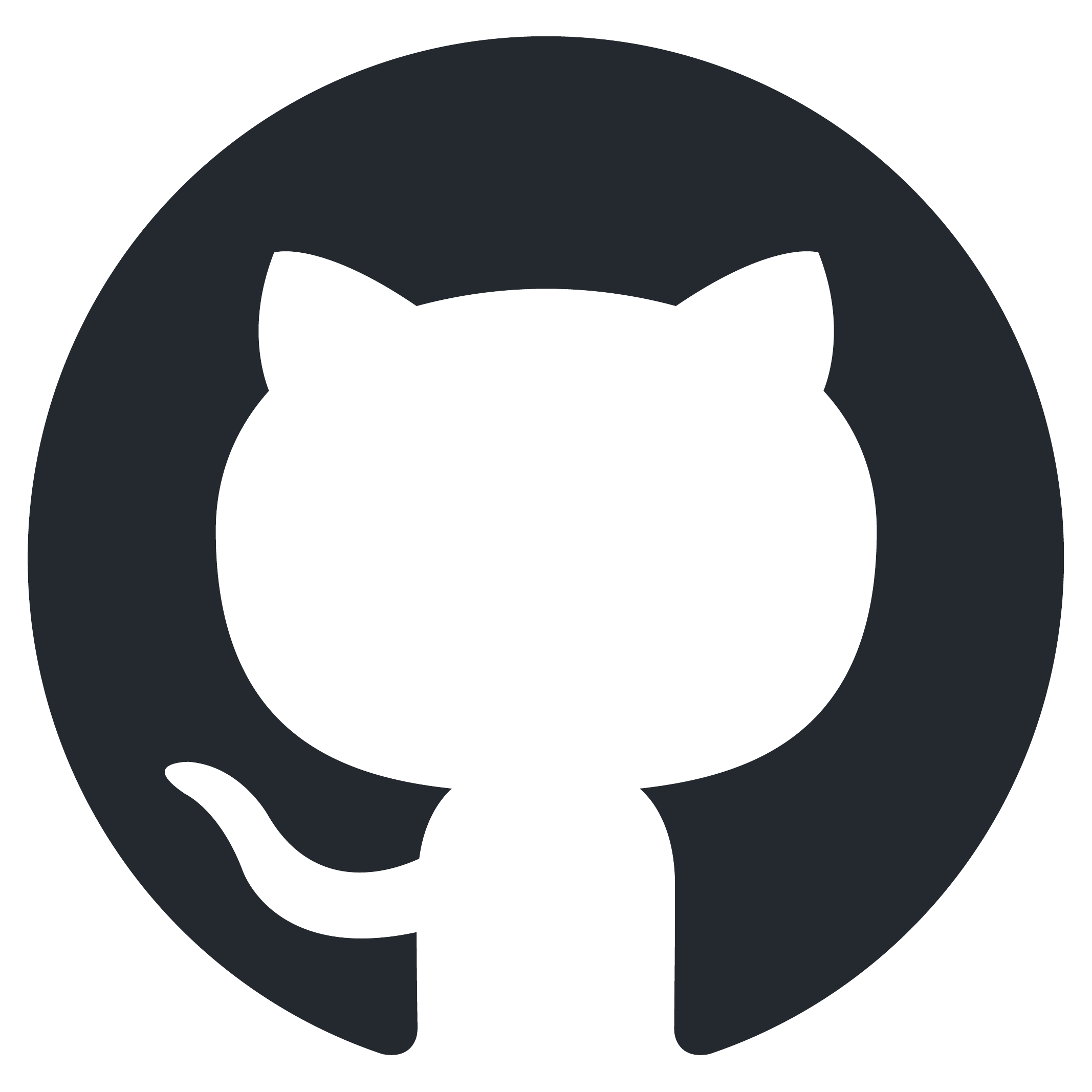}}

\title{Finance Language Model Evaluation (\papertitle)}
\author{%
  Glenn Matlin\textsuperscript{$\dagger\diamond$}\quad
  Mika Okamoto\textsuperscript{$\diamond$}\quad
  Huzaifa Pardawala\textsuperscript{$\diamond$}\quad
  Yang Yang\quad
  Sudheer Chava\\
  Georgia Institute of Technology\\
  \github \href{https://github.com/gtfintechlab/FLaME}{\texttt{github.com/gtfintechlab/FLaME}}\\
  \huggingface \href{https://huggingface.co/collections/gtfintechlab/flame-6831e0e48d4ce092550dfac7}{\texttt{huggingface.co/gtfintechlab/FLaME}}
}
\begin{document}
\maketitle
\begin{abstract}
Language Models (LMs) have demonstrated impressive capabilities with core Natural Language Processing (NLP) tasks. The effectiveness of LMs for highly specialized knowledge-intensive tasks in finance remains difficult to assess due to major gaps in the methodologies of existing evaluation frameworks. These gaps have caused an erroneous belief in a far lower bound of LMs' performance on common Finance NLP (FinNLP) tasks. To accurately assess LM capabilities and demonstrate their potential for FinNLP tasks, we present the first holistic benchmarking suite for \textbf{\textit{Financial Language Model Evaluation}} (\papertitle). Our work includes the first comprehensive empirical study comparing standard LMs with 'reasoning-reinforced' LMs with \nummodels foundation LMs over \numtasks core financial NLP tasks. We open-source our framework software along with all data and results.
\end{abstract}

\begin{figure*}[!b]
    \centering
    \includegraphics[width=1\textwidth,
    height=0.25\textheight,
    keepaspectratio
    ]{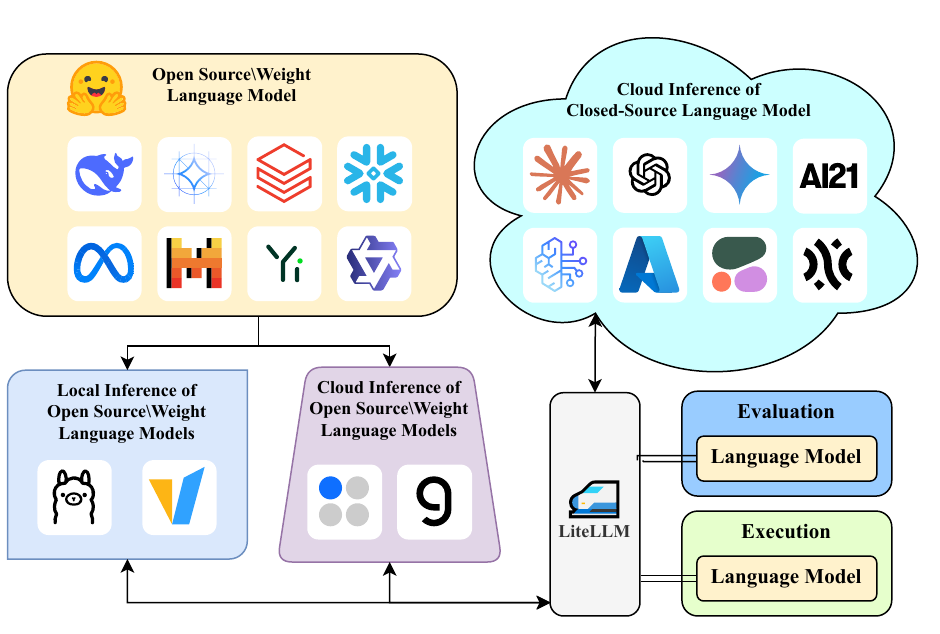}
    \caption{\textbf{Technical Overview}: \papertitle uses a unified inference hub, providing a single, model-agnostic API across three deployment modes: (i) proprietary cloud APIs (e.g., Claude 4, Gemini 2.5 Pro), (ii) cloud-hosted open-weight models (e.g., OLMo 2, Qwen 2.5) served by either cloud providers (TogetherAI, HuggingFace), and (iii) fully local inference backends (e.g., vLLM, Ollama). This modular software design abstracts deployment complexity, enabling rapid experimentation and comprehensive benchmarking across a diverse range of language models, significantly simplifying evaluation workflows to promote replication and transparency in FinNLP research.}
    \label{fig:overview_tech}
\end{figure*}
\section{Introduction}\label{sec:introduction}
\begin{table*}[!t]
\centering
\renewcommand{\arraystretch}{1.1} 
\resizebox{0.98\textwidth}{!}{%
\begin{tabular}{l l c c c c c c c c c c}
\toprule
\textbf{Benchmark Suite} & \textbf{Languages} & \makecell{\textbf{Core} \\ \textbf{NLP} \\ \textbf{Datasets}}
 & \makecell{\textbf{Model} \\ \textbf{Families} \\ \textbf{Evaluated}} & \makecell{\textbf{Foundation} \\ \textbf{LMs} \\ \textbf{Evaluated}} & \makecell{\textbf{Reasoning-} \\ \textbf{Reinforced} \\ \textbf{LMs Evaluated}} & \makecell{\textbf{Standardized} \\ \textbf{Evaluations}} & \makecell{\textbf{Recognition of} \\ \textbf{Incompleteness}} & \makecell{\textbf{Multi-} \\ \textbf{Metric} \\ \textbf{Evaluation}} & \makecell{\textbf{Data} \\ \textbf{Quality}\\ \textbf{Assurance}} & \makecell{\textbf{Taxonomy} \\ \textbf{of Scenarios}} & \makecell{\textbf{Public} \\ \textbf{Benchmark} \\ \textbf{Leaderboard}} \\
\midrule
FLUE \citeFLUE & ENG  & 6  & 0  & 0 & 0  & \xmark  & \xmark  & \xmark  & \xmark  & \xmark  & \xmark \\
FLARE \cite{Xie2023-xf} & ENG  & 9 & 1  & 1 & 0  & \xmark  & \xmark  & \xmark  & \xmark  & \xmark  & \xmark \\
CFBenchmark \cite{Lei2023-ox} & CHI  & 3 & 8  & 11 & 0  & \xmark  & \xmark  & \xmark  & \xmark  & \xmark  & \xmark \\
BizBench \citep{Koncel-Kedziorski2023-fx} & ENG  & 8  & 7  & 16 & 0 & \cmark  & \cmark  & \xmark  & \cmark  & \xmark  & \xmark \\
FinBen \citep{Xie2024-pn} & ENG  & 22 & 7 & 9 & 0 & \cmark  & \xmark  & \xmark  & \xmark  & \xmark  & \xmark \\
Golden Touchstone \cite{Wu2024-df} & \makecell{CHI + ENG} & 20 & 4  & 4  & 0  & \cmark  & \xmark  & \cmark  & \xmark  & \xmark  & \xmark \\
\papertitle & ENG & \numtasks & 12  & \nummodels & 3  & \cmark  & \cmark  & \cmark  & \cmark  & \cmark  & \cmark \\

\bottomrule
\end{tabular}
}
\caption{\textbf{Comparison of Benchmark Suites for Financial NLP.} We compare this work against state-of-the-art benchmark suites for financial NLP tasks across datasets, scenario coverage, number of \textit{foundation} model families and individual models evaluated. \papertitle is the only benchmark that qualifies as \textit{holistic}.}
\label{tab:us-vs-them}
\end{table*}

\begin{figure*}
    \centering
    \includegraphics[width=1\linewidth]{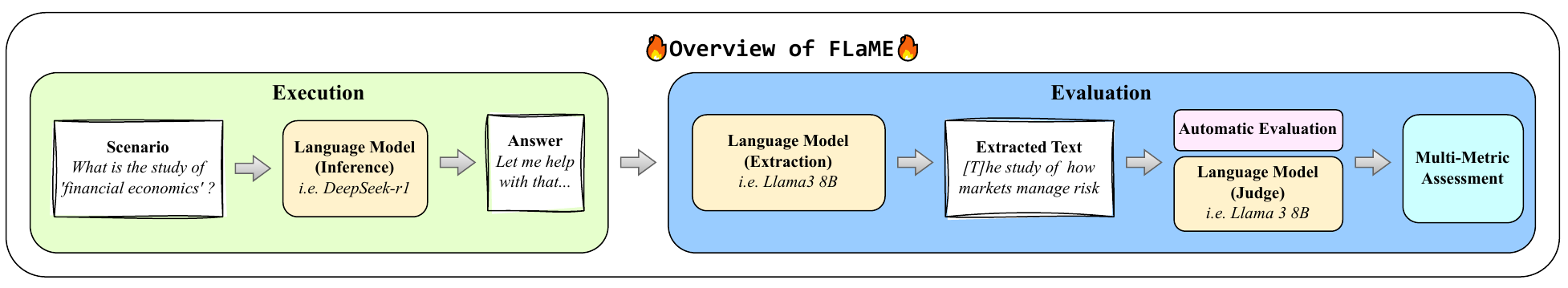}
    \caption{\textbf{Functional Overview}: In the Execution phase, a language model (e.g., DeepSeek-r1) generates responses to financial queries. During the Evaluation phase, text spans are extracted from the generated text by an LM (e.g., Llama3 3B), followed by either directly verifying the answer or using automated evaluation performed by a judge LM (e.g., Llama3 8B). \papertitle's main contribution is providing a comprehensive software package and standardized methodology for reproducible multi-metric assessment of LM performance on core FinNLP tasks.}
    \label{fig:overview}
\end{figure*}
\renewcommand{\thefootnote}{}%
\footnotetext{%
$\dagger$~Corresponding author: \href{mailto:glenn@gatech.edu}{glenn@gatech.edu}\\
$\diamond$~These authors contributed equally to this work.
}%
\setcounter{footnote}{0}\renewcommand{\thefootnote}{\arabic{footnote}5}
Benchmarks and datasets are the foundation for Artificial Intelligence (AI) research. How the research community collectively defines \textit{'success'} directly shapes researchers' priorities and goals \cite{Raji2021-kj}. Benchmarks enable the wider research community to understand and track progress in AI development \cite{Birhane2022-az}. Recent developments enabling the general commercial availability of foundation Language Models (LMs) \cite{Bommasani2021-ro, Zhao2023-mn} (\eg, ChatGPT \cite{Brown2020-zf}, Claude \cite{AnthropicUnknown-vl}, Gemini \cite{Gemini-Team2023-gv}, etc. have fueled widespread interest in tracking their progress \cite{Chang2023-wf,Nie2024-mh}. The widespread availability of LMs has spurred research into AI capabilities for highly specialized and knowledge-intensive domains, such as medicine, law, and finance \cite{Guha2024-ox,Chen2024-zb,Kaddour2023-sg}.

Prior research has raised serious concerns about the ability of LMs to generalize their reasoning or adapt to specialized domains \citep{Bender2021-wf, Kocon2023-py}, particularly finance \citealp{Kang2023-hn, Zhao2024-eh, Dong2024-xd, Chen2024-zb}. Despite this explosion of interest and skepticism, there has not yet been a sufficiently rigorous and \textbf{\textit{holistic evaluation}} of the performance of foundation LMs for core NLP tasks in finance. Existing state-of-the-art efforts lack sufficient standardization and rigor to identify the true performance bounds of foundation LMs. Poor understanding of these errors leads to real-world failures in financial computing systems. 
The risk of failures in AI-enabled financial systems should be a primary concern for both academia and industry. Without a deep understanding of common failures in LM-enabled finance NLP tasks (e.g., generating incorrect financial data), these systems may mislead users, leading to substantial harm. Misinformation stemming from analytical failures, flawed reasoning, or outright hallucinations remains a persistent challenge and may be difficult, if not impossible, to fully eliminate \citep{Ye2023-qn, Li2023-oo, Xu2024-cm, Ji2022-fk}.

Over the past few years, multiple benchmark evaluation suites have emerged to assess model performance on finance-oriented NLP tasks. However, these efforts typically:
\begin{enumerate}
    \item Are collections of benchmarks without establishing an in-depth taxonomy,
    \item Lack standardized criteria for data selection or evaluation,
    \item Omit a systematic recognition of incompleteness of their current methods, and
    \item Narrow evaluation scope with only fine-tuned or closed-source LMs.
\end{enumerate}

\textbf{\textit{Holistic evaluations}} are critical for AI in finance. System failures, caused by an insufficient understanding of LM weaknesses on core financial NLP tasks, will cause serious public harm and entail significant economic and legal consequences for businesses an financial institutions. We adopt the widely accepted meaning of \textit{holistic evaluation} from \citet{Liang2022-ew}, which requires: \textbf{(1) standardization}, \textbf{(2) recognition of incompleteness}, and \textbf{(3) multi-metric evaluation}.  \textbf{\textit{Holistic benchmark suites}} help prevent these errors by identifying gaps in data coverage in their dataset taxonomy, encouraging comprehensive study of model behavior, and providing a reliable and repeatable method for comparison.
However, no benchmark suites for evaluating core NLP finance tasks on LMs meet the definition of 'holistic.' In \reftab{us-vs-them}, we assess other existing benchmarks and highlight how they fail to meet the criteria for a holistic evaluation.
To solve this critical gap for our community, we propose \papertitle, which provides the following novel contributions:
\begin{enumerate}
    \item \textbf{Standardized Evaluation Framework}: We release an \textbf{open-source software} for creating standardized pipelines for LM evaluations for core financial NLP tasks. Our configurable pipeline (see \reffig{overview}) handles the complete evaluation process.
    \item \textbf{Large-Scale Model Assessment}: We conduct \textbf{extensive evaluations} of \nummodels open-weight and proprietary LMs, exposing strengths and weaknesses across \numtasks financial benchmarks (see \reffig{overview_tech}). We provide a \textbf{meta-analysis} of the results, including a \textit{study on the performance/cost trade-off space}. Our in-depth \textbf{error analysis} offers more insight into recurring model failures.
    \item \textbf{Living Benchmark}: We provide a \textbf{public leaderboard} to encourage continuous updates. Researchers and practitioners can contribute new datasets or model results, extending \papertitle beyond our initial contributions. By design, this effort \textit{\textbf{explicitly}} welcomes peer review and invites ongoing collaboration.
    \item \textbf{Taxonomy and Dataset Selection}: We present a holistic taxonomy for financial NLP tasks, detailing the financial domain scenario and categorizing benchmarking tasks. We also establish \textbf{clear inclusion criteria} (domain relevance, licensing, label quality).
\end{enumerate}

\section{Related Work}\label{sec:relatedwork}
\subsection{Foundation Language Models}
Recent LM progress (as discussed in Section 1) has driven state-of-the-art performance across many core NLP tasks, including in finance. LMs exhibit strong performance on both general-domain benchmarks and increasingly complex tasks (\eg, multi-hop reasoning, tool use, multi-modal tasks)
The term "Large Language Model" has increased rapidly in use; however, its definition is broad enough to encompass fine-tuned models or systems. We define a \textit{language model} (LM) as probabilistic model for natural language and a \textit{foundation language model} as those trained on broad datasets (typically using large-scale self-supervision) that can be adapted (i.e., fine-tuned) for a wide range of downstream tasks. \cite{Bommasani2021-ro}. Our study aims for a robust and holistic understanding of LM performance rather than use-case-specific adaptations. We prioritize studying foundation LMs, as all fine-tuned models originate from a foundation model. The performance of fine-tuned models heavily depends on the pre-training stage (\ie, self-supervised learning) of the foundation model \cite{Chia2023-sb}.
\subsection{Language Model Evaluation}
\textit{\textbf{Domain-specific}} evaluations for knowledge-intensive fields (\eg, medicine, law, computing) have seen much research interest \cite{Guha2024-ox,Chen2024-zb,Kaddour2023-sg}. However, finance-specific evaluations remain relatively under-studied. As highlighted in \refsec{introduction}, deploying LMs in financial systems without thorough, domain-specific evaluations can lead to incorrect predictions, misinterpretations of regulatory text, flawed market analysis, and other significant financial risks.
A robust body of research has focused on developing benchmarks to measure the evolving capabilities of LMs in broad NLP contexts. Landmark resources such as GLUE \citep{Wang2018-qm}, SuperGLUE \citep{Wang2019-jb}, SQuAD \cite{Rajpurkar2016-si}, HellaSwag \cite{Zellers2019-lp}, and others have helped standardize the evaluation of general natural language understanding for AI. Subsequent benchmarking efforts including MMLU \citep{Hendrycks2020-rz, Wang2024-pk}, Dynabench \cite{Kiela2021-pi, Ma2021-zb}, BigBench \citep{Srivastava2022-tn, Suzgun2022-pp}, the AI2 Reasoning Challenge (ARC) \citep{Clark2018-yg}, and many others have introduced more challenging domains, spanning multi-step reasoning, commonsense tasks, and even agent interactions.
\citepHELM in their Holistic Evaluation of Language Models (HELM) framework, advocate for standardized methods, multi-metric assessments, and explicit recognition of benchmark incompleteness, principles we adopt (see \refsec{introduction}). While these benchmarks have significantly helped with research on general LM capabilities, they do not explicitly address the intricacies of finance-specific applications, such as handling financial definitions, regulatory language, and domain-specific reasoning.
\subsection{Financial Task Benchmarks}
Datasets and benchmarks serve a foundational role in the evaluation of AI systems for finance. Although researchers investigated LMs for finance \cite{Wu2023-ph}, evaluating these models rigorously remains an open challenge. \papertitle builds on these general and domain-specific insights to provide the finance-specific holistic evaluation framework. While several finance-tailored benchmark suites exist (see Table 1), none fully meet the holistic criteria outlined by \citepHELM. In \reftab{us-vs-them}, we assess these existing benchmarks and highlight how these benchmarks fail to meet the criteria for a holistic evaluation. We provide a full discussion and comparison of \papertitle with prior works in \refapp{relatedwork}.
\section{Methodology}\label{sec:methodology}
We present our methodology for holistic financial language model evaluation. \papertitle is the first \textit{\textbf{holistic}} benchmark suite for core NLP tasks in finance. This methodology enables researchers to evaluate the fundamental abilities of foundation models systematically.
\subsection{\papertitle}\label{sec:flame}
We conducted quality checks (license validation, label audits) to ensure each dataset meets the \textbf{inclusion criteria} described in \refapp{datasets}. We give full credit and acknowledgment is given to the authors of these benchmarks. We provide all the pre-processing code for these datasets and direct reader traffic to their original hosting sources. We encourage all readers to refer to our extensive discussion in \refapp{ethicslegal} on the ethics and legal matters regarding appropriate use by others. To promote collaboration and transparent reporting, \papertitle provides a public leaderboard.\\
The evaluation pipeline proceeds in stages:
\begin{enumerate}
    \item \textbf{Configuration:} Users select desired tasks, datasets, and model parameters.
    \item \textbf{Model Interaction:} The system queries each LM \dash via local instantiation or a remote API \dash to collect its outputs. We automatically handle token limits, rate-limiting, and retry logic for cloud services.
    \item \textbf{Post-processing and Extraction:} Generated text undergoes parsing, ensuring any structured output is normalized.
    \item \textbf{Metric Computation:} User-specified metrics are computed. All parameters (prompt, settings) are logged.
\end{enumerate}
This modular design \emph{decomposes} complex tasks, allowing researchers to customize each step \dash e.g., incorporating novel prompt engineering techniques or adding new metrics. By default, \papertitle \emph{checkpoints} each step to guarantee reproducibility and traceability of results. We anticipate the community will extend or refine these modules as FinNLP evolves.

\subsection{Taxonomy}\label{sec:taxonomy}
To address the nonstandard task definitions common in previous benchmark suites (see \refsec{relatedwork}), \papertitle uses a scenario-based taxonomy. Our taxonomy improves on prior works by defining the complex scenario space within FinNLP. Unlike prior works, the \papertitle taxonomy categorizes financial data based on their primary characteristics and attributes. We define our taxonomy based on these characteristics to avoid creating superfluous categories that unnecessarily add complexity by diverging from established NLP terminology. Our taxonomy is intentionally designed to rely on broad categories (with subcategories as appropriate) to maintain a balance between simplicity and granularity. 
By detailing the complex space of different financial scenarios, our taxonomy highlights the current paucity of data and the need for more research work on financial LM benchmarks. The \papertitle website allows users to browse all available datasets and results using our taxonomy. This taxonomic framework enables researchers to analyze the availability and quality of benchmarking datasets in depth. We posit that every possible financial scenario (i.e., what the LM should do) can be represented with a combination of three attributes: tasks, domains, and languages.
\begin{figure}
    \centering
    \includegraphics[width=1\linewidth]{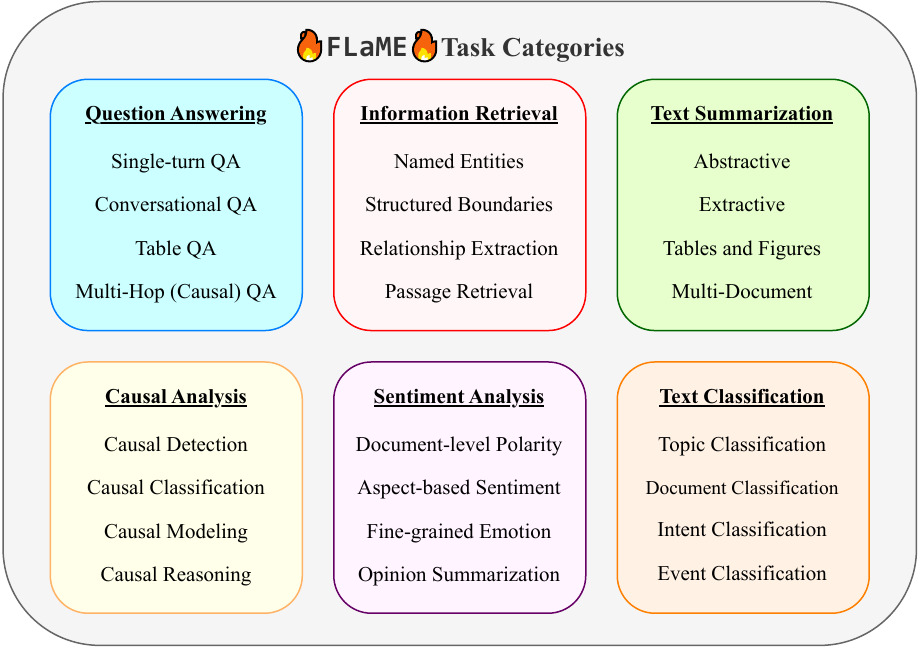}
    \caption{\textbf{Illustrative breakdown for each of the six core NLP task categories.} While our taxonomy groups these tasks broadly, each category can encompass numerous specialized variants depending on data format, user needs, and domain constraints. We provide a limited set of specific examples to illustrate the concepts.}
    \label{fig:methodology_tasks}
\end{figure}
\begin{figure*}
    \centering
    \includegraphics[width=1\linewidth]{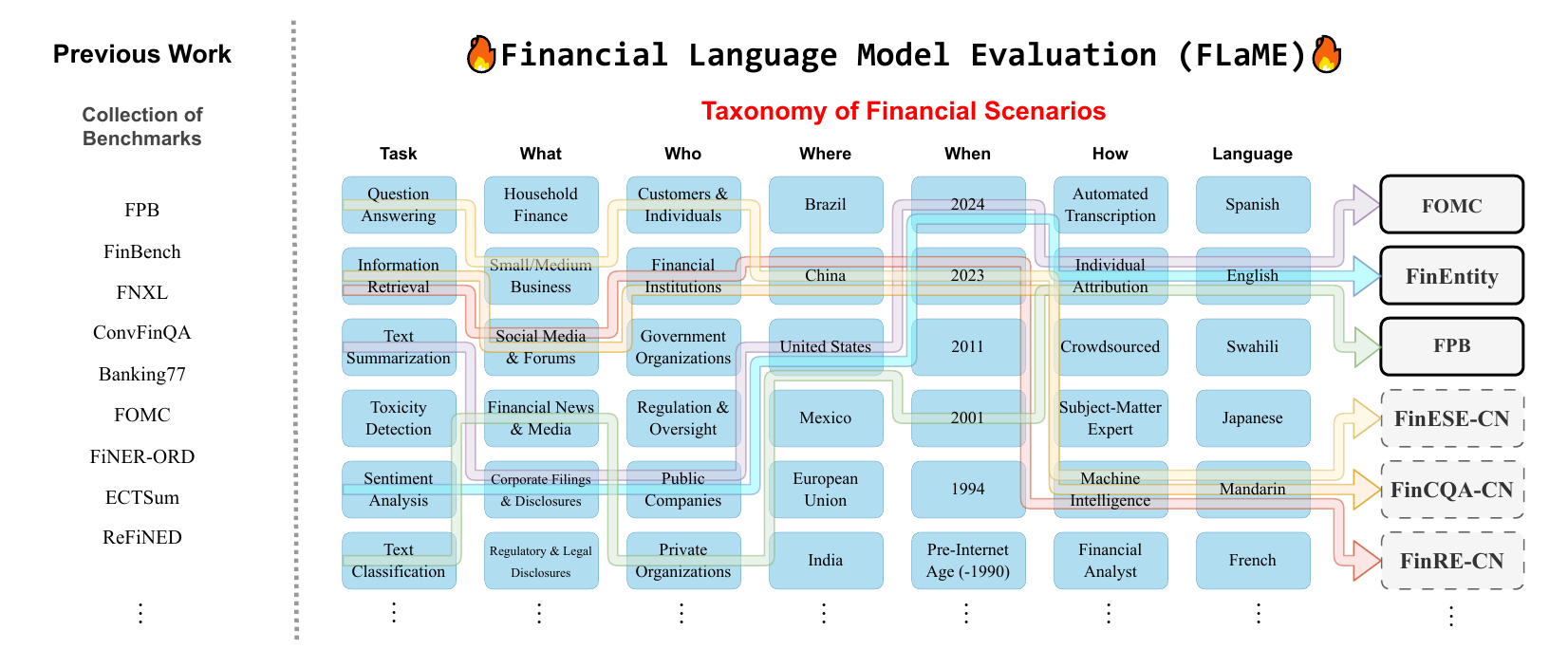}
    \caption{\textbf{Holistic Taxonomy for \papertitle.} Unlike prior FinNLP benchmark suites, which primarily collect individual datasets aligned to specific tasks or metrics, \papertitle adopts a holistic perspective, systematically mapping benchmarks across multiple dimensions such as tasks, scenarios, and contextual attributes. We track not only the currently implemented datasets within the \papertitle repository (solid-line boxes), but we also track datasets yet to be implemented (dotted-lined boxes) and produce desiredata by identifying scenarios with missing data. \papertitle's explicit delineation highlights gaps in FinNLP data, providing actionable direction for future dataset development and facilitating contributions from the broader research community.
    }
    \label{fig:methodology_domain}
\end{figure*}
\paragraph{Tasks.} In \papertitle, we consider six core FinNLP tasks (see \reffig{methodology_tasks}), each selected for their relevance to real-world financial applications, such as information retrieval, text classification, and sentiment analysis. The categories are designed to be broad enough to capture most FinNLP applications while remaining specific enough to support rigorous evaluation.
\paragraph{Domains.} Each dataset is classified by its domain, which considers what the data represents, who produced it, where it originates, when it was generated, how it was created, and why it exists. Domains include financial institutions, regulators, news media, small businesses, and individual investors. \citet{Liang2022-ew} organizes \textit{\textbf{domains}} primarily by the "3 W’s," describing what (genre of text), when (time period), and who (demographic or author source). We expand on this definition for finance by detailing additional attributes such as "where" origin (\eg \textit{specific} regulatory bodies) and "how" for data types (e.g., \textit{transcribed} earnings transcripts, \textit{human-annotated} SEC filings). This refinement ensures we capture the domain complexity unique to financial text sources.
\paragraph{Languages.} Our taxonomy currently focuses on English-language financial datasets but acknowledges the need for multilingual FinNLP resources, particularly for global markets.
\subsection{Datasets}\label{sec:datasets}
We construct \papertitle’s dataset suite according to explicit selection criteria that ensure \textbf{financial domain relevance}, \textbf{fair usage licensing}, \textbf{annotation quality}, and \textbf{task substance}.  Datasets must focus primarily on \textit{financial} text rather than tangential business or economic references. We exclude datasets that are not publicly available to researchers, without research-friendly licensing, or that do not explicitly credit original data authors. While \papertitle primarily covers \textbf{\textit{core}} NLP tasks (\reffig{methodology_tasks}), certain \textbf{\textit{frontier scenarios}} (e.g., decision-making, tool-use, market forecasting) lie outside this initial scope. These tasks require deeper domain knowledge, additional metrics, and robust guardrails. We aim to incorporate them in future expansions. 

After applying the above criteria, we selected \numtasks datasets for \papertitle.
\refapp{datasets} provides a complete list of each dataset, along with domain type, annotation method, and usage license. We perform \textbf{quality assurance} on each dataset for label consistency, domain specificity, and minimal data leakage. When previous studies or the community flag serious issues (e.g., skewed entity labeling, incomplete coverage), we either exclude the dataset or advise caution. For instance, prior work identified that some “CRA NER” corpora have oversimplified entity types, potentially distorting real-world distribution. We exclude such datasets or relegate them to an \emph{experimental} status if they do not meet our threshold for reliability. We also exclude benchmarks that attempt purely numeric or time-series forecasting with no natural language component, as these do not align with our focus on core NLP tasks. Please see \refapp{datasets} for full details on data selection criteria, along with additional discussion on data leakage, recommended salted hashes, and excluded datasets.
\subsection{Evaluation}\label{sec:models}
\para{Models.} We select models that are not multi-modal to focus our study on their NLP capabilities. Multi-modal models are an aspect of frontier research that deserves a separate dedicated research study (see \refapp{incomplete} for details).\\
We study the following LM families with \papertitle: Proprietary closed source systems GPT 4o \& o1-mini, Gemini-1.5, Claude3, and Cohere Command R. Along with open weight models including Llama 3, DeepSeekV3 \& R1, Qwen-2 \& QwQ, Mistral, Gemma-1 \& 2, Mixtral, WizardLM2, and DBRX. All experiments involving large language models (LMs) were conducted using cloud-based APIs. We utilized commercial API access for the proprietary models listed above, such as OpenAI’s GPT, Google’s Gemini, and Anthropic’s Claude, and others.\\
We include known details on foundation LMs such as architecture, training data, and model parameters, etc. in \reftab{model-detail}. Results from open-source or open-weight models offer greater transparency into LM performance compared to closed-source systems, due to the lack of reproducibility and transparency regarding any closed-source models or systems.\\
\para{Extractions.} During evaluation, the primary language model generates responses to task-specific inputs. These responses undergo a structured extraction process using a separate language model to identify relevant output elements. This two-stage approach separates the generation and extraction steps, enabling robust evaluation across different response formats. The extraction phase employs rule-based pattern matching and regular expressions to identify specific elements within LM outputs. This systematic approach ensures consistent response parsing across different tasks and model architectures. The framework maintains separate evaluation criteria for financial classification, numerical reasoning, and text generation tasks.\\
\para{Evaluation.} Performance measurement occurs through task-specific metrics, including accuracy, F1 scores, precision, recall, and BLEU scores for generation tasks. These metrics are computed using standardized implementations to ensure consistency across evaluations. \papertitle aggregates results by grouping scores according to task categories and financial domains. A configurable weighting system allows adjustment of score importance based on task difficulty and domain relevance. The final meta-score computation accounts for the relative performance range of models across tasks, providing a balanced assessment of financial language understanding capabilities.\\
\para{Generation.} Decoding strategies are methods that determine how an LM generates text tokens \cite{Wiher2022-pe}. Decoding strategy involve different settings for temperature, top-p, and repetition penalty, which influence the randomness and diversity of the output token sequence. Our \textit{‘deterministic’} strategy uses a temperature of 0.0, top-p of 0.9, and when possible, a repetition penalty of 1. We chose this deterministic decoding strategy to obtain predictable and consistent results across samples, which is crucial for benchmarking where accuracy and reliability are emphasized.. Deterministic decoding is most important for tasks common in finance such as data extraction or structured text generation due to the improved performance from low temperatures \cite{Liang2024-cz,Zarriess2021-ty}
\begin{table*}[h]
\centering
\resizebox{\textwidth}{!}{%
\begin{tabular}{|l|cccc|c|ccc|cc|ccccc|ccc|cc|}
\hline
Dataset Type & \multicolumn{4}{c|}{Information Retrieval} & \multicolumn{1}{c|}{*}& \multicolumn{3}{c|}{Sentiment Anal.} & \multicolumn{2}{c|}{Causal Anal.} & \multicolumn{5}{c|}{Text Classification} & \multicolumn{3}{c|}{Question Answering} & \multicolumn{2}{c|}{Summarization} \\ 
\hline
Model/Dataset & FiNER & FR & RD & FNXL & FE & FiQA & SQA & FPB & CD & CC & B77 & FB & FOMC & NC & HL & CFQA & FinQA & TQA & ECTSum & EDTSum \\
\hline
Metric Used & \multicolumn{5}{c|}{F1 Score} & \multicolumn{1}{c|}{MSE} & \multicolumn{8}{c|}{F1 Score} & \multicolumn{4}{c|}{Accuracy}&\multicolumn{2}{c|}{BERTScore F1}\\ 
\hline
Llama 3 70B Instruct & .701 & .332 & .883 & .020 & .469 & .123 & .535 & .902 & .142 & .192 & .645 & .309 & .652 & .386 & .811 & .709 & .809 & .772 & .754 & \cellcolor{green!50}{.817} \\ 
Llama 3 8B Instruct & .565 & .289 & .705 & .003 & .350 & .161 & \cellcolor{Green!70}{.600} & .698 & .049 & .234 & .512 & .659 & .497 & .511 & .763 & .268 & .767 & .706 & .757 & .811 \\ 
DBRX Instruct & .489 & .304 & .778 & .009 & .006 & .160 & .436 & .499 & .087 & .231 & .574 & .483 & .193 & .319 & .746 & .252 & .738 & .633 & .729 & .806 \\ 
DeepSeek LLM (67B) & .745 & .334 & .879 & .007 & .416 & .118 & .462 & .811 & .025 & .193 & .578 & .492 & .407 & .151 & .778 & .174 & .742 & .355 & .681 & .807 \\
Gemma 2 27B & .761 & .356 & .902 & .006 & .298 & \cellcolor{Green!70}{.100} & .515 & .884 & .133 & .242 & .621 & .538 & .620 & .408 & .808 & .268 & .768 & .734 & .723 & .814 \\ 
Gemma 2 9B & .651 & .331 & .892 & .005 & .367 & .189 & .491 & \cellcolor{green!50}{.940} & .105 & .207 & .609 & .541 & .519 & .365 & \cellcolor{Green!70}{.856} & .292 & .779 & .750 & .585 & \cellcolor{green!50}{.817} \\
Mistral (7B) Instruct v0.3 & .526 & .276 & .771 & .004 & .368 & .135 & .522 & .841 & .052 & .227 & .528 & .503 & .542 & .412 & .779 & .199 & .655 & .553 & .750 & .811 \\ 
\emph{Mixtral-8x22B Instruct} & .635 & .367 & .811 & .009 & .435 & .221 & .510 & .776 & .125 & \cellcolor{Green!70}{.308} & .602 & .221 & .465 & .513 & \cellcolor{green!20}{.835} & .285 & .766 & .666 & .758 & .815 \\ 
\emph{Mixtral-8x7B Instruct}z & .598 & .282 & .845 & .009 & .267 & .208 & .498 & .893 & .055 & .229 & .547 & .396 & .603 & .583 & .805 & .315 & .611 & .501 & .747 & .810 \\ 
\emph{Qwen 2 Instruct (72B)} & .748 & .348 & .854 & .012 & .483 & .205 & .576 & .901 & .190 & .184 & .627 & .495 & .605 & .639 & .830 & .269 & .819 & .715 & .752 & .811 \\ 
\emph{WizardLM-2 8x22B} & .744 & .355 & .852 & .008 & .226 & .129 & .566 & .779 & .114 & .201 & .648 & .500 & .505 & .272 & .797 & .247 & .796 & .725 & .735 & .808 \\ 
\emph{DeepSeek-V3} & \cellcolor{green!20}{.790} & \cellcolor{green!50}{.437} & .934 & \cellcolor{green!20}{.045} & .549 & .150 & \cellcolor{green!20}{.583} & .814 & \cellcolor{green!20}{.198} & .170 & \cellcolor{green!50}{.714} & .487 & .578 & .675 & .729 & .261 & \cellcolor{green!50}{.840} & \cellcolor{green!20}{.779} & .750 & .815 \\ 
\emph{\textbf{DeepSeek R1}} & \cellcolor{Green!70}{.807} & .393 & \cellcolor{Green!70}{.952} & \cellcolor{Green!70}{.057} & \cellcolor{green!20}{.587} & .110 & .499 & .902 & \cellcolor{Green!70}{.337} & .202 & \cellcolor{Green!70}{.763} & .419 & \cellcolor{green!50}{.670} & .688 & .769 & \cellcolor{Green!70}{.853} & \cellcolor{green!20}{.836} & \cellcolor{Green!70}{.858} & .759 & .804 \\ 
\textbf{QwQ-32B-Preview} & .685 & .270 & .656 & .001 & .005 & .141 & .550 & .815 & .131 & .220 & .613 & \cellcolor{green!50}{.784} & .555 & .020 & .744 & .282 & .793 & \cellcolor{green!50}{.796} & .696 & \cellcolor{green!50}{.817} \\ \hline
Jamba 1.5 Mini & .552 & .284 & .844 & .005 & .132 & .119 & .418 & .765 & .043 & \cellcolor{green!50}{.270} & .508 & \cellcolor{Green!70}{.898} & .499 & .151 & .682 & .218 & .666 & .586 & .741 & .816 \\ 
Jamba 1.5 Large & .693 & .341 & .862 & .005 & .397 & .183 & .582 & .798 & .074 & .176 & .628 & .618 & .550 & .541 & .782 & .225 & .790 & .660 & .734 & \cellcolor{Green!70}{.818} \\ 
Claude 3.5 Sonnet & \cellcolor{green!50}{.799} & \cellcolor{Green!70}{.439} & .891 & \cellcolor{green!50}{.047} & \cellcolor{green!50}{.655} & \cellcolor{green!50}{.101} & .553 & \cellcolor{Green!70}{.944} & .196 & .197 & .668 & .634 & \cellcolor{Green!70}{.674} & \cellcolor{green!20}{.692} & .827 & .402 & \cellcolor{Green!70}{.844} & .700 & \cellcolor{green!20}{.767} & .813 \\
Claude 3 Haiku & .711 & .285 & .883 & .015 & .494 & .167 & .463 & .908 & .081 & .200 & .622 & .022 & .631 & .558 & .781 & .421 & .803 & .733 & .646 & .808 \\ 
Cohere Command R 7B & .748 & .194 & .845 & .018 & .441 & .164 & .532 & .840 & .057 & \cellcolor{green!20}{.255} & .516 & \cellcolor{green!20}{.762} & .459 & .068 & .770 & .212 & .709 & .716 & .750 & .815 \\ 
Cohere Command R + & .756 & .333 & .922 & .021 & .452 & \cellcolor{green!20}{.106} & .533 & .699 & .080 & .238 & .651 & .684 & .393 & .118 & .812 & .259 & .776 & .698 & .751 & .810 \\
Google Gemini 1.5 Pro & .712 & .374 & \cellcolor{green!50}{.944} & .019 & .393 & .144 & \cellcolor{green!50}{.593} & .885 & .196 & .217 & .418 & .336 & .579 & .525 & \cellcolor{green!50}{.837} & .280 & .829 & .763 & \cellcolor{Green!70}{.777} & \cellcolor{green!50}{.817} \\ 
OpenAI gpt-4o & .766 & .399 & \cellcolor{green!20}{.942} & .037 & .523 & .184 & .541 & \cellcolor{green!20}{.928} & .130 & .222 & \cellcolor{green!20}{.710} & .524 & \cellcolor{green!20}{.664} & \cellcolor{Green!70}{.750} & .824 & \cellcolor{green!20}{.749} & .836 & .754 & \cellcolor{green!50}{.773} & .816 \\ 
\textbf{OpenAI o1-mini} & .761 & \cellcolor{green!20}{.403} & .876 & .010 & \cellcolor{Green!70}{.662} & .120 & .542 & .917 & \cellcolor{green!50}{.289} & .209 & .670 & .612 & .635 & \cellcolor{green!50}{.720} & .769 & \cellcolor{green!50}{.840} & .799 & .698 & .763 & .816 \\ \hline
\end{tabular}

}
\caption{\textbf{Overview of \papertitle Results.} This table compares results across all datasets and all models in \papertitle. We note reasoning-reinforced models as \textbf{bold text} and mixture of expert models with \emph{italics}. For full dataset details, see \refapp{datasets}. * indicates the dataset belongs in both IR and SA.}
\label{tab:main_table}
\end{table*}
\section{Experiments and Results}\label{sec:results}
In this section, we present the results of our holistic evaluation of LMs across a variety of core NLP tasks for finance, focusing on multiple dimensions: \emph{performance} and \emph{efficiency} in terms of inference overhead and cost.
We evaluated \nummodels{} language models (LMs) on the \papertitle{} benchmark suite.
\reftab{main_table} provides a high-level scoreboard across six main task categories:\footnote{Datasets are introduced in \refsec{datasets}.} 
We also detail each dataset's unique domain requirements, the metrics used, and final model performances in separate tables (see \refapp{results}).
Overall, the results reveal three key insights:
\begin{enumerate}
    \item No single LM performs the best across all tasks, but a handful of models show strong overall performance.
    \item Performance depends heavily on the domain and task structure, \eg numeric reasoning vs entity classification. 
    \item Open-weight and mid-scale models demonstrated strong cost/performance efficiency, highlighting the importance of further scientific research.
\end{enumerate}
We organize the following subsections around a {meta-analysis} of our results across all models. For the \textit{model-specific} observations or \textit{per-task} discussion, please refer to \refapp{task_results}
\subsection{Meta-Analysis of Results}
\label{sec:meta-analysis}
\para{Key Takeaways.} 
\reftab{main_table} shows that certain LMs consistently perform well on multiple tasks\emdash~\eg \textbf{DeepSeek R1} leads in many IR tasks and advanced QA settings, \textbf{Claude 3.5 Sonnet} excels in sentiment (\textsc{FPB}) and some IR tasks (\textsc{FinRED}), and \textbf{GPT-4o} hovers near the top in classification and summarization. 
Nevertheless, there was \emph{no single model that wins overall}: while \textbf{DeepSeek R1} dominates multi-step QA (e.g., \textsc{ConvFinQA}, \textsc{TATQA}), trails in summarization. Performance can vary between even similar tasks, as \textbf{Claude 3.5 Sonnet} leads \textsc{FinQA}, but not necessarily multi-turn \textsc{ConvFinQA}.\\
\para{Domain-Specific Challenges.} 
Numeric reasoning tasks (like \textsc{FNXL} for numeric labeling or \textsc{ConvFinQA} for multi-step financial statements) remain especially challenging, with F1 scores for \textsc{FNXL} often below 0.06, signaling that even large models struggle to precisely map an extremely large amount of categories to numeric content. The relatively low scores on \textsc{ConvFinQA} compared to basic classification or retrieval tasks like \textsc{REFinD} and \textsc{Headlines} suggest that LMs suffer from sharp performance drops on tasks requiring step-by-step deductions, calculations, or cross-referencing, which could impede their application to financial forecasting and decision-making.

By contrast, summarization tasks yield relatively high BERTScores (0.75\dash0.82 for most models), indicating that summarization in financial contexts\emdash though non-trivial\emdash seems more tractable or amenable to the generic capabilities of foundation LMs. This could be due to those tasks only requiring LMs to identify and output the key parts of the input task, rather than having to generate text or reason through a problem.\\
\para{Inconsistent Scaling.}
Our results corroborate that \emph{larger parameter sizes do not strictly guarantee higher performance}: 
For instance, \textsc{Jamba 1.5 Mini} outperforms many bigger models in \textsc{FinBench}, and \textsc{Gemma 2 9B} can match or exceed larger model variants on \textsc{Banking77} or \textsc{Headlines}.

\subsection{Further Error Analysis and Discussion}

In addition to the aggregate results, we highlight some error patterns:

\para{Numeric Reasoning Gaps.} 
Despite partial success in \textsc{FinQA} or \textsc{ConvFinQA}, many LM outputs fail to produce consistent numeric or textual formats (e.g., rounding vs.\ decimal, underscore vs. dash) or handle cross-sentence references. This can be especially detrimental in \textsc{FNXL} labeling.

\para{Language Drift and Prompt Issues.}
Some models (e.g., Qwen\,2\,72B) occasionally drift into non-English outputs for summarization. Additionally, longer label sets (e.g., \textsc{Banking77} with 77 classes) can yield off-list label predictions, decreasing F1 scores. This could be due to models struggling to precisely remember everything in their context window.

\para{Causal Data Scarcity.}
Given the specialized financial domain, training data for causal detection or classification is limited. Our results reinforce that this scarcity remains a bottleneck; external knowledge or additional reasoning modules might be necessary to improve performance on causal tasks.

A detailed summary of model‐specific and task‐specific errors is provided in Appendix~\ref{app:error-analysis}, Tables~\ref{tab:error-by-model} and~\ref{tab:error-by-task} respectively.

\subsection{Efficiency Analysis of Model Performance}\label{sec:meta-analysis-results}
Beyond raw accuracy or F1, a critical factor for FinNLP is \emph{efficiency}. Tasks such as multi-turn financial question answering (\textsc{ConvFinQA}) and advanced causal classification require lengthy in-context prompts, leading to high inference costs. Notably, smaller models sometimes outperform larger ones by offering a superior trade-off between \emph{throughput} and \emph{accuracy}, making them more viable for real-world applications.\\
For all of our inference runs, DeepSeek R1 cost approximately \$260 USD compared to Claude 3.5 Sonnet's and o1-mini's ~\$105 USD and Meta Llama 3.1 8b's ~\$4 USD. This dramatic price difference suggests that users should choose models carefully based on use-case, as slightly lower performing models might have dramatically cheaper inference costs. For example, models such as Llama 3.1 70b and DeepSeek-V3 cost less than \$25 USD.\\(See \refapp{efficiency-analysis} for full details and cost.)
\section{Conclusion}\label{sec:conclusions}
We present \papertitle, a robust evaluation framework and open-source software package for conducting holistic evaluation of language models for finance. \papertitle provides standardized multi-metric evaluation for finance-specific datasets and evaluation methods. This framework provides a valuable foundation for building, testing, and advancing high-performance NLP models tailored to the unique challenges of financial language understanding. We believe that the adoption of a collaborative evaluation framework like \papertitle will be used by researchers to easily conduct holistic evaluations of any generally available LM for core FinNLP tasks.\\

Our evaluation underscores the complex landscape of FinNLP. Our key insights are as follows: 
\begin{enumerate}
    \item No single LM outperforms all others across every task, but a few models \emdash namely Deepseek R1, OpenAI o1-mini, and Anthropic Claude 3.5 Sonnet \emdash demonstrate strong overall performance. Despite their capabilities, these large models come with significant cost trade-offs compared to smaller, more affordable alternatives. 
    \item Model performance varies significantly based on the domain and task structure, with notable differences observed between tasks such as summarization and multi-turn question answering. 
    \item Open-weight and mid-scale models such as DeepSeek-V3 and Llama 3.1 70B demonstrate a strong balance between cost-efficiency and performance, underscoring the need for further research to optimize their effectiveness in FinNLP.
    \item There is a notable dearth of datasets across most languages and tasks within the taxonomy. The predominant languages in FinNLP remain English and Chinese.
    \item The taxonomy is a collaborative and evolving framework that requires continuous expansion with additional tasks to adapt to the field's advancements.
\end{enumerate}

Key directions for future research include advanced prompt engineering, domain-adaptive training (particularly for numeric/causal tasks), and benchmarking efficiency trade-offs. We hope these results guide both industry practitioners and NLP researchers in developing robust financial systems.

\section{Limitations}\label{sec:limitations}
\papertitle has several notable limitations that should be acknowledged. First, there are many limitations to be noted that together could significantly impact the robustness and reliability of \papertitle. We discuss these limitations in extensive detail to illuminate the community on where we believe the most effort is needed for additional research. The recognition of incompleteness is a major requirement for holistic LM evaluation.
The limited size and diversity of datasets significantly affects our ability to measure the robustness and generalization of model performance across different scenario contexts. We highlight these areas of incompleteness with out taxonomy.

Budgets associated with computational cost were another major limiting favor for our study. In order to gather so many results from high-cost proprietary models, we conducted only zero-shot evaluations. We acknowledge the limitation of this research as techniques such as chain-of-thought and program-of-thought can significantly increase inference costs. 
Adaptation (i.e. model prompting techniques) are not covered within this paper as the importance of in-context learning, structured analytical techniques, or evoking chains of 'reasoning' all are deserving of their own individual study. The benefit of these techniques has been noted and is worth of further research. The goals of our study are to focus on the zero-shot un-adapted and un-augmented performance of the selected foundation LMs. We believe that existing research has demonstrated the benefits of these techniques enough to warrant widespread adoption and therefore allocated the computational budget towards exploring more models rather than prompt engineering.

Finally, the tasks associated with the first version of \papertitle all primarily rely on the English language due to English being the primary language of not only the authors, but of many FinNLP benchmarks /cite{Longpre2024-op}. The focus on English for this \textbf{\textit{first iteration}} of \papertitle limits our ability to draw conclusions on multi-lingual performance for these models. However, the authors already have begun work to expand our benchmark to include multi-lingual coverage. 

Further, we solve for this limitation by establishing a living and community-governed benchmark for researchers to collaboratively build. We seek collaboration to work alongside other researchers to continually push for updates with new tasks and models. To assist others, we defined clearly and narrowly defined requirements for inclusion along with a standardized python implementation recipe to ensure fair evaluation in \refapp{datasets} and \refapp{ethicslegal}.
Despite our efforts to include a wide range of tasks, these datasets do not even begin to capture the breadth and complexity of human cognition required for real-world financial scenarios. The current tasks  overlook many highly specialized use cases, local or regional knowledge, or emerging financial products or events. 

Finally, although \papertitle is easily extensible, the nature of changes in financial academics and practice means that benchmarks can lose their effectiveness. Modern financial economics undergoes rapid evolution and change. Due to this dynamic nature it is very difficult for any benchmark to capture the variability of out-of-sample data. By adopting a collaborative and extensible framework for our benchmark suite, we attempt to mitigate the risks associated with benchmarks becoming trivial to solve or irrelevant to current practice. For a full in-depth discussion recognizing the incompleteness and the limits of this work, please refer to \refapp{incomplete}.\\

\section*{Ethics Statement}\label{sec:ethics}
All datasets and resources in this benchmark are used and shared per their respective licenses. We have audited the license of each included dataset and provided this information in our documentation. The ACL responsible research checklist recommends providing license or terms of use for any dataset or software artifact \cite{ACL2022-qf}. We follow this by explicitly stating each dataset's license (e.g., CC-BY, MIT, etc.) in \ref{app:datasets} and documentation. We will update the final manuscript for publication to include all details on the leaderboard, an exploration of the user experience, and visualizations for metrics.
Finally, the authors of \papertitle disclaim and do not accept any liability for financial damages or losses associated with the use of the materials contained within this manuscript. This document and its related materials are only for academic and educational purposes. No commentary provided by the authors or this manuscript should used as financial, investing, or legal advice. Readers of our findings should seek the consultation of professionals before any use of these materials. Any use of our academic research constitutes indemnification of the authors against any claims from its use. Please see Appendix \ref{app:ethicslegal} for further discussion on our research's ethics and legal aspects, along with our proposed collaboration's governance policies. During writing, the authors used AI tools, including ChatGPT, Gemini, and Claude, for writing assistance, editing, and LaTeX code generation. All usage was in accordance with ACL guidelines and limited to non-substantive tasks, such as formatting, grammar suggestions, and refining phrasing. No AI-generated text was included as original scientific contributions in this work.
\section*{Acknowledgments} \label{sec:acknowledgements}
First, we collectively would like to thank our anonymous reviewers for their comments and feedback. We appreciate the help with revisions and reviews from Isaac Song, Siddharth Siddharth, Aryan Shah, and Anant Gupta. GM would like to acknowledge Agam Shah for his work on FLUE, and Michael Galarnyk for his advice and feedback. This work is funded in part by generous support to GM from \textsc{TogetherAI}.
\bibliography{paperpile}
\appendix
\section*{Contributions} \label{sec:contributions}
\begin{enumerate}
    \item \textbf{Glenn Matlin}
    \begin{enumerate}
    \item Primarily responsible for the the core methodology of the paper, the holistic benchmarking approach, designed the taxonomy and qualitative evaluation, and developed the engineering design of the software  framework.
    \item Managed the curation and processing of benchmark data, and lead the design and processing of evaluations.
    \item Author of manuscript and appendix, and created figures, ran statistical analyses, and interpreted results.
    \item Coordinated the collaboration of team members, provided leadership, facilitated discussions and experiments.
    \item Responsibility for addressing feedback from any issues arising from the current state of the project, management of the project's software repository and its software framework,  and the incorporation of new datasets and evaluation methods.
    \end{enumerate}
    \textit{GM wishes to emphasize that co-authors should not be held liable for errors, as they are not responsible for updating \papertitle.}
    \item \textbf{Mika Okamoto}
    \begin{enumerate}
        \item Contributed significantly to the overall framework of the paper and design and development of the codebase.
        \item Developed core components of the Python package, focusing on inference, extraction, and evaluation modules. 
        \item Authored parts of sections 1 (Introduction), 3 (Methodology), 4 (Experiments and Results), 5 (Conclusion), and parts of the Appendix. 
        \item Revised and proofread the manuscript to enhance clarity, coherence, and technical accuracy.
        \item Assisted with dataset curation in Hugging Face.
        \item Conducted inference and evaluation experiments on datasets FinRED, FinQA, EDTSum, REFinD, FiQA, FPB, Banking77, FOMC, FinBench, News Headline, and Causal Detection; also supported experiments on TATQA and ConvFinQA.
        \item Ideation and implementation of LLM-as-a-judge evaluation method and metric, applied to FinQA, TATQA, and ConvFinQA. 
        \item Performed efficiency and cost-performance analysis of models and tasks; analyzed token lengths and costs, created related figures, and authored Section 4.3 and Appendix F.4.
        \item Worked on error analysis efforts by collecting failure cases and analyzing model performance.
        \item Applied prompt optimization techniques to improve base prompt effectiveness across various tasks.
    \end{enumerate}
    \item \textbf{Huzaifa Pardawala}
\begin{enumerate}
  \item Significant contributions to the main research objectives and overall framework of the paper.
  \item Contributed to the development of the Python package from scratch to support data ingestion, preprocessing, and evaluation.
  \item Authored parts of sections 1 (Introduction), 4 (Experiments and Results), 5 (Conclusion), and 6 (Limitations).
  \item Wrote Appendix subsections B, C, D, E, and F, detailing task definitions, dataset descriptions, model descriptions, and evaluation protocols.
  \item Performed revisions and proof‐reading to improve clarity, flow, and technical accuracy.
  \item Collected and curated data for eight datasets, including uploading each to Hugging Face with full dataset cards.
  \item Conducted inference and evaluation experiments on FNXL, FinEntity, ECTSum, NumClaim, SubjECTive‐QA, Causality Classification, and FiNER-ORD.
  \item Ideation and implementation of new evaluation metrics for FinEntity and FNXL.
\end{enumerate}

    \item \textbf{Yang Yang}
    \begin{enumerate}
    \item Significant contributions to the main research objectives and overall framework of the paper.
    \item Contributed to section 4 (Experiments and Results) and section C of the Appendix.
    \item Created initial end-to-end inference and extraction scripts for FinQA, ConvFinQA, TATQA.
    \item Conducted inference and evaluation experiments on ConvFinQA and TATQA.
    \item Collected and curated data for five datasets, including uploading the cleaned data to Hugging Face.
\end{enumerate}
    \item \textbf{Sudheer Chava} provided their expertise and feedback which provided valuable perspective throughout this project and helped refine our approach at key stages.
\end{enumerate}
\section{Taxonomy of Financial Scenarios}\label{app:taxonomy}
\paragraph{Tasks.}
We focus on six core NLP tasks \dash \textit{question answering}, \textit{information retrieval}, \textit{summarization}, \textit{sentiment analysis}, \textit{toxicity detection}, and \textit{text classification} category for miscellaneous labeling tasks. These tasks are \textit{user-facing} for finance: they reflect practical objectives like extracting key information from company filings, summarizing earnings reports, detecting false or harmful content in financial forums, and classifying transactions or documents. Although many sub-categories of tasks exist within each broad task category (\eg, named entity recognition, structured boundary detection, causal reasoning), we group them under broader categories where possible, to keep the focus on the end-user or enterprise-facing application in financial scenarios.

\begin{table*}[h!]
  \centering
   \large                             
  \resizebox{\linewidth}{!}{%
    \renewcommand{\arraystretch}{1.15}

\begin{tabular}{|l|p{3.5cm}|p{4cm}|p{4.5cm}|p{3.75cm}|p{2.5cm}|p{3cm}|p{2cm}|}
\hline
\textbf{Dataset} & \textbf{Task} & \textbf{What} & \textbf{Who} & \textbf{Where} & \textbf{When} & \textbf{How} & \textbf{Language} \\
\hline
FinQA & QA & Earnings reports & S\&P 500 companies & Collected from FinTabNet dataset & 1999-2019 & Expert annotation & EN \\
\hline
ConvFinQA & QA & Earnings reports & S\&P 500 companies & Built on top of FinQA dataset & 1999-2019 & Expert annotation & EN \\
\hline
TAT-QA & QA & Tables and relevant text from 500 financial reports & Public companies & www.annualreports.com & 2019-2021 & Expert annotation & EN \\
\hline
ECTSum & TS & Transcripts of earnings calls & Russell 3000 Index companies & The Motley Fool & 2019-2022 & Analysts and experts wrote summaries for the ECTs & EN \\
\hline
EDTSum & TS & News articles including a type of corporate event & PRNewswire, Businesswire, GlobeNewswire authors & PRNewswire, Businesswire, GlobeNewswire & 2020-2021 & Sampling and filtering based on corporate event & EN \\
\hline
FiNER-ORD & IR & Financial news articles & Article writers, 10-K filings came from public companies & webz.io & NS & Manual annotation & EN \\
\hline
FinRED & IR & 47,851 finance news articles and 4,713 earnings call transcripts & Public companies for ECT data & Webhose and Seeking Alpha & Jun 2019 - Sep 2019 & NS & EN \\
\hline
REFinD & IR & 10-K filings from SEC & Public companies & SEC database & 2016-2017 & Crowdsourced annotation that was reviewed by experts & EN \\
\hline
FNXL & IR & Filings for 2,339 companies & Public company filings & SEC database & 2019-2021 & Annotations made by the company that is filing & EN \\
\hline
FinEntity & IR; SA & Finance news & Reuters & Refinitiv Reuters DB & NS & 12 senior undergrads in finance or business annotated & EN \\
\hline
SubjECTive-QA & SA & 2,747 QA Pairs from Earnings Calls Transcripts & NYSE Companies & Investor relations' sections of the companies' websites & 2007-2021 & Manual annotation & EN \\
\hline
FiQA & SA & Social media and investor forums & Individuals and households & Social media (i.e., Reddit) & 2016-2018 & Crowdsourced & EN \\
\hline
FPB & SA & 10,000 finance news articles & All companies in OMX Helsinki & LexisNexis database & NS & 16 annotators with adequate financial knowledge & EN \\
\hline
NumClaim & TC & Analyst reports and earnings call reports & Analysts and NASDAQ 100 companies & Zacks Equity Research and public data & 2017-2020 for analyst reports and 2017-2023 for earnings calls & Manual annotation & EN \\
\hline
Banking77 & TC & 13,083 customer service queries with 77 intents & NS & Customer service interactions & NS & NS & NS \\
\hline
FinBench & TC & 10 high-quality datasets for financial risk prediction & Dataset creators & Kaggle & NS & NS & NS \\
\hline
News Headline Classification & TC & 11,412 news headlines about commodities, particularly gold & NS & Reuters, The Hindu, The Economic Times, Bloomberg, Kitco, MetalsDaily, etc. & 2000-2019 & Expert annotation & NS \\
\hline
FOMC & TC & FOMC meeting mins, press conference transcripts, and speeches & Federal Open Market Committee & www.federalreserve.gov & Meeting mins 1996-2022 and press conference transcripts 2011-2022 & Manual annotation & EN \\
\hline
FinCausal-SC & CA & Financial news provided (14,000 websites) & Article writers & Qwam & 2019 & Individual authors & NS \\
\bottomrule
\end{tabular}
\label{tab:taxonomy_data}
  }

  \caption{Financial NLP datasets and their characteristics.  
           IR = Information Retrieval, SA = Sentiment Analysis, TS = Text Summarization, QA = Question Answering, CA = Causal Analysis, 
           TC = Text Classification, EN = English, NS = Not Specified.}
  \label{tab:app_taxonomy_data}
\end{table*}

\paragraph{Domains.}
We define a domain by \textit{what} is the type of data, \textit{who} produced it, \textit{when} it was created, \textit{where} did it originate, \textit{how} it was generated, and \textit{why} is it useful. Examples of domains include (i) \textit{publicly-traded corporations} producing investor filings, (ii) \textit{regulatory bodies} issuing policies and enforcement documents, (iii) \textit{news media} offering breaking market updates, (iv) \textit{SMBs} managing internal accounting ledgers, and (v) \textit{individual investors} discussing trades on social media. Each domain introduces unique formats (\eg, structured filings vs. informal posts) and unique constraints (\eg, legal compliance vs. personal expression). By taxonomizing these domains, researchers can use \papertitle to identify coverage gaps and propose new benchmark datasets for under-served financial scenarios
\paragraph{What (Type of Data/Annotations).} This refers to the nature of the dataset, whether it includes structured financial records (e.g., SEC filings), informal text (e.g., social media discussions), regulatory reports, or analyst commentary. Annotations can range from human-labeled categories to machine-generated insights.
\paragraph{Who (Data Source).} The entity that produced or collected the dataset, such as individuals (personal finance data), businesses (corporate records), financial institutions (bank transactions), regulators (policy statements), or media sources (news articles).
\paragraph{Where (Data Origination \& Distribution).} The source repository of the dataset \dash e.g., regulatory databases, company websites, news platforms, or user-generated content from social media.
\paragraph{When (Time Sensitivity \& Temporal Scope).} The time period of the dataset, distinguishing between historical, recent, and real-time data. Financial data has strong temporal relevance, affecting its usability for different research tasks.
\paragraph{How (Data Generation \& Annotation).} Describes whether the dataset was self-reported, institutionally recorded, scraped from public sources, or generated synthetically. Annotation can be performed by experts, crowd workers, automated scripts, or AI models.
\subsection{Tasks}\label{app:tasks}
\paragraph{Question answering.}
In financial QA, models answer questions about company disclosures, regulatory text, or market data. For example, a user may ask, “What was Company X’s net income last quarter?” or “Under which clause must this fund disclose assets?” These tasks can be open-book (access to filings or transcripts) or closed-book (testing a model’s internalized domain knowledge). Accuracy and factual correctness are paramount, as erroneous answers can mislead analysts or investors.
\paragraph{Information retrieval.}
Here, the system locates relevant text or documents from large financial corpora, such as retrieving the correct section in an SEC filing that addresses a particular risk factor. This typically involves ranking passages or paragraphs by relevance. Good performance in financial IR helps analysts quickly navigate extensive disclosures, saving time and reducing information overload.
\paragraph{Summarization.}
Summaries condense lengthy financial documents like earnings reports or regulatory proposals into concise abstracts. Abstractive summarization can highlight key takeaways for investors, while extractive approaches ensure faithfulness to the original text. Faithfulness is critical in finance; hallucinated or misleading summaries can create compliance issues or misinform market participants.
\paragraph{Sentiment analysis.}
Sentiment tasks in finance often involve gauging the emotional tone of news headlines, social media chatter, or analyst commentary. Models can help traders or risk managers track public sentiment around specific stocks, detect shifts in market mood, or monitor customer feedback. Unlike general sentiment tasks, financial sentiment often leans heavily on domain-specific lexicons and context (\eg, “downward revision” vs. “positive guidance”).
\paragraph{Causal Analysis.}
Causal analysis in finance focuses on identifying cause-and-effect relationships within economic events, financial policies, or market movements. Models can help analysts determine whether a policy change influenced stock prices or assess the impact of macroeconomic factors on investment trends. Unlike general causal inference tasks, financial causal analysis often relies on structured data, temporal dependencies, and domain-specific knowledge (\eg, “interest rate hike leading to capital outflows” vs. “regulatory easing boosting market liquidity”).
\paragraph{Text classification.}
Beyond these core tasks, many finance-specific classification needs arise, such as identifying fraudulent activities (\eg, “phishing scam” vs. “legitimate inquiry”), labeling compliance documents by topic, or categorizing support tickets (\eg, “credit card issue” vs. “mortgage application”). This \emph{miscellaneous} category accommodates various text classification tasks at different granularity.

\subsection{Domains}\label{app:domains}
\subsubsection{What}\label{app:domain_what}
\textit{"What is the type of data/annotations?"}
\paragraph{Personal Finances.} 
Personal finances include documents and records related to individual households’ finances. This category broadly covers self-generated financial records such as personal budgets, expense logs, cash flow statements, and official documents like individual income tax filings (\eg, IRS Form 1040). In addition, the category covers data collected about individuals by financial institutions, including bank statements, transaction logs, and credit reports. These data sources are used in various NLP tasks such as information extraction, summarization, sentiment analysis (\eg, for credit risk), and the generation of personalized financial advice. A clear distinction should made between first-party data (directly produced or owned by individuals) and third-party data (collected about individuals by institutions), with derived data and metrics (\eg, credit reporting and scores) recognized as distinct types.
\paragraph{SMB Finances.} 
Small and Medium Business (SMB) finances include the financial records generated and maintained by small enterprises. This category comprises internal documents such as accounting statements (balance sheets, income statements, and cash flow statements), invoices, payroll records, and business tax filings. It also encompasses external data collected about SMBs by financial institutions and credit bureaus, such as transaction logs and business credit reports. NLP applications for this data focus on information extraction, text classification, and summarization tasks. The category includes data produced directly by SMBs (first-party data) and data collected by third-party entities (external assessments).
\paragraph{Social Media \& Investor Forums.} 
This includes content from public platforms where individual investors discuss financial markets. Social media posts are real-time and high-volume, often opinionated and informal (emojis, memes, humor, or hyperbole). Annotation often relies on crowd-sourcing of sentiment and toxicity labels. Examples of tasks include sentiment analysis, toxicity detection, text classification, and summarization. The category includes data (i.e., post text and image) produced directly by individuals (first-party), as well as data collected about the individual or their user behavior (third-party). 
\paragraph{Financial News \& Media.} 
Produced by major news agencies, news about current events and finance informs markets about macroeconomics, company earnings, and opinionated analysis. News types range from real-time reports and market analyses to press releases. Financial news is high-frequency, continuously updated, and distributed via news terminals, APIs, and web sources. Annotations can include topic categories, sentiment scores, and event classifications. NLP tasks include information retrieval, text classification, sentiment analysis, and summarization. 
\paragraph{Corporate Disclosures \& Filings.}
Corporate disclosures include financial reports such as 10-K annual reports, 10-Q quarterly reports, earnings call transcripts, and press releases. These documents are produced by public corporations, primarily for legal compliance, investor transparency, and shaping market sentiment. They consist of formal reports, earnings transcripts, and event-driven disclosures. The frequency varies, with periodic reports released annually or quarterly and event-driven disclosures appearing as needed. Creation follows regulatory formats, typically unannotated, but some datasets add expert labels for sentiment analysis and summarization. Distribution occurs through company websites, regulatory databases, and press release services. Example tasks include summarization, information extraction, sentiment analysis, and question-answering.
\paragraph{Regulatory \& Legal Disclosures.} 
This includes regulatory filings, policy statements, legislation, and central bank reports. Producers include financial regulators, central banks, and legislative bodies, aiming to ensure transparency, market regulation, and compliance guidance. These texts range from proposed rules and legislation to policy statements and enforcement actions, with varying publication frequency. Regulatory texts are formal and often lengthy, with limited public annotation. NLP tasks include text classification, summarization, information extraction, and stance detection.
\paragraph{Analyst \& Research Reports.} 
These reports are created by investment banks, rating agencies, and independent analysts to provide in-depth financial analysis and recommendations. They include equity research reports, macroeconomic outlooks, and credit rating evaluations, which are published periodically and are event-driven. Reports are proprietary, limiting public access, though some analyst reports appear in regulatory filings. NLP tasks include sentiment analysis, recommendation classification, summarization, and information extraction.
\paragraph{Emerging \& Alternative Finance.} This category includes cryptocurrency whitepapers, FinTech credit reporting data, and novel forms of financial products. Data producers range from blockchain communities to financial regulators. Alternative data is diverse in format and frequency. NLP tasks include entity recognition, scam detection, summarization, and bias analysis.
\subsubsection{Who}\label{app:domain_who}
\textit{"Who generated the data/annotations?"}
\paragraph{Individuals \& Households.} 
This category covers the financial data originating from individuals' activity. It includes self-generated financial records (such as budgets, expenses, and receipts) and data produced by financial institutions on behalf of individuals ( bank statements, loan documents, etc.).
\paragraph{Small and Medium Businesses (SMBs).} 
This category pertains to the financial data produced by SMBs. It involves internally generated documents such as accounting records, invoices, payroll information, and tax filings, alongside externally collected data like business credit reports and bank transaction records. NLP systems may use this data to automate financial management tasks, improve risk assessments, and facilitate credit underwriting for smaller enterprises. Differentiations are made between first-party data (generated by the SMB) and third-party data (collected about the SMB).
\paragraph{Commercial \& Retail Banks.} 
Banking institutions accept deposits, extend credit, and provide loans to consumers and businesses. Larger banks have lines of business that include retail banking (i.e., individual customers), business banking (small and medium companies), and commercial banking (enterprise clients) operations. They generate extensive text-based data, including annual reports, quarterly earnings reports, and shareholder letters. Regulatory reports such as SEC 10-K/10-Q forms disclose financials and risks. Internally, banks maintain risk management reports, compliance documents, and customer communications (emails, chat logs). Most internal documents are proprietary, while investor reports and required filings are public.
\paragraph{Investment Banks \& Brokerage Firms.} 
Investment banks facilitate securities offerings, mergers and acquisitions, and other complex financial transactions. Brokerage firms execute trades for clients. These institutions produce financial research reports, prospectuses, and offering memoranda for investment offerings. Internally, they generate pitch books, trading desk reports, and compliance documentation. Public documents include financial research and regulatory filings, while deal-related and internal reports remain proprietary.
\paragraph{Asset Management Firms.} 
Asset managers invest pooled funds on behalf of clients, including mutual funds, pension funds, and investment advisors. They produce fund prospectuses, shareholder reports, investor letters, and market outlooks. Internally, they maintain investment committee memos, research reports, and risk reports. Public mutual fund documents and investor letters are available, whereas internal research and risk memos usually remain confidential.
\paragraph{Hedge Funds \& Private Investment Firms.} 
Hedge funds and private investment firms manage private capital with flexible investment strategies. They produce strategy documents, trading models, and investor update letters. Capital-raising documents such as Private Placement Memoranda (PPM) outline strategies, risks, and terms. Regulatory filings like Form 13F are public, but trading strategies and internal risk/compliance reports remain confidential.
\paragraph{Insurance Companies.} 
Insurance firms underwrite risk policies and manage significant investment portfolios. They generate insurance policy contracts, actuarial reports, claims reports, and risk assessments. Regulatory filings include financial statements and risk-based capital reports. Public documents include policies and financial reports, whereas underwriting guidelines and claims analyses remain proprietary.
\paragraph{Regulators \& Central Banks.} 
Regulators oversee financial markets, ensuring stability and compliance. Examples include the Security Exchange Commission (SEC), the Federal Reserve, the Basel Committee on Banking Supervision, and the European Central Bank. These entities produce regulations and guidance documents, monetary policy statements, financial stability reports, and enforcement rulings. Many regulatory texts are public, though supervisory communications and compliance assessments remain private.
\paragraph{Government Finance Departments.} 
Finance ministries manage government fiscal policy and economic regulation. They produce budget statements, policy white papers, press releases, and financial analysis reports. Most documents are public, though some internal memos and briefings remain confidential.
\paragraph{Financial Technology Companies.} 
Financial Technology companies (FinTech) engage in financial services innovation through technology, including digital banking, AI agents, investment technologies, cryptocurrency exchanges, and others. They produce customer agreements, product documentation, and white papers. Some FinTechs generate regulatory filings and compliance reports. Customer-facing documents are typically public, while internal reports and transaction logs remain private.
\paragraph{Legal \& Compliance Bodies.} 
These entities ensure regulatory adherence and oversee legal aspects of finance. They generate compliance manuals and audit reports (i.e., Suspicious Activity Reports) and publish legal advisories. While many compliance documents remain internal, some client advisories and industry guidelines are publicly available.
\subsubsection{Where}\label{app:domain_where}
\textit{"Where was the data generated/annotated?"}
In finance, textual data arises from multiple channels. Corporate disclosures are uploaded to regulatory databases (\eg, the SEC's Electronic Data Gathering, Analysis, and Retrieval (EDGAR)), press releases appear on news-wires or company websites, and social media data is generated globally. Annotation can be handled by specialized providers (\eg, rating agencies for risk labeling) or crowd-sourced platforms. Consequently, the “where” dimension includes the physical location of data creators or annotators and the digital repositories hosting the final datasets (\eg, regulatory websites, aggregator platforms, or data brokers).
\subsubsection{When}\label{sec:taxonomy:when}
\textit{"When was the data generated/annotated?"}
Finance is \textbf{time-sensitive}. Data from an older annual report (\eg, 2010) may be of historical research value, while a live earnings call is relevant to immediate trading decisions. Datasets could be divided into further categories \emph{historical}, \emph{recent}, or \emph{live-streaming}. The time also affects legal obligations (\eg, updated regulations), context relevance (macroeconomic conditions), and any potential dataset drift over time (\eg, new financial terminology, products, services).
\subsubsection{Why}\label{sec:taxonomy:why}
\textit{"Why would the data/annotations be used?"}
In finance, the motivations range from legal compliance (meeting regulatory disclosure requirements) to investor relations (transparency for shareholders) or internal risk management (spotting financial misconduct). Data often enables specific downstream applications—like building credit-scoring models or automating customer support. Understanding “why” data is created or used helps identify nuances in the data (\eg, self-reported vs. legally mandated) and the real-world implications for any NLP-driven downstream uses. We consider the real-world uses of benchmark datasets during categorization or metric selections. For example, data sets related to anti-money laundering focus on text classification to detect fraud and might prioritize recall to catch potential wrongdoing. In contrast, a financial analyst focuses on text classification for document classification.
\subsubsection{How}\label{app:domain_how}
\textit{"How was the data generated/annotated?"}
Financial data generation spans official reporting (formal documentation mandated by regulations) and user-generated content (social media, customer chats). Annotation might be done by subject matter experts (\eg, compliance officers labeling risk factors), professional analysts (\eg, rating agencies), crowd workers (\eg, annotator labeling), or machines (\eg, AI labeling services). The expertise needed often correlates with the data’s complexity—highly technical documents (\eg, derivative contracts) demand specialized annotators to ensure label accuracy. Annotations may be partially or fully automated, leveraging pattern-matching or prior language models to reduce costs.
\subsection{Language}\label{app:language}
\textit{"Language used for data/annotations?"}
Currently, \papertitle focuses on \textbf{English}, reflecting its widespread use in global financial markets and regulatory documents. However, finance also includes other major world languages for company disclosures, investor communications, and cross-border transactions. Future expansions may incorporate multilingual corpora to reflect cross-national markets better. For now, we emphasize that language coverage remains incomplete and is a major area for community-driven growth.

\section{Framework}\label{app:framework}
\para{Python Package.} We provide \papertitle as an open-source Python package under a Creative Commons Non-Commercial 4.0 License, offering the research community a \textit{generalizable} framework for reliable and reproducible evaluation of LMs on core NLP tasks for finance. \papertitle standardizes all steps of the evaluation process \dash downloading datasets, setting prompt templates, and computing metrics \dash such that researchers can fairly compare LMs on core NLP tasks across any selected scenario. Our software addresses prior issues of uncoordinated benchmarking by (1) making \emph{all} code, data, and results publicly available, (2) enforcing uniform data-loading pipelines, and (3) logging all inference parameters (e.g., temperature, context window) for transparency. We believe \papertitle will encourage more comprehensive study of new tasks, deeper error analysis, and rapid benchmarking of new models after release. We build our evaluation framework using LiteLLM, which acts as our “universal gateway” to bridge across any local inference engines or cloud API endpoint. This ensures identical prompting and evaluation logic for all models, regardless of whether the model is closed-source or open-weight.\\
\paragraph{Transparency and Reproducibility.}
Throughout, \papertitle stores complete metadata for every submission including model version, parameter count, datetime stamps, dataset versioning tags, evaluation settings, prompt templates, decoding parameters, and more. All final results (raw completions, logs, metrics) are compiled and serialized for secondary analysis and auditing. We aim to make \papertitle a trustworthy and collaborative anchor for ongoing financial LM research and take all steps needed to ensure the authenticity of all data used.

\section{Datasets}\label{app:datasets}
\para{Dataset Repository.} \papertitle also hosts a centralized repository of all benchmark datasets on HuggingFace \texttt{dataset} objects for consistent and immediate use by the community. We make these datasets available to users \textbf{\textit{only with the permission of the original authors}}. \papertitle boosts adoption by both academic and industry users by streamlines the evaluation process and (1) guaranteeing all evaluations use standardized formatting, (2) verifying correct annotation labels and dataset splits, and (3) facilitating future expansions by our community  (e.g., new language coverage, updates to annotations, data de-duplication).\\
\subsection{Selection Criteria}
\para{Domain}: We require that a \textit{majority} of the dataset’s content be directly relevant to finance (e.g., investor filings, policy statements). Datasets that are only tangentially financial (e.g., general news with minor finance topics) are excluded.\\
\para{Purpose}: We do not include massive corpora intended purely for model \textit{pre-training} or fine-tuning. Instead, we focus on evaluating zero/few-shot performance of foundation LMs.\\
\para{Task Substance}: The dataset should exercise real finance knowledge or language capabilities (e.g., extracting risk factors, classifying research reports). Overly trivial tasks or single-label corpora are discouraged.\\
\para{Difficulty}: The dataset should not be trivial for state-of-the-art LMs, yet solvable by domain experts. This ensures the benchmark is challenging enough to reveal meaningful differences in model performance.\\
\para{Simplicity}: Where possible, tasks should be feed-forward (one input → one output) and not rely on elaborate prompt engineering. We want to measure foundational LM performance rather than specialized engineering hacks.\\
\para{License and Attribution}: Any dataset in \papertitle must allow open research use and provide attribution for original data authors.\\
\para{Fairness and Quality.} We require transparent sourcing (first-party or third-party) and minimal risk of label corruption or poor annotation. We strongly prefer tasks built on \textit{novel} data or curated expansions of existing public data to reduce the risk of model contamination.\\
\para{Bounded Complexity.} We target tasks suitable for foundational LMs in zero-shot settings rather than massive pre-training sets. Long or multi-document tasks must still fit practical LM context windows. For specialized tasks (e.g., advanced numeric forecasting from documents), we will extend our work in the future.\\
\subsection{Frontier Scenarios and Future Additions}
We identify multiple \textbf{\textit{frontier scenarios}}—reasoning-based tasks (mathematical or causal), decision-making (market forecasts), advanced knowledge (fact completion, cross-lingual QA), and more \citepHELM. These go beyond standard NLP tasks and often demand specialized labeling or multi-modal input. Our plan is to collaborate with domain experts and the broader community to gradually incorporate these frontiers into \papertitle.
\subsection{Data Quality Assurance}
\para{Data Integrity.} We conducted comprehensive validation to ensure that all datasets used in \papertitle were of acceptable quality for use. Before including a dataset, we conduct manual or semi-automated checks for label mismatch, duplicate entries, and incomplete annotations. If the dataset is well-documented and widely cited as reliable, we fast-track its inclusion.\\
\para{Community Collaboration.} We invite researchers to submit new datasets or highlight issues in existing ones. Our open GitHub issue tracker logs reported label noise, mismatch between dataset documentation and raw text, or potential duplication with a model’s training set. Our philosophy is that the best finance LM benchmark emerges from open-source commmunities and iterative improvement.\\
\para{Contamination Risks} Because finance data may appear in large pre-training corpora, we encourage dataset creators to embed “salted” verifiers (hash tokens). \papertitle aims to mitigate unintentional memorization or partial overlap in training data by carefully tracking dataset versions and urging the community to keep \textit{private} test splits off the open web.\\
\para{Datasets Excluded}
We identified concerns regarding certain datasets during our survey. For these reason we exclude datasets which are being flagged as concern by others. Label quality is a major factor in the selection of our datasets. We choose datasets where the quality of the datasets has not been noted by the community to have issues. datasets like the CRA NER dataset \cite{Alvarado2015-oe} has been noted by others \cite{Wu2023-ph,Wu2024-df,Lu2025-iq} as having quality issues with labels due to using a limited selection of only four entity types. Using only four entity types leads to a severely skewed distributions of entity types due to the limited data.\\
The appropriate use of datasets is important. we exclude datasets that focus on evaluating tabular time series data using a standard language model. there is reasons to believe and show interest in transformers and decoders as symbolic reasoners over time series numerical data, but language models are not trained for time series forecasting. As others have noted \cite{Wu2024-df} this type of data and task tend to be ineffective and not useful for understanding the capability of a language model to generate a forecast.\\
In addition we also exclude datasets that are (i) purely tabular/time-series data that lacks semantic meaning or human-readable text, (ii) proprietary or undisclosed corpora that are not shared publicly or verified, (iii) modified subsets of widely used corpora, if they do not offer new annotations or insights.
\subsection{Datasets}
\paragraph{Question Answering.}
\begin{itemize}
    \item 
\textbf{FinQA} \cite{Chen2021-hr} is a large-scale dataset designed for numerical reasoning over financial data, consisting of 8,281 question-answer pairs derived from financial reports authored by experts. The dataset addresses the complexity of analyzing financial statements, which requires both deep understanding and intricate numerical reasoning. Unlike general QA tasks, FinQA focuses on questions that demand the interpretation of financial data and multi-step reasoning to reach an answer. The dataset is fully annotated with reasoning programs to ensure explainability, making it a valuable resource for advancing research in automated financial analysis. For evaluation, we prompted the language models to output the answer of each question. The FinQA dataset is licensed under the Creative Commons Attribution-NonCommercial 4.0 International \textbf{(CC BY-NC 4.0) license.}

The Zero-Shot prompt used for FinQA is given in Figure \ref{fig:finqa_prompt}.

\begin{figure}[h!]
    \centering
    \footnotesize
    \begin{tcolorbox}
    [colback=gray!5,colframe=black!75,title= FinQA Zero-Shot Prompt]
        \small
\begin{verbatim}
"""
Discard all the previous instructions. 
Behave like you are a financial expert 
in question answering. Your task is to 
answer a financial question based on 
the  provided context.\n\n 
The context: {document}. Repeat your 
final answer at the end of your response.
"""
\end{verbatim}
    \end{tcolorbox}
    \caption{Zero-shot prompt used for FinQA.}
    \label{fig:finqa_prompt}
\end{figure}
    \item \textbf{ConvFinQA (CFQA)}\cite{Chen2022-ae} multi-turn question answering is a large-scale dataset designed to explore the chain of numerical reasoning in conversational question-answering within the financial domain. It consists of 3,892 conversations and 14,115 questions, where the conversations are split between 2,715 simple and 1,177 hybrid conversations. ConvFinQA focuses on modeling complex, long-range numerical reasoning paths found in real-world financial dialogues. The dataset is a response to the growing need to study complex reasoning beyond pattern matching, and it includes experiments with neural symbolic and prompting-based methods to analyze reasoning mechanisms. This resource pushes the boundaries of research on numerical reasoning and conversational question-answering in finance. For evaluation, we prompted the language models to answer the question given context from a previous question and answer. The ConvFinQA dataset is released under the \textbf{MIT License.}

The Zero-Shot prompt used for ConvFinQA is given in Figure \ref{fig:convfinqa_prompt}.

\begin{figure}[h!]
    \centering
    \footnotesize
    \begin{tcolorbox}
    [colback=gray!5,colframe=black!75,title= ConvFinQA Zero-Shot Prompt]
        \small
\begin{verbatim}
"""
Discard all previous instructions. 
You are a financial expert 
specializing in answering questions.
The context provided includes a previous 
question and its answer, followed by a 
new question that you need to answer.
Focus on answering only the final 
question based on the entire provided 
context: {document}.
Answer the final question based on 
the context above. Repeat your final 
answer at the end of your response. 
"""
\end{verbatim}
    \end{tcolorbox}
    \caption{Zero-shot prompt used for ConvFinQA.}
    \label{fig:convfinqa_prompt}
\end{figure}

    \item \textbf{TAT-QA (TQA)} \cite{Zhu2021-ig} is a large-scale question-answering (QA) dataset designed for hybrid data sources, combining both tabular and textual content, particularly from financial reports. The dataset emphasizes numerical reasoning, requiring operations such as addition, subtraction, comparison, and more to infer answers from both tables and text. Extracted from real-world financial reports, TAT-QA challenges QA models to handle complex data formats, addressing a gap in existing research which often overlooks hybrid data. A new model, TAGOP, was introduced to tackle this challenge by extracting relevant cells and text spans for symbolic reasoning, achieving an F1 score of 58.0\%, though still falling short of expert human performance (90.8\%). TAT-QA provides a critical benchmark for advancing QA models in finance. For evaluation, we prompted the language models to output the answer given the context and question for each sample. The TAT-QA dataset is licensed under the Creative Commons Attribution 4.0 International \textbf{(CC BY 4.0) License.}

The Zero-Shot prompt used for TAT-QA is given in Figure \ref{fig:tatqa_prompt}.

\begin{figure}[h!]
    \centering
    \footnotesize
    \begin{tcolorbox}
    [colback=gray!5,colframe=black!75,title= TAT-QA Zero-Shot Prompt]
        \small
\begin{verbatim}
"""
Discard all previous instructions. 
Behave like an expert in table-and-
text-based financial question answering.
Your task is to answer a question by 
extracting relevant information from both 
tables and text provided in the context. 
Ensure that you use both sources 
comprehensively to generate an accurate 
response. Repeat your final answer at the
end of your response.  \n\n{text}
"""
\end{verbatim}
    \end{tcolorbox}
    \caption{Zero-shot prompt used for TAT-QA.}
    \label{fig:tatqa_prompt}
\end{figure}

\end{itemize}
\paragraph{Text Summarization.}
\begin{itemize}
    \item \textbf{ECTSum} \cite{Mukherjee2022-cj} is designed for bullet-point summarization of long earnings call transcripts (ECTs) in the financial domain. It consists of 2,425 document-summary pairs, with the transcripts sourced from publicly traded companies' earnings calls between January 2019 and April 2022. Each transcript is a lengthy, unstructured document, and the summaries are concise, telegram-style bullet points extracted from Reuters articles. These summaries focus on key financial metrics such as earnings, sales, and trends discussed during the calls. ECTSum addresses the challenge of summarizing complex financial data into short, meaningful summaries, making it a valuable benchmark for evaluating summarization models, particularly in the context of financial reporting. For evaluation, we prompted the language models to output a bullet point summary from each sample, and compared that summary to the ground truth summary with BERTScore. The ECTSum dataset is released under the \textbf{GPL-3.0 license}.

The Zero-Shot prompt used for ECTSum is given in Figure \ref{fig:ectsum_prompt}.

\begin{figure}[h!]
    \centering
    \footnotesize
    \begin{tcolorbox}
    [colback=gray!5,colframe=black!75,title= ECTSum Zero-Shot Prompt]
        \small
\begin{verbatim}
"""
Discard all the previous instructions.
Behave like you are an expert at 
summarization tasks. Below an earnings call 
transcript of a Russell 3000 Index company
is provided. Perform extractive summarization 
followed by paraphrasing the transcript in 
bullet point format according to the
experts-written short telegram-style bullet 
point summaries derived from corresponding 
Reuters articles. The target length of
the summary should be at most 50 words. 
\n\n The document: {document}
"""
\end{verbatim}
    \end{tcolorbox}
    \caption{Zero-shot prompt used for ECTSum.}
    \label{fig:ectsum_prompt}
\end{figure}

    \item \textbf{EDTSum} \cite{Xie2024-pn} is a financial news summarization resource designed to evaluate the performance of large language models (LLMs) in generating concise and informative summaries. It comprises 2,000 financial news articles, each paired with its headline serving as the ground-truth summary. These articles were manually selected and cleaned from the dataset introduced by  to ensure high-quality annotations. The original dataset \cite{Zhou2021-il} focuses on corporate event detection and text-based stock prediction, containing 9,721 news articles with token-level event labels and 303,893 first-hand news articles with minute-level timestamps and comprehensive stock price labels. For evaluation, we prompted the language models to output a summary given an article, and compared that summary to the ground truth summary with BERTScore. The EDTSum dataset provides a benchmark for financial text summarization. The EDTSum dataset is \textbf{publicly available}.

The Zero-Shot prompt used for EDTSum is given in Figure \ref{fig:edtsum_prompt}.

\begin{figure}[h!]
    \centering
    \footnotesize
    \begin{tcolorbox}
    [colback=gray!5,colframe=black!75,title= EDTSum Zero-Shot Prompt]
        \small
\begin{verbatim}
"""
Discard all the previous instructions. 
Behave like you are an expert at summarization 
tasks. You are given a text that consists 
of multiple sentences. Your task is to perform 
abstractive summarization on this text. 
Use your understanding of the content to 
express the main ideas and crucial details 
in a shorter, coherent, and natural sounding 
text. \nThe text:\n{document}.\nOutput your 
concise summary below. Try to keep your summary 
to one sentence and a maximum of 50 words, 
preferably around 25 words.
"""
\end{verbatim}
    \end{tcolorbox}
    \caption{Zero-shot prompt used for EDTSum.}
    \label{fig:edtsum_prompt}
\end{figure}

\end{itemize}
\paragraph{Information Retrieval.}
\begin{itemize}
    \item \textbf{FiNER-Open Research Dataset (FiNER-ORD)} \cite{Shah2023-hr} is a manually annotated dataset comprising 47,851 financial news articles (in English) collected from webz.io. Each article is a JSON document containing metadata such as the source, publication date, author, and title. A subset of 220 randomly sampled documents was manually annotated, with 201 remaining after filtering out empty articles. The dataset was manually labeled using Doccano, an open-source annotation tool, with annotations for person (PER), location (LOC), and organization (ORG) entities. This annotated dataset benchmarks model performance for financial named entity recognition. Further annotation guidelines are available in the dataset's documentation. We did not perform any additional preprocessing; the test set of the dataset is used in its original publicly available form. The main metric used for evaluations of the models for the FiNER-ORD dataset is Macro F1. The FiNER-Open Research Dataset (FiNER-ORD) is available under the Creative Commons Attribution-NonCommercial 4.0 International \textbf{(CC BY-NC 4.0) license }.

The Zero-Shot prompt used for FiNER is given in Figure \ref{fig:finer_prompt}.

\begin{figure}[h!]
    \centering
    \footnotesize
    \begin{tcolorbox}
    [colback=gray!5,colframe=black!75,title= FiNER Zero-Shot Prompt]
        \small
\begin{verbatim}
"""
Discard all the previous instructions. 
Behave like you are an expert named entity
identifier. Below a sentence is tokenized 
and each list item contains a word token 
from the sentence. Identify ‘Person’, 
‘Location’, and ‘Organisation’ from them and 
label them. If the entity is multi token use 
post-fix_B for the first label and _I for 
the remaining token labels for that 
particular entity. The start of 
the separate entity should always use _B 
post-fix for the label. If the token doesn’t 
fit in any of those three categories or 
is not a named entity label it ‘Other’. 
Do not combine words yourself. Use a colon t
o separate token and label.
So the format should be token:label. 
\n\n + {sentence}
"""
\end{verbatim}
    \end{tcolorbox}
    \caption{Zero-shot prompt used for FiNER.}
    \label{fig:finer_prompt}
\end{figure}

    \item \textbf{FinEntity (FE)} \cite{Tang2023-sm} is an entity-level sentiment classification dataset designed for financial news analysis. It contains 979 financial news paragraphs, featuring 2,131 manually-annotated financial entities classified into positive, negative, and neutral sentiment categories. The dataset was sourced from Refinitiv Reuters Database, ensuring high-quality financial news coverage. Data collection focused on financial entities such as companies, organizations, and asset classes, excluding persons, locations, and events. The dataset employs a BILOU labeling scheme for entity tagging and sentiment classification. Fine-tuned BERT and FinBERT models significantly outperform ChatGPT in this task. Additionally, the FinEntity dataset has been applied to cryptocurrency news (15,290 articles from May 2022 to February 2023), demonstrating stronger correlations between entity-level sentiment and cryptocurrency prices compared to traditional sequence-level sentiment models. Only the test set from the FinEntity dataset is used, with no additional preprocessing applied. However, we do not consider the start and end boundary tags during evaluation; they are therefore excluded from the assessment. The FinEntity dataset is licensed under the Open Data Commons Attribution License \textbf{(ODC-BY) license.}

Previous work on FinEntity, such as \cite{Xing2025-qo},
focuses on sentiment classification and does not account for entity extraction in the same manner. Specifically, prior approaches often introduce random insertions to handle unclear or irrelevant outputs, which is not applicable to our evaluation setting where exact entity matching is also considered.

The FinEntity task involves entity extraction and sentiment classification. For our evaluations, span boundary detection is not considered. This evaluation metric treats outputs as sets rather than enforcing exact span alignment.

\textbf {Entity-Based Comparison}

Given the predicted and ground-truth entity sets:

$$
E_p = \{e_{p1}, e_{p2}, \ldots, e_{p_{N_p}}\}
$$

$$
E_t = \{e_{t1}, e_{t2}, \ldots, e_{t_{N_t}}\}.
$$

Each entity \( e \) is represented as:

\[
e = (\text{value},\, \text{tag},\, \text{label}).
\]

An entity in the predicted set is considered a match if it exactly equals any ground-truth entity:

\[
M = \{ e \in E_p : e \in E_t \}.
\]

\textbf{Proposed Evaluation Metric}

We compute:

$$
P = \frac{|M|}{|E_p|}, \\
$$

$$
R = \frac{|M|}{|E_t|}, \\
$$

$$
F1 = \frac{2 P R}{P + R}, \\
$$

$$
\text{Accuracy} = \frac{|M|}{|E_t|}.
$$

Since our evaluation is entity-level, accuracy is equivalent to recall. Unlike prior work that enforces strict length matching, we adopt a more flexible metric to better align with the nature of LLM outputs. This allows for partial credit and avoids assigning a score of zero when predictions differ in length from the ground truth.

The Zero-Shot prompt used for FinEntity is given in Figure \ref{fig:finentity_prompt}.

\begin{figure}[h!]
    \centering
    \footnotesize
    \begin{tcolorbox}
    [colback=gray!5,colframe=black!75,title= FinEntity Zero-Shot Prompt]
        \small
\begin{verbatim}
"""
Discard all the previous instructions. 
Behave like you are an expert entity 
recognizer and sentiment classifier. 
Identify the entities which are companies 
or organizations from the following 
content and classify the sentiment of the 
corresponding entities into ‘Neutral’ 
‘Positive’ or ‘Negative’ classes. 
Considering every paragraph as a String in 
Python, provide the entities with the 
start and end index to mark the 
boundaries of it including spaces and 
punctuation using zero-based indexing. 
In the output,  Tag means sentiment; 
value means entity name. If no entity is 
found in the paragraph, the response 
should be empty. Only give the output, 
not python code. The output should be a 
list that looks like:
[{{'end': int,
'label': 'Neutral',
'start': int,
'tag': 'Neutral',
'value': str}},
{{'end': int, 'label': 'Neutral', 
'start': int, 'tag': 'Neutral', 
'value': str}}]
Do not repeat any JSON object in the list. 
Evey JSON object should be unique.
The paragraph: {paragraph}
"""
\end{verbatim}
    \end{tcolorbox}
    \caption{Zero-shot prompt used for FinEntity.}
    \label{fig:finentity_prompt}
\end{figure}

    \item \textbf{The Financial Numeric Extreme Labeling (FNXL) dataset} \cite{Sharma2023-ir} addresses the challenge of automating the annotation of numerals in financial statements with appropriate labels from a vast taxonomy. Sourced from the U.S. Securities and Exchange Commission's (SEC) publicly available annual 10-K reports from 2019 to 2021, the FNXL dataset comprises 79,088 sentences containing 142,922 annotated numerals, categorized under 2,794 distinct labels.

The FNXL task involves extracting numerical values associated with specific XBRL tags. Unlike traditional named entity recognition, this task requires set-based numerical comparison. Thus, we cannot use Entity F1 scores directly. 

Normalization is applied consistently across all datasets to reduce inconsistencies, including case standardization and whitespace stripping, but we do not explicitly define it per dataset.

\textbf{Set-Based Comparison and Partial Credit}

Each tag is associated with a set of numerical values, and we evaluate based on set overlap rather than exact string matching. Given the predicted and ground-truth mappings:

$
T_p = \{(t_p, S_p)\}
$

$
T_t = \{(t_t, S_t)\},
$

where \( S_p \) and \( S_t \) are sets of numerical values, we compute:

$M_t = S_p \cap S_t,$ \\
$TP = \sum_{t} |M_t|$, \\ 
$FP = \sum_{t} |S_p - M_t|$ \\ 
$FN = \sum_{t} |S_t - M_t|.$

The total actual and predicted values are given by:

$
\text{Total}_{\text{actual}} =  \sum_{t} |S_t|
$

$
\text{Total}_{\text{predicted}} = \sum_{t} |S_p|.
$

\textbf{Evaluation Metrics}

We compute precision, recall, and F1 score using standard formulae.

Additionally, we define a Jaccard-inspired accuracy measure:

$
\text{Accuracy} = \frac{TP}{\text{Total}_{\text{actual}} + \text{Total}_{\text{predicted}} - TP}
$

This evaluation metric allows for partial credit by considering numerical overlaps instead of enforcing exact matches, which is crucial given the nature of LLM predictions.

The Zero-Shot prompt used for FNXL is given in Figure \ref{fig:fnxl_prompt}.

\begin{figure}[h!]
    \centering
    \footnotesize
    \begin{tcolorbox}
    [colback=gray!5,colframe=black!75,title= FNXL Zero-Shot Prompt]
        \small
\begin{verbatim}
"""
Discard all the previous instructions.
Behave like you are an SEC reporting expert. 
Given a sentence from a financial filing, 
do the following two things:
1) Identify every numeral in the sentence.
2) For each numeral, assign the most 
appropriate US-GAAP XBRL tag based on context. 
If no tag is appropriate, label it as "other".

Return only valid JSON in this format:
```json
{
"12.0": "us-gaap:Revenue",
"9.5": "us-gaap:SomeExpense",
"100.0": "other"
}```
The sentence is: {sentence}
"""
\end{verbatim}
    \end{tcolorbox}
    \caption{Zero-shot prompt used for FNXL.}
    \label{fig:fnxl_prompt}
\end{figure}

    \item \textbf{FinRED (FR)} \cite{Sharma2022-dt} dataset is a specialized relation extraction dataset tailored to the financial domain, created to address the gap where existing models trained on general datasets fail to transfer effectively to financial contexts. It comprises data curated from financial news and earnings call transcripts, with financial relations mapped using a distance supervision method based on Wikidata triplets. To ensure robust evaluation, the test data is manually annotated. The dataset provides a benchmark for evaluating relation extraction models, revealing a significant performance drop when applied to financial relations, highlighting the need for more advanced models in this domain. For evaluation, we prompted the language models to output the relation of an entity pair given the list of possible relations, the entity, and the statement. The FinRED dataset is released under the Creative Commons Attribution 4.0 International \textbf{(CC BY 4.0) license.}

The Zero-Shot prompt used for FinRED is given in Figure \ref{fig:finred_prompt}.

\begin{figure}[h!]
    \centering
    \footnotesize
    \begin{tcolorbox}
    [colback=gray!5,colframe=black!75,title= FinRED Zero-Shot Prompt]
        \small
\begin{verbatim}
"""
Classify what relationship {entity2} (the head) 
has to {entity1} (the tail) within the 
following sentence: "{sentence}" 
The relationship should match one of the 
following categories, where the relationship 
is what the head entity is to the tail 
entity: {", ".join(possible_relationships)}. 
You must output one, and only one, 
relationship out of the previous list that 
connects the head entity {entity2} to the 
tail entity {entity1}. Find what relationship 
best fits {entity2} 'RELATIONSHIP' {entity1} 
for this sentence.
"""
\end{verbatim}
    \end{tcolorbox}
    \caption{Zero-shot prompt used for FinRED.}
    \label{fig:finred_prompt}
\end{figure}

    \item \textbf{REFinD (RD)} \cite{Kaur2023-we} is a specialized relation extraction dataset created to address the unique challenges of extracting relationships between entity pairs from financial texts. With approximately 29,000 annotated instances and 22 distinct relations across 8 types of entity pairs, it stands out as the largest-scale dataset of its kind, specifically generated from financial documents, including Securities and Exchange Commission (SEC) filings. This dataset aims to fill the gap left by existing relation extraction datasets, which are predominantly compiled from general sources like Wikipedia or news articles. For evaluation, we prompted the language models to output the relation of an entity pair given the sentence and entity pairs. We did not count no relationship entity pairs. The REFinD dataset is licensed under the Creative Commons Attribution-NonCommercial 4.0 International \textbf{(CC BY-NC 4.0)  License}.

The Zero-Shot prompt used for ReFinD is given in Figure \ref{fig:refind_prompt}.

\begin{figure}[h!]
    \centering
    \footnotesize
    \begin{tcolorbox}
    [colback=gray!5,colframe=black!75,title= ReFinD Zero-Shot Prompt]
        \small
\begin{verbatim}
"""
Classify the following relationship between
ENT1 (the subject) and ENT2 (the object). 
The entities are marked by being enclosed 
in [ENT1] and [/EN1] and [ENT2] and [/ENT2]
respectively. The subject entity will
either be a person (PER) or an organization 
(ORG). The possible relationships are as 
follows, with the subject listed first and 
object listed second:
PERSON/TITLE - person subject, title object,
relation title 
PERSON/GOV_AGY - person subject, 
government agency object, relation member_of
PERSON/UNIV - person subject, 
university object, relation employee_of, 
member_of, attended PERSON/ORG - person 
subject, organization object, 
relation employee_of,
member_of, founder_of ORG/DATE - organization
subject, date object,
relation formed_on, 
acquired_on ORG/MONEY - organization subject,
money object, relation revenue_of, profit_of,
loss_of, cost_of ORG/GPE - 
organization subject,
geopolitical entity object, relation
headquartered_in, operations_in, formed_in
ORG/ORG - organization subject, 
organization object, relation shares_of,
subsidiary_of, acquired_by, agreement_with
Text about entities: {entities}
"""
\end{verbatim}
    \end{tcolorbox}
    \caption{Zero-shot prompt used for ReFinD.}
    \label{fig:refind_prompt}
\end{figure}

\end{itemize}
\paragraph{Sentiment Analysis.}
\begin{itemize}
    \item \textbf{FiQA} \cite{Maia2018-hg} has two sub tasks. \textbf{\textit{FiQA Task 1}} focuses on aspect-based financial sentiment analysis. Given a financial text, such as microblog posts or news headlines, systems are tasked with identifying the specific target aspects mentioned and predicting their corresponding sentiment scores on a continuous scale from -1 (negative) to 1 (positive). The challenge involves accurately linking financial entities or topics to the appropriate sentiment, such as distinguishing between corporate strategy decisions of companies. For evaluation, systems are measured on their ability to correctly classify aspects, attach sentiment to those aspects, and predict sentiment with metrics like precision, recall, F1-score, and regression-based measures (MSE and R-squared). For evaluation, we prompted the language models to output a sentiment score given each sample financial text.
\textbf{\textit{FiQA Task 2}} addresses opinion-based question answering (QA) over financial data, where systems must answer natural language questions by retrieving relevant financial opinions and facts from a knowledge base of structured and unstructured documents (such as reports, news, and microblogs). This task requires systems to either rank relevant documents from the knowledge base or generate answers directly. Opinion-based questions require identifying entities, aspects, sentiment, and opinion holders, with performance evaluated on metrics like F-score, Normalized Discounted Cumulative Gain (NDCG), and Mean Reciprocal Rank (MRR). The QA test collection includes diverse sources like StackExchange, Reddit, and StockTwits, focusing on ranking and answering accuracy.

The Zero-Shot prompt used for FiQA is given in Figure \ref{fig:fiqa_prompt}.

\begin{figure}[h!]
    \centering
    \footnotesize
    \begin{tcolorbox}
    [colback=gray!5,colframe=black!75,title= FiQA Zero-Shot Prompt]
        \small
\begin{verbatim}
"""
You are a financial sentiment analysis expert. 
Analyze the provided sentence, 
identify relevant target aspects 
(such as companies, products, or strategies), 
and assign a sentiment score for each target. 
The sentiment score should be between -1 
(highly negative) and 1 (highly positive), 
using up to three decimal places to capture 
nuances in sentiment. 
Financial sentence: {sentence}
"""
\end{verbatim}
    \end{tcolorbox}
    \caption{Zero-shot prompt used for FiQA.}
    \label{fig:fiqa_prompt}
\end{figure}

    \item \textbf{Financial Phrase Bank (FPB)} \cite{Malo2013-el}, is a dataset for sentiment analysis in financial news. It contains 4,840 sentences sourced from English-language financial news articles, categorized by sentiment as positive, negative, or neutral. Each sentence reflects the sentiment an investor might perceive from the news with respect to its influence on stock prices. The dataset is annotated by a group of 16 annotators with a background in finance, using a majority vote approach. It is available in four different configurations based on annotator agreement levels (50\%, 66\%, 75\%, and 100\%). FPB is used as resource for financial sentiment analysis, especially for training and benchmarking models in the financial domain. For evaluation, we prompted the language models to output the sentiment of each sample, given the choices positive, negative, and neutral. The Financial Phrase Bank (FPB) dataset is licensed under the Creative Commons Attribution-NonCommercial-ShareAlike 3.0 Unported \textbf{(CC BY-NC-SA 3.0) License}.

The Zero-Shot prompt used for FPB is given in Figure \ref{fig:fpb_prompt}.

\begin{figure}[h!]
    \centering
    \footnotesize
    \begin{tcolorbox}
    [colback=gray!5,colframe=black!75,title= FPB Zero-Shot Prompt]
        \small
\begin{verbatim}
"""
Discard all the previous instructions. 
Behave like you are an expert sentence 
classifier. Classify the following sentence 
into ‘NEGATIVE’, ‘POSITIVE’, or ‘NEUTRAL’ 
class. Label ‘NEGATIVE’ if it is corresponding 
to negative sentiment, ‘POSITIVE’ if it is 
corresponding to positive sentiment, or 
‘NEUTRAL’ if the sentiment is neutral. 
Provide the label in the first line and 
provide a short explanation in the second line. 
This is the sentence: {sentence}
"""
\end{verbatim}
    \end{tcolorbox}
    \caption{Zero-shot prompt used for FPB.}
    \label{fig:fpb_prompt}
\end{figure}

    \item \textbf{SubjECTive-QA (SQA)} \cite{Pardawala2024-lj} is a manually-annotated dataset focusing on subjectivity and soft misinformation in Earnings Call Transcripts (ECTs), specifically in their long-form QA sessions. It includes 49,446 annotations across 2,747 QA pairs from 120 ECTs spanning 2007 to 2021. Each QA pair is labeled on six subjectivity features: Assertive, Cautious, Optimistic, Specific, Clear, and Relevant. The dataset was benchmarked using RoBERTa-base and Llama-3-70b-Chat, showing varying performance based on feature subjectivity. Additionally, cross-domain evaluation on White House Press Briefings demonstrated its broader applicability. The SubjECTive-QA dataset is licensed under the Creative Commons Attribution 4.0 International \textbf{(CC BY 4.0) License}.

The Zero-Shot prompt used for SubjECTiveQA is given in Figure \ref{fig:subjectiveqa_prompt}.

\begin{figure}[h!]
    \centering
    \footnotesize
    \begin{tcolorbox}
    [colback=gray!5,colframe=black!75,title= SubjECTiveQA Zero-Shot Prompt]
        \small
\begin{verbatim}
"""
Discard all the previous instructions. 
Given the following feature:
{feature} and its corresponding definition:
{definition}\n. Give the answer a 
rating of:\n
2: If the answer positively 
demonstrates the chosen feature, with 
regards to the question.\n
1: If there is no evident/neutral 
correlation between the question and the 
answer for the feature.\n
0: If the answer negatively correlates 
to the question on the chosen feature.\n
Provide the rating only. No explanations. 
This is the question: {question} and this 
is the answer: {answer}.
"""
\end{verbatim}
    \end{tcolorbox}
    \caption{Zero-shot prompt used for SubjECTiveQA.}
    \label{fig:subjectiveqa_prompt}
\end{figure}

    \item FiNER falls under Information Retrieval and Sentiment Analysis, see Information Retrieval section for the dataset information.
\end{itemize}
\paragraph{Text Classification.}
\begin{itemize}
    \item \textbf{Banking77 (B77)} \cite{Casanueva2020-oa} is a fine-grained dataset designed for intent detection within the banking domain. It comprises 13,083 customer service queries annotated with 77 unique intents, such as card\_arrival and lost\_or\_stolen\_card. The dataset focuses on single-domain intent classification, providing a granular view of customer queries in the banking sector. With 10,003 training and 3,080 test examples, Banking77 offers a valuable resource for evaluating machine learning models in intent detection. The dataset has been curated to fill the gap in existing intent detection datasets, which often feature fewer intents or cover multiple domains without the depth offered here. For evaluation, we prompted the language models to identify each sample's intent from the list of intents. The Banking77 dataset is publicly available under the \textbf{MIT License}.\citet{Ying2022-fo} investigates potential labeling errors in Banking77, but further studies are required before a determination can be made.

The Zero-Shot prompt used for Banking77 is given in Figure \ref{fig:banking77_prompt}.

\begin{figure}[h!]
    \centering
    \footnotesize
    \begin{tcolorbox}
    [colback=gray!5,colframe=black!75,title= Banking77 Zero-Shot Prompt]
        \small
\begin{verbatim}
"""
Discard all the previous instructions. 
Behave like you are an expert at
fine-grained single-domain intent detection. 
From the following list: 
["activate_my_card", "age_limit", ...,
"wrong_exchange_rate_for_cash_withdrawal"],
identify which category the following sentence
belongs to. The sentence: {sentence}
"""
\end{verbatim}
    \end{tcolorbox}
    \caption{Zero-shot prompt used for Banking77.}
    \label{fig:banking77_prompt}
\end{figure}

    \item \textbf{FinBench (FB)} \cite{Yin2023-nf} is a dataset designed to evaluate the performance of machine learning models using both tabular data and profile text inputs, specifically within the context of financial risk prediction. The FinBench dataset consists of approximately 333,000 labeled instances, covering three primary financial risks: default, fraud, and churn. Each instance is labeled as "high risk" or "low risk". The time frame of data collection varies by dataset. The dataset accompanies FinPT, an approach that leverages Profile Tuning using foundation LMs. The core task is to transform tabular data into natural-language customer profiles via LMs for enhanced prediction accuracy. For evaluation, we prompted the language models to output high risk or low risk given the profile text. This benchmark falls under financial risk prediction. The FinBench dataset is licensed under the Creative Commons Attribution-NonCommercial 4.0 International \textbf{(CC BY-NC 4.0) license}.

The Zero-Shot prompt used for FinBench is given in Figure \ref{fig:finbench_prompt}.

\begin{figure}[h!]
    \centering
    \footnotesize
    \begin{tcolorbox}
    [colback=gray!5,colframe=black!75,title= FinBench Zero-Shot Prompt]
        \small
\begin{verbatim}
"""
Discard all the previous instructions. 
Behave like you are an expert risk assessor.
Classify the following individual as either 
‘LOW RISK’ or ‘HIGH RISK’ for approving a loan 
for. Categorize the person as ‘HIGH RISK’ if 
their profile indicates that they will likely 
default on the loan and not pay it back, and 
‘LOW RISK’ if it is unlikely that they will 
fail to pay the loan back in full.
Provide the label in the first line and 
provide a short explanation in the second 
line. Explain how you came to your 
classification decision and output the label 
that you chose. Do not write any code, 
simply think and provide your decision.
Here is the information about the person:
\nProfile data: {profile}\nPredict the risk 
category of this person:
"""
\end{verbatim}
    \end{tcolorbox}
    \caption{Zero-shot prompt used for FinBench.}
    \label{fig:finbench_prompt}
\end{figure}

    \item \textbf{Numerical Claim Detection Dataset (NC)} \cite{Shah2024-cr} is an expert-annotated dataset designed for detecting fine-grained investor claims within financial narratives, with a focus on the role of numerals. The dataset was constructed by sampling and annotating financial-numeric sentences from a large collection of 87,536 analyst reports (2017–2020) and 1,085 earnings call transcripts (2017–2023). Specifically, 96 analyst reports (two per sector per year) were sampled, containing 2,681 unique financial-numeric sentences, alongside 12 randomly selected earnings call transcripts (two per year), contributing 498 additional financial-numeric sentences. Each sentence was manually labeled as either "In-claim" or "Out-of-claim" by two annotators with foundational expertise in finance, ensuring high-quality annotations. This dataset facilitates the study of numerical claim detection in financial discourse and serves as a resource for argument mining and investor sentiment analysis. For evaluation, we prompted the language models to output if each sample was in claim or out of claim. The Numerical Claim Detection dataset is licensed under the Creative Commons Attribution 4.0 International \textbf{(CC BY 4.0) license}.

The Zero-Shot prompt used for NumClaim is given in Figure \ref{fig:numclaim_prompt}.

\begin{figure}[h!]
    \centering
    \footnotesize
    \begin{tcolorbox}
    [colback=gray!5,colframe=black!75,title= NumClaim Zero-Shot Prompt]
        \small
\begin{verbatim}
"""
Discard all the previous instructions. 
Behave like you are an expert sentence 
sentiment classifier. Classify the following 
sentence into ‘INCLAIM’, or ‘OUTOFCLAIM’ class.
Label ‘INCLAIM’ if it consists of a claim and 
not just factual past or present information, 
or ‘OUTOFCLAIM’ if it has just factual past or
present information. 
Provide the label in the first line and 
provide a short explanation in the second 
line. The sentence:{sentence}
"""
\end{verbatim}
    \end{tcolorbox}
    \caption{Zero-shot prompt used for NumClaim.}
    \label{fig:numclaim_prompt}
\end{figure}

    \item News \textbf{Headline (HL)} Classification \cite{Sinha2021-ax} dataset consists of 11,412 human-annotated financial news headlines focused on commodities, particularly gold. The dataset spans a collection period from 2000 to 2019. It includes publication date, article URL, and the news headline itself, and binary indicators that capture key financial aspects, including whether the headline mentions a price, the direction of price movement, and references to past or future prices and news. This dataset is valuable for analyzing sentiment and market trends based on news articles, making it a useful resource for financial analysis, trading strategy development, and research in sentiment analysis within the financial domain. For evaluation, we prompted the language models to output answers to the 7 different questions given the sample headline, such as whether the headline contains a price. The News Headline Classification dataset is licensed under the Creative Commons Attribution-ShareAlike 3.0 \textbf{(CC BY-SA 3.0) license.}

The Zero-Shot prompt used for News Headlines is given in Figure \ref{fig:newsheadlines_prompt}.

\begin{figure}[h!]
    \centering
    \footnotesize
    \begin{tcolorbox}
    [colback=gray!5,colframe=black!75,title= News Headlines Zero-Shot Prompt]
        \small
\begin{verbatim}
"""
Discard all the previous instructions. 
Behave like you are an expert at analyzing 
headlines. Give a score of 0 for each of the
following attributes if the news 
headline does not contain the following 
information or 1 if it does.
Price or Not: Does the news item talk 
about price or not.
Direction Up: Does the news headline talk 
about price going up or not?
Direction Down: Does the news headline 
talk about price going down or not?
Direction Constant: Does the news headline talk 
about price remaining constant 
or not?
Past Price: Does the news headline talk 
about an event in the past?
Future Price: Does the news headline talk 
about an event in the future?
Past News: Does the news headline talk 
about a general event (apart from prices) 
in the past?
The news headline is: {sentence}
"""
\end{verbatim}
    \end{tcolorbox}
    \caption{Zero-shot prompt used for News Headlines.}
    \label{fig:newsheadlines_prompt}
\end{figure}

    \item \textbf{Federal Open Market Committee (FOMC)} \cite{Shah2023-bh} dataset is a large-scale, tokenized, and annotated dataset designed to analyze the impact of monetary policy announcements on financial markets. It comprises FOMC speeches, meeting minutes, and press conference transcripts collected from 1996 to 2022. The dataset introduces a novel task of hawkish-dovish classification, where the goal is to classify the stance of FOMC communications into hawkish (policy tightening), dovish (policy easing), or neutral categories. The dataset is accompanied by various metadata, including the speaker and publication date. It was curated using both rule-based methods and manual annotation, and it has been benchmarked using state-of-the-art pre-trained models like RoBERTa, BERT, and others. The dataset aims provides resource for understanding how FOMC communications influence financial markets, including stock and treasury yields. For evaluation, we prompted the language models to output the stance of each sample given the choices hawkish, dovish, and neutral. The Federal Open Market Committee (FOMC) dataset is publicly available under the Creative Commons Attribution-NonCommercial 4.0 International \textbf{(CC BY-NC 4.0) license.}

The Zero-Shot prompt used for FOMC is given in Figure \ref{fig:fomc_prompt}.

\begin{figure}[h!]
    \centering
    \footnotesize
    \begin{tcolorbox}
    [colback=gray!5,colframe=black!75,title= FOMC Zero-Shot Prompt]
        \small
\begin{verbatim}
"""
Discard all the previous instructions.
Behave like you are an expert 
sentence classifier. Classify the following 
sentence from FOMC  into ‘HAWKISH’, 
‘DOVISH’, or ‘NEUTRAL’ class. 
Label ‘HAWKISH’ if it is corresponding to 
tightening of the monetary policy,
‘DOVISH’ if it is corresponding to easing 
of the monetary policy, or ‘NEUTRAL’ if the 
stance is neutral. 
Provide the label in the first line and 
provide a short explanation in the 
second line. This is the sentence: {sentence}
"""
\end{verbatim}
    \end{tcolorbox}
    \caption{Zero-shot prompt used for FOMC.}
    \label{fig:fomc_prompt}
\end{figure}

\end{itemize}
\paragraph{Causal Analysis.}
\begin{itemize}
    \item \textbf{FinCausal-SC} \cite{Mariko2020-by} is a dataset for cause-effect analysis in financial news texts. It consists of 29,444 text sections (each containing up to three sentences), with 2,136 annotated as causal and accompanied by cause-effect spans. 
FinCausal focuses on two tasks: 

\textbf{(1) Causality Classification (CC).}
Determine if a given text section contains a causal relation. Each text section is labeled with 
Gold = 1 if a causal statement is present and 0 otherwise.

\textbf{(2) Causality Detection (CD).}
For those text sections identified as causal, the task is to extract the Cause and Effect spans. In total, there are 796 instances annotated for cause-effect extraction. These include both unicausal cases (with an average of  621.67 instances) and multicausal cases (with an average of 174.33 instances). This task challenges models to handle potentially complex causal chains, where one event can trigger multiple consequences or multiple factors can lead to a single outcome. 

FinCausal-SC pushes beyond simple keyword matching toward more nuanced and context-aware understanding of financial news articles. This dataset is published under the \textbf{CC0 License}.

The Zero-Shot prompt used for Causal Detection is given in Figure \ref{fig:causaldetection_prompt}.

\begin{figure}[h!]
    \centering
    \footnotesize
    \begin{tcolorbox}
    [colback=gray!5,colframe=black!75,title= Causal Detection Zero-Shot Prompt]
        \small
\begin{verbatim}
"""
You are an expert in detecting cause and effect
phrases in text.
You are given the following tokenized 
sentence. For each token, assign one of 
these labels:
- 'B-CAUSE': The first token of a cause phrase.
- 'I-CAUSE': A token inside a cause phrase, 
but not the first token.
- 'B-EFFECT': The first token of an 
effect phrase.
- 'I-EFFECT': A token inside an effect phrase, 
but not the first token.
- 'O': A token that is neither part of a cause 
nor an effect phrase.

Return only the list of labels in the same 
order as the tokens, without additional 
commentary or repeating the tokens themselves.

Tokens: {", ".join(tokens)}
"""
\end{verbatim}
    \end{tcolorbox}
    \caption{Zero-shot prompt used for Causal Detection.}
    \label{fig:causaldetection_prompt}
\end{figure}

\end{itemize}
\section{Models} \label{app:models}
In this section we detail the various models evaluated on the benchmarks along with the associated evaluation costs. The details of the models are displayed in Table~\ref{tab:model-detail}.
\begin{table*}[t]
\centering
\resizebox{0.95\textwidth}{!}{%
\renewcommand{\arraystretch}{1.1}
\begin{tabular}{llllllcc}
\toprule
Model & Organization & Provider & Size & Notes & Source & \makecell{Input Token Cost\\(\$USD / 1M Tokens)} & \makecell{Output Token Cost\\(\$USD / 1M Tokens)} \\
\midrule
GPT-4o & OpenAI & OpenAI & -- & -- & \texttt{openai/gpt-4o-2024-08-06} & 2.5 & 10 \\
OpenAI o1-mini & OpenAI & OpenAI & -- & -- & \texttt{openai/o1-mini} & 1.1 & 4.4 \\
\hdashline
Claude-3.5-Sonnet& Anthropic & Anthropic & -- & -- & \texttt{anthropic/claude-3-5-sonnet-20240620} & 3 & 15 \\
Claude-3-Haiku & Anthropic & Anthropic & -- & -- & \texttt{anthropic/claude-3-haiku-20240307} & 0.25 & 1.25 \\
\hdashline
Gemini-1.5-Pro & Google & Google & -- & -- & \texttt{gemini/gemini-1.5-pro} & 1.25 & 5.0 \\
\midrule
Llama-3-70B & Meta & Together AI& 70B & Dense & \texttt{meta-llama/Llama-3-70b-chat-hf} & 0.90 & 0.90 \\
Llama-3-8B & Meta & Together AI & 8B & Dense & \texttt{meta-llama/Llama-3-8b-chat-hf} & 0.20 & 0.20 \\
Llama-2-13B & Meta & Together AI & 13B & Dense & \texttt{meta-llama/Llama-2-13b-chat-hf} & 0.30 & 0.30 \\
\hdashline
DBRX & Databricks & Together AI & 132B & MoE & \texttt{databricks/dbrx-instruct} & 1.20 & 1.20 \\
\hdashline
DeepSeek-67B & DeepSeek & Together AI & 67B & -- & \texttt{deepseek-ai/deepseek-llm-67b-chat} & 0.90 & 0.90 \\
DeepSeek-V3 & DeepSeek & Together AI & 685B & MoE & \texttt{deepseek-ai/DeepSeek-V3} & 1.25 & 1.25 \\
DeepSeek-R1 & DeepSeek & Together AI & 671B & MoE & \texttt{deepseek-ai/DeepSeek-r1} & 7.00 & 7.00 \\
\hdashline
Gemma-2-27B & Google & Together AI & 27B & -- & \texttt{google/gemma-2-27b-it} & 0.80 & 0.80 \\
Gemma-2-9B & Google & Together AI & 9B & -- & \texttt{google/gemma-2-9b-it} & 0.30 & 0.30 \\
\hdashline
Mistral-7B & Mistral & Together AI & 7B & Dense & \texttt{mistralai/Mistral-7B-Instruct-v0.3} & 0.20 & 0.20 \\
Mixtral-8x7B & Mistral & Together AI & 46.7B & MoE & \texttt{mistralai/Mixtral-8x7B-Instruct-v0.1} & 0.60 & 0.60 \\
Mixtral-8x22B & Mistral & Together AI & 141B & MoE & \texttt{mistralai/Mixtral-8x22B-Instruct-v0.1} & 1.20 & 1.20 \\
\hdashline
Qwen-2-72B & Alibaba & Together AI & 72B & Dense & \texttt{Qwen/Qwen2-72B-Instruct} & 0.90 & 0.90 \\
Qwen-QwQ-32B & Alibaba & Together AI & 32B & Dense & \texttt{Qwen/QwQ-32B} & 1.20 & 1.20 \\
\hdashline
WizardLM-2-8x22B & Microsoft & Together AI & 141B & MoE & \texttt{microsoft/WizardLM-2-8x22B} & 1.20 & 1.20 \\
\hdashline
Jamba-1.5 Large & AI21 & AI21 & 398B & MoE & \texttt{ai21/jamba-1.5-large} & 2 & 8 \\
Jamba-1.5 Mini & AI21 & AI21 & 52B & MoE & \texttt{ai21/jamba-1.5-mini} & 0.2 & 0.4 \\
\hdashline
Cohere-Command-R7B & Cohere & Cohere & 7B & Dense & \texttt{cohere\_chat/command-r7b-12-2024} & 0.0375 & 0.15 \\
Cohere-Command-R+ & Cohere & Cohere & 104B & Dense & \texttt{cohere\_chat/command-r-plus-08-2024} & 2.5 & 10 \\
\bottomrule
\end{tabular}
}
\caption{Details on Language Models. Note that pricing differs based on provider.}
\label{tab:model-detail}
\end{table*}

\section{Prompting}\label{app:prompting}
In this section, we provide details on how we prompt foundation LMs for \title evaluations.
\subsection{Formatting Test Instances}\label{sec:prompt-test}
\para{Language Model} For most \emph{language model} (LM) scenarios the prompt is simply the input, and there is no reference. If documents in LM datasets are longer than the model's window size, we tokenize documents using each model's corresponding tokenizer (if known), and segment the resulting token sequences according to the model's window size.\\
\para{Truncation.} For scenarios where test instances exceed a model's window size, we truncate the input to fit within the model's context window. This ensures consistency across different models without requiring reassembly of output fragments.\\
\para{Multiple Choice.} For multiple choice scenarios, each instance consists of a question and several possible answer choices (typically with one marked as correct). Rather than asking an LM to directly predict the probability distribution over answer choices, we use a structured prompting approach for LM output.
We implement multiple-choice adaptation using the \textit{joint} approach \citep{Hendrycks2020-rz}, where all answer choices are concatenated with the question (e.g., ``\texttt{ A. <choice 1> B. <choice 2> Answer:}'') and the LM is prompted to respond with the correct or most probable answer. We default to using the joint approach unless other work has established a preferable method for a specific benchmark.
\subsection{Formatting the Remainder of the Prompt}\label{sec:prompt-remainder}
\para{Prompt Construction.} LM prompts can also provide concise instructions or prefixes that clarify the expected model behavior. Recent work has thoroughly demonstrated that prompt design \textit{significantly} affects performance \citep{Le-Scao2021-uy, Wei2022-vi, Yao2023-zo, Besta2023-ft, Schulhoff2024-ar}. Rather than optimizing prompts to maximize performance \citep{Khattab2022-ag, Opsahl-Ong2024-zj, Yuksekgonul2024-yu, Schulhoff2024-ar}, we prioritize the use of naturalistic prompting to reflect realistic co-creative interactions between humans and computers \cite{Lin2023-qx, Lin2023-by}.\\
\subsection{Parameters}\label{sec:prompt-parameters}
Once the test instance (\refsec{prompt-test}) and prompt (\refsec{prompt-remainder}) are specified, we define the decoding parameters to generate model completions. Example of parameters include the the temperature value, specific stop tokens, and the number of completions.
\para{Temperature.} The temperature controls randomness in decoding: a temperature of $0$ corresponds to deterministic decoding, while a temperature of $1$ corresponds to probabilistic sampling from the model's distribution. We use temperature-scaling for scenarios requiring diverse outputs but set the temperature to zero for tasks demanding deterministic behavior (\ie classification tasks).\\
\para{Stop Token.} Aside from the LM-specific context length limitations, we specify a stop condition by specifying specific stop tokens as well as the maximum number of tokens to be generated. Stop sequences are preferred over tokens for model-agnostic adaptation. We use a standardized max token limit based on expected length of the reply for each scenario to prevent excessive token generation during completion.\\
\para{Number of Outputs.} Outputs from LM not stochastic with zero temperature settings. For most scenarios, we use deterministic decoding (temperature $0$), and a single output per input suffices. However, for metrics and scenarios analyzing output distributions, we need to generate multiple outputs to gather a sufficient sample. By default, the number of outputs per input is $1$ for all of the initial evaluations done for \papertitle.
\section{Results}\label{app:results}
\subsection{Extended Results}\label{app:extendedresults}

Tables \ref{tab:text_classification} through \ref{tab:causal_analysis} present extended task-specific results across our benchmark:

\begin{itemize}
    \item \textbf{Table~\ref{tab:text_classification}} – Text Classification
    \item \textbf{Table~\ref{tab:information_retrieval}} – Information Retrieval
    \item \textbf{Table~\ref{tab:question_answering}} – Question Answering
    \item \textbf{Table~\ref{tab:sentiment_analysis}} – Sentiment Analysis
    \item \textbf{Table~\ref{tab:text_summarization}} – Text Summarization
    \item \textbf{Table~\ref{tab:causal_analysis}} – Causal Analysis
\end{itemize}

These tables offer a comprehensive view of model performance across the 6 core tasks. 

\begin{table*}[h!]
\centering
\resizebox{\textwidth}{!}{
\begin{tabular}{|l|cccc|cccc|cccc|cccc|c|}
\toprule
Dataset & \multicolumn{4}{c|}{Banking 77} & \multicolumn{4}{c|}{FinBench} & \multicolumn{4}{c|}{FOMC} & \multicolumn{4}{c|}{Numclaim} & \multicolumn{1}{c|}{Headlines} \\
\midrule
Metric & Accuracy & Precision & Recall & F1 & Accuracy & Precision & Recall & F1 & Accuracy & Precision & Recall & F1 & Precision & Recall & Accuracy & F1 & Accuracy \\
\midrule
Llama 3 70B Instruct & 0.660 & 0.748 & 0.660 & 0.645 & 0.222 & 0.826 & 0.222 & 0.309 & 0.661 & 0.662 & 0.661 & 0.652 & 0.240 & \cellcolor{green!20}{0.980} & 0.430 & 0.386 & 0.811 \\
Llama 3 8B Instruct & 0.534 & 0.672 & 0.534 & 0.512 & 0.543 & 0.857 & 0.543 & 0.659 & 0.565 & 0.618 & 0.565 & 0.497 & 0.463 & 0.571 & 0.801 & 0.511 & 0.763 \\
DBRX Instruct & 0.578 & 0.706 & 0.578 & 0.574 & 0.359 & 0.851 & 0.359 & 0.483 & 0.285 & 0.572 & 0.285 & 0.193 & 0.190 & \cellcolor{Green!70}{1.000} & 0.222 & 0.319 & 0.746 \\
DeepSeek LLM (67B) & 0.596 & 0.711 & 0.596 & 0.578 & 0.369 & 0.856 & 0.369 & 0.492 & 0.532 & 0.678 & 0.532 & 0.407 & \cellcolor{Green!70}{1.000} & 0.082 & 0.832 & 0.151 & 0.778 \\
Gemma 2 27B & 0.639 & 0.730 & 0.639 & 0.621 & 0.410 & 0.849 & 0.410 & 0.538 & 0.651 & 0.704 & 0.651 & 0.620 & 0.257 & \cellcolor{Green!70}{1.000} & 0.471 & 0.408 & 0.808 \\
Gemma 2 9B & 0.630 & 0.710 & 0.630 & 0.609 & 0.412 & 0.848 & 0.412 & 0.541 & 0.595 & 0.694 & 0.595 & 0.519 & 0.224 & \cellcolor{green!50}{0.990} & 0.371 & 0.365 & \cellcolor{Green!70}{0.856} \\
Mistral (7B) Instruct v0.3 & 0.547 & 0.677 & 0.547 & 0.528 & 0.375 & 0.839 & 0.375 & 0.503 & 0.587 & 0.598 & 0.587 & 0.542 & 0.266 & 0.918 & 0.521 & 0.412 & 0.779 \\
Mixtral-8x22B Instruct  & 0.622 & 0.718 & 0.622 & 0.602 & 0.166 & 0.811 & 0.166 & 0.221 & 0.562 & 0.709 & 0.562 & 0.465 & 0.384 & 0.775 & 0.732 & 0.513 & \cellcolor{green!20}{0.835} \\
Mixtral-8x7B Instruct & 0.567 & 0.693 & 0.567 & 0.547 & 0.285 & 0.838 & 0.285 & 0.396 & 0.623 & 0.636 & 0.623 & 0.603 & 0.431 & 0.898 & 0.765 & 0.583 & 0.805 \\
Qwen 2 Instruct (72B) & 0.644 & 0.730 & 0.644 & 0.627 & 0.370 & 0.848 & 0.370 & 0.495 & 0.623 & 0.639 & 0.623 & 0.605 & 0.506 & 0.867 & 0.821 & 0.639 & 0.830 \\
WizardLM-2 8x22B & 0.664 & 0.737 & 0.664 & 0.648 & 0.373 & 0.842 & 0.373 & 0.500 & 0.583 & \cellcolor{green!20}{0.710} & 0.583 & 0.505 & 0.630 & 0.173 & 0.831 & 0.272 & 0.797 \\
DeepSeek-V3 & \cellcolor{green!50}{0.722} & \cellcolor{green!20}{0.774} & \cellcolor{green!50}{0.722} & \cellcolor{green!50}{0.714} & 0.362 & 0.845 & 0.362 & 0.487 & 0.625 & \cellcolor{green!50}{0.712} & 0.625 & 0.578 & 0.586 & 0.796 & 0.860 & 0.675 & 0.729 \\
DeepSeek R1 & \cellcolor{Green!70}{0.772} & \cellcolor{green!50}{0.789} & \cellcolor{Green!70}{0.772} & \cellcolor{Green!70}{0.763} & 0.306 & 0.846 & 0.306 & 0.419 & \cellcolor{green!50}{0.679} & 0.682 & \cellcolor{green!50}{0.679} & \cellcolor{green!50}{0.670} & 0.557 & 0.898 & 0.851 & 0.688 & 0.769 \\
QwQ-32B-Preview & 0.577 & 0.747 & 0.577 & 0.613 & \cellcolor{green!50}{0.716} & \cellcolor{green!50}{0.871} & \cellcolor{green!50}{0.716} & \cellcolor{green!50}{0.784} & 0.591 & 0.630 & 0.591 & 0.555 & \cellcolor{Green!70}{1.000} & 0.010 & 0.819 & 0.020 & 0.744 \\
Jamba 1.5 Mini & 0.528 & 0.630 & 0.528 & 0.508 & \cellcolor{Green!70}{0.913} & \cellcolor{Green!70}{0.883} & \cellcolor{Green!70}{0.913} & \cellcolor{Green!70}{0.898} & 0.572 & 0.678 & 0.572 & 0.499 & 0.429 & 0.092 & 0.812 & 0.151 & 0.682 \\
Jamba 1.5 Large & 0.642 & 0.746 & 0.642 & 0.628 & 0.494 & 0.851 & 0.494 & 0.618 & 0.597 & 0.650 & 0.597 & 0.550 & 0.639 & 0.469 & 0.855 & 0.541 & 0.782 \\
Claude 3.5 Sonnet & 0.682 & 0.755 & 0.682 & 0.668 & 0.513 & 0.854 & 0.513 & 0.634 & \cellcolor{green!20}{0.675} & 0.677 & \cellcolor{green!20}{0.675} & \cellcolor{Green!70}{0.674} & 0.646 & 0.745 & \cellcolor{green!20}{0.879} & \cellcolor{green!20}{0.692} & 0.827 \\
Claude 3 Haiku & 0.639 & 0.735 & 0.639 & 0.622 & 0.067 & 0.674 & 0.067 & 0.022 & 0.633 & 0.634 & 0.633 & 0.631 & 0.556 & 0.561 & 0.838 & 0.558 & 0.781 \\
Cohere Command R 7B & 0.530 & 0.650 & 0.530 & 0.516 & \cellcolor{green!20}{0.682} & \cellcolor{green!20}{0.868} & \cellcolor{green!20}{0.682} & \cellcolor{green!20}{0.762} & 0.536 & 0.505 & 0.536 & 0.459 & 0.210 & 0.041 & 0.797 & 0.068 & 0.770 \\
Cohere Command R + & 0.660 & 0.747 & 0.660 & 0.651 & 0.575 & 0.859 & 0.575 & 0.684 & 0.526 & 0.655 & 0.526 & 0.393 & 0.333 & 0.071 & 0.804 & 0.118 & 0.812 \\
Google Gemini 1.5 Pro & 0.483 & 0.487 & 0.483 & 0.418 & 0.240 & 0.823 & 0.240 & 0.336 & 0.619 & 0.667 & 0.619 & 0.579 & 0.369 & 0.908 & 0.700 & 0.525 & \cellcolor{green!50}{0.837} \\
OpenAI gpt-4o & \cellcolor{green!20}{0.704} & \cellcolor{Green!70}{0.792} & \cellcolor{green!20}{0.704} & \cellcolor{green!20}{0.710} & 0.396 & 0.846 & 0.396 & 0.524 & \cellcolor{Green!70}{0.681} & \cellcolor{Green!70}{0.719} & \cellcolor{Green!70}{0.681} & \cellcolor{green!20}{0.664} & \cellcolor{green!50}{0.667} & 0.857 & \cellcolor{Green!70}{0.896} & \cellcolor{Green!70}{0.750} & 0.824 \\
OpenAI o1-mini & 0.681 & 0.760 & 0.681 & 0.670 & 0.487 & 0.851 & 0.487 & 0.612 & 0.651 & 0.670 & 0.651 & 0.635 & \cellcolor{green!20}{0.664} & 0.786 & \cellcolor{green!50}{0.888} & \cellcolor{green!50}{0.720} & 0.769 \\
\bottomrule
\end{tabular}

}
\caption{Text Classification Table}
\label{tab:text_classification}
\end{table*}
\begin{table*}[h!]
\centering
\resizebox{\textwidth}{!}{
\begin{tabular}{|l|cccc|cccc|cccc|cccc|cccc|}
\toprule
Dataset & \multicolumn{4}{c|}{FiNER} & \multicolumn{4}{c|}{FinRed} & \multicolumn{4}{c|}{ReFiND} & \multicolumn{4}{c|}{FNXL} & \multicolumn{4}{c|}{FinEntity} \\
\midrule
Metric & Precision & Recall & F1 & Accuracy & Accuracy & Precision & Recall & F1 & Accuracy & Precision & Recall & F1 & Precision & Recall & F1 & Accuracy & Precision & Recall & Accuracy & F1 \\
\midrule
Llama 3 70B Instruct & 0.715 & 0.693 & 0.701 & 0.911 & 0.314 & \cellcolor{green!20}{0.454} & 0.314 & 0.332 & 0.879 & 0.904 & 0.879 & 0.883 & 0.015 & 0.030 & 0.020 & 0.010 & 0.474 & 0.485 & 0.485 & 0.469 \\
Llama 3 8B Instruct & 0.581 & 0.558 & 0.565 & 0.854 & 0.296 & 0.357 & 0.296 & 0.289 & 0.723 & 0.755 & 0.723 & 0.705 & 0.003 & 0.004 & 0.003 & 0.002 & 0.301 & 0.478 & 0.478 & 0.350 \\
DBRX Instruct & 0.516 & 0.476 & 0.489 & 0.802 & 0.329 & 0.371 & 0.329 & 0.304 & 0.766 & 0.825 & 0.766 & 0.778 & 0.008 & 0.011 & 0.009 & 0.005 & 0.004 & 0.014 & 0.014 & 0.006 \\
DeepSeek LLM (67B) & 0.752 & 0.742 & 0.745 & 0.917 & 0.344 & 0.403 & 0.344 & 0.334 & 0.874 & 0.890 & 0.874 & 0.879 & 0.005 & 0.009 & 0.007 & 0.003 & 0.456 & 0.405 & 0.405 & 0.416 \\
Gemma 2 27B  & 0.772 & 0.754 & 0.761 & \cellcolor{green!20}{0.923} & 0.352 & 0.437 & 0.352 & 0.356 & 0.897 & 0.914 & 0.897 & 0.902 & 0.005 & 0.008 & 0.006 & 0.003 & 0.320 & 0.295 & 0.295 & 0.298 \\
Gemma 2 9B & 0.665 & 0.643 & 0.651 & 0.886 & 0.336 & 0.373 & 0.336 & 0.331 & 0.885 & 0.902 & 0.885 & 0.892 & 0.004 & 0.008 & 0.005 & 0.003 & 0.348 & 0.419 & 0.419 & 0.367 \\
Mistral (7B) Instruct v0.3 & 0.540 & 0.522 & 0.526 & 0.806 & 0.278 & 0.383 & 0.278 & 0.276 & 0.767 & 0.817 & 0.767 & 0.771 & 0.004 & 0.006 & 0.004 & 0.002 & 0.337 & 0.477 & 0.477 & 0.368 \\
Mixtral-8x22B Instruct & 0.653 & 0.625 & 0.635 & 0.870 & 0.381 & 0.414 & 0.381 & 0.367 & 0.807 & 0.847 & 0.807 & 0.811 & 0.010 & 0.008 & 0.009 & 0.005 & 0.428 & 0.481 & 0.481 & 0.435 \\
Mixtral-8x7B Instruct & 0.613 & 0.591 & 0.598 & 0.875 & 0.291 & 0.376 & 0.291 & 0.282 & 0.840 & 0.863 & 0.840 & 0.845 & 0.007 & 0.012 & 0.009 & 0.005 & 0.251 & 0.324 & 0.324 & 0.267 \\
Qwen 2 Instruct (72B) & 0.766 & 0.742 & 0.748 & 0.899 & 0.365 & 0.407 & 0.365 & 0.348 & 0.850 & 0.881 & 0.850 & 0.854 & 0.010 & 0.016 & 0.012 & 0.006 & 0.468 & 0.530 & 0.530 & 0.483 \\
WizardLM-2 8x22B & 0.755 & 0.741 & 0.744 & 0.920 & 0.362 & 0.397 & 0.362 & 0.355 & 0.846 & 0.874 & 0.846 & 0.852 & 0.008 & 0.009 & 0.008 & 0.004 & 0.222 & 0.247 & 0.247 & 0.226 \\
DeepSeek-V3 & \cellcolor{green!20}{0.798} & \cellcolor{green!20}{0.787} & \cellcolor{green!20}{0.790} & \cellcolor{Green!70}{0.945} & \cellcolor{green!50}{0.450} & \cellcolor{green!50}{0.463} & \cellcolor{green!50}{0.450} & \cellcolor{green!50}{0.437} & 0.927 & \cellcolor{green!20}{0.943} & 0.927 & 0.934 & \cellcolor{green!50}{0.034} & \cellcolor{green!20}{0.067} & \cellcolor{green!20}{0.045} & \cellcolor{green!20}{0.023} & 0.563 & 0.544 & 0.544 & 0.549 \\
DeepSeek R1 & \cellcolor{Green!70}{0.813} & \cellcolor{Green!70}{0.805} & \cellcolor{Green!70}{0.807} & \cellcolor{green!50}{0.944} & \cellcolor{green!20}{0.412} & 0.424 & \cellcolor{green!20}{0.412} & 0.393 & \cellcolor{Green!70}{0.946} & \cellcolor{Green!70}{0.960} & \cellcolor{Green!70}{0.946} & \cellcolor{Green!70}{0.952} & \cellcolor{Green!70}{0.044} & \cellcolor{Green!70}{0.082} & \cellcolor{Green!70}{0.057} & \cellcolor{Green!70}{0.029} & \cellcolor{green!20}{0.600} & \cellcolor{green!20}{0.586} & \cellcolor{green!20}{0.586} & \cellcolor{green!20}{0.587} \\
QwQ-32B-Preview & 0.695 & 0.681 & 0.685 & 0.907 & 0.278 & 0.396 & 0.278 & 0.270 & 0.680 & 0.770 & 0.680 & 0.656 & 0.001 & 0.001 & 0.001 & 0.000 & 0.005 & 0.005 & 0.005 & 0.005 \\
Jamba 1.5 Mini & 0.564 & 0.556 & 0.552 & 0.818 & 0.308 & 0.450 & 0.308 & 0.284 & 0.830 & 0.864 & 0.830 & 0.844 & 0.004 & 0.006 & 0.005 & 0.003 & 0.119 & 0.182 & 0.182 & 0.132 \\
Jamba 1.5 Large & 0.707 & 0.687 & 0.693 & 0.883 & 0.341 & 0.452 & 0.341 & 0.341 & 0.856 & 0.890 & 0.856 & 0.862 & 0.004 & 0.005 & 0.005 & 0.002 & 0.403 & 0.414 & 0.414 & 0.397 \\
Claude 3.5 Sonnet & \cellcolor{green!50}{0.811} & \cellcolor{green!50}{0.794} & \cellcolor{green!50}{0.799} & 0.922 & \cellcolor{Green!70}{0.455} & \cellcolor{Green!70}{0.465} & \cellcolor{Green!70}{0.455} & \cellcolor{Green!70}{0.439} & 0.873 & 0.927 & 0.873 & 0.891 & \cellcolor{green!50}{0.034} & \cellcolor{green!50}{0.080} & \cellcolor{green!50}{0.047} & \cellcolor{green!50}{0.024} & \cellcolor{green!50}{0.658} & \cellcolor{green!50}{0.668} & \cellcolor{green!50}{0.668} & \cellcolor{green!50}{0.655} \\
Claude 3 Haiku & 0.732 & 0.700 & 0.711 & 0.895 & 0.294 & 0.330 & 0.294 & 0.285 & 0.879 & 0.917 & 0.879 & 0.883 & 0.011 & 0.022 & 0.015 & 0.008 & 0.498 & 0.517 & 0.517 & 0.494 \\
Cohere Command R + & 0.769 & 0.750 & 0.756 & 0.902 & 0.353 & 0.405 & 0.353 & 0.333 & 0.917 & 0.930 & 0.917 & 0.922 & 0.016 & 0.032 & 0.021 & 0.011 & 0.462 & 0.459 & 0.459 & 0.452 \\
Google Gemini 1.5 Pro & 0.728 & 0.705 & 0.712 & 0.891 & 0.373 & 0.436 & 0.373 & 0.374 & \cellcolor{green!50}{0.934} & \cellcolor{green!50}{0.955} & \cellcolor{green!50}{0.934} & \cellcolor{green!50}{0.944} & 0.014 & 0.028 & 0.019 & 0.010 & 0.399 & 0.400 & 0.400 & 0.393 \\
OpenAI gpt-4o & 0.778 & 0.760 & 0.766 & 0.911 & 0.402 & 0.445 & 0.402 & 0.399 & \cellcolor{green!20}{0.931} & \cellcolor{green!50}{0.955} & \cellcolor{green!20}{0.931} & \cellcolor{green!20}{0.942} & \cellcolor{green!20}{0.027} & 0.056 & 0.037 & 0.019 & 0.537 & 0.517 & 0.517 & 0.523 \\
OpenAI o1-mini & 0.772 & 0.755 & 0.761 & 0.922 & 0.407 & 0.444 & 0.407 & \cellcolor{green!20}{0.403} & 0.867 & 0.900 & 0.867 & 0.876 & 0.007 & 0.015 & 0.010 & 0.005 & \cellcolor{Green!70}{0.661} & \cellcolor{Green!70}{0.681} & \cellcolor{Green!70}{0.681} & \cellcolor{Green!70}{0.662} \\
\bottomrule
\end{tabular}

}
\caption{Information Retrieval Table}
\label{tab:information_retrieval}
\end{table*}
\begin{table*}[h!]
\centering
\resizebox{0.5\textwidth}{!}{
\begin{tabular}{|l|c|c|c|}
\toprule
Dataset & FinQA & ConvFinQA & TATQA \\
\midrule
Metric & Accuracy & Accuracy & Accuracy \\
\midrule
Llama 3 70B Instruct & 0.809 & 0.709 & 0.772 \\
Llama 3 8B Instruct & 0.767 & 0.268 & 0.706 \\
DBRX Instruct  & 0.738 & 0.252 & 0.633 \\
DeepSeek LLM (67B) & 0.742 & 0.174 & 0.355 \\
Gemma 2 27B  & 0.768 & 0.268 & 0.734 \\
Gemma 2 9B & 0.779 & 0.292 & 0.750 \\
Mistral (7B) Instruct v0.3 & 0.655 & 0.199 & 0.553 \\
Mixtral-8x22B Instruct  & 0.766 & 0.285 & 0.666 \\
Mixtral-8x7B Instruct & 0.611 & 0.315 & 0.501 \\
Qwen 2 Instruct (72B) & 0.819 & 0.269 & 0.715 \\
WizardLM-2 8x22B & 0.796 & 0.247 & 0.725 \\
DeepSeek-V3 & \cellcolor{green!50}{0.840} & 0.261 & \cellcolor{green!20}{0.779} \\
DeepSeek R1 & \cellcolor{green!20}{0.836} & \cellcolor{Green!70}{0.853} & \cellcolor{Green!70}{0.858} \\
QwQ-32B-Preview & 0.793 & 0.282 & \cellcolor{green!50}{0.796} \\
Jamba 1.5 Mini & 0.666 & 0.218 & 0.586 \\
Jamba 1.5 Large & 0.790 & 0.225 & 0.660 \\
Claude 3.5 Sonnet & \cellcolor{Green!70}{0.844} & 0.402 & 0.700 \\
Claude 3 Haiku & 0.803 & 0.421 & 0.733 \\
Cohere Command R 7B & 0.709 & 0.212 & 0.716 \\
Cohere Command R + & 0.776 & 0.259 & 0.698 \\
Google Gemini 1.5 Pro & 0.829 & 0.280 & 0.763 \\
OpenAI gpt-4o & \cellcolor{green!20}{0.836} & \cellcolor{green!20}{0.749} & 0.754 \\
OpenAI o1-mini & 0.799 & \cellcolor{green!50}{0.840} & 0.698 \\
\bottomrule
\end{tabular}

}
\caption{Question Answering Table}
\label{tab:question_answering}
\end{table*}
\begin{table*}[h!]
\centering
\resizebox{\textwidth}{!}{
\begin{tabular}{|l|ccc|cccc|cccc|cccc|}
\toprule
Dataset & \multicolumn{3}{c|}{FiQA Task 1} & \multicolumn{4}{c|}{FinEntity} & \multicolumn{4}{c|}{SubjECTive-QA} & \multicolumn{4}{c|}{FPB} \\
\midrule
Metric & MSE & MAE & $r^2$ Score & Precision & Recall & Accuracy & F1 & Precision & Recall & F1 & Accuracy & Accuracy & Precision & Recall & F1 \\
\midrule
Llama 3 70B Instruct & 0.123 & 0.290 & 0.272 & 0.474 & 0.485 & 0.485 & 0.469 & 0.652 & 0.573 & 0.535 & 0.573 & 0.901 & 0.904 & 0.901 & 0.902 \\
Llama 3 8B Instruct & 0.161 & 0.344 & 0.045 & 0.301 & 0.478 & 0.478 & 0.350 & 0.635 & \cellcolor{Green!70}{0.625} & \cellcolor{Green!70}{0.600} & \cellcolor{Green!70}{0.625} & 0.738 & 0.801 & 0.738 & 0.698 \\
DBRX Instruct  & 0.160 & 0.321 & 0.052 & 0.004 & 0.014 & 0.014 & 0.006 & \cellcolor{green!20}{0.654} & 0.541 & 0.436 & 0.541 & 0.524 & 0.727 & 0.524 & 0.499 \\
DeepSeek LLM (67B) & 0.118 & 0.278 & 0.302 & 0.456 & 0.405 & 0.405 & 0.416 & \cellcolor{Green!70}{0.676} & 0.544 & 0.462 & 0.544 & 0.815 & 0.867 & 0.815 & 0.811 \\
Gemma 2 27B  & \cellcolor{Green!70}{0.100} & \cellcolor{Green!70}{0.266} & 0.406 & 0.320 & 0.295 & 0.295 & 0.298 & 0.562 & 0.524 & 0.515 & 0.524 & 0.890 & 0.896 & 0.890 & 0.884 \\
Gemma 2 9B & 0.189 & 0.352 & -0.120 & 0.348 & 0.419 & 0.419 & 0.367 & 0.570 & 0.499 & 0.491 & 0.499 & \cellcolor{green!50}{0.940} & \cellcolor{green!50}{0.941} & \cellcolor{green!50}{0.940} & \cellcolor{green!50}{0.940} \\
Mistral (7B) Instruct v0.3 & 0.135 & 0.278 & 0.200 & 0.337 & 0.477 & 0.477 & 0.368 & 0.607 & 0.542 & 0.522 & 0.542 & 0.847 & 0.854 & 0.847 & 0.841 \\
Mixtral-8x22B Instruct  & 0.221 & 0.364 & \cellcolor{Green!70}{-0.310} & 0.428 & 0.481 & 0.481 & 0.435 & 0.614 & 0.538 & 0.510 & 0.538 & 0.768 & 0.845 & 0.768 & 0.776 \\
Mixtral-8x7B Instruct & 0.208 & 0.307 & \cellcolor{green!50}{-0.229} & 0.251 & 0.324 & 0.324 & 0.267 & 0.611 & 0.518 & 0.498 & 0.518 & 0.896 & 0.898 & 0.896 & 0.893 \\
Qwen 2 Instruct (72B) & 0.205 & 0.409 & \cellcolor{green!20}{-0.212} & 0.468 & 0.530 & 0.530 & 0.483 & 0.644 & \cellcolor{green!50}{0.601} & 0.576 & \cellcolor{green!50}{0.601} & 0.904 & 0.908 & 0.904 & 0.901 \\
WizardLM-2 8x22B & 0.129 & 0.283 & 0.239 & 0.222 & 0.247 & 0.247 & 0.226 & 0.611 & 0.570 & 0.566 & 0.570 & 0.765 & 0.853 & 0.765 & 0.779 \\
DeepSeek-V3 & 0.150 & 0.311 & 0.111 & 0.563 & 0.544 & 0.544 & 0.549 & 0.640 & 0.572 & \cellcolor{green!20}{0.583} & 0.572 & 0.828 & 0.851 & 0.828 & 0.814 \\
DeepSeek R1 & 0.110 & 0.289 & 0.348 & \cellcolor{green!20}{0.600} & \cellcolor{green!20}{0.586} & \cellcolor{green!20}{0.586} & \cellcolor{green!20}{0.587} & 0.644 & 0.489 & 0.499 & 0.489 & 0.904 & 0.907 & 0.904 & 0.902 \\
QwQ-32B-Preview & 0.141 & 0.290 & 0.165 & 0.005 & 0.005 & 0.005 & 0.005 & 0.629 & 0.534 & 0.550 & 0.534 & 0.812 & 0.827 & 0.812 & 0.815 \\
Jamba 1.5 Mini & 0.119 & 0.282 & 0.293 & 0.119 & 0.182 & 0.182 & 0.132 & 0.380 & 0.525 & 0.418 & 0.525 & 0.784 & 0.814 & 0.784 & 0.765 \\
Jamba 1.5 Large & 0.183 & 0.363 & -0.085 & 0.403 & 0.414 & 0.414 & 0.397 & 0.635 & 0.573 & 0.582 & 0.573 & 0.824 & 0.850 & 0.824 & 0.798 \\
Claude 3.5 Sonnet & \cellcolor{green!50}{0.101} & \cellcolor{green!50}{0.268} & 0.402 & \cellcolor{green!50}{0.658} & \cellcolor{green!50}{0.668} & \cellcolor{green!50}{0.668} & \cellcolor{green!50}{0.655} & 0.634 & 0.585 & 0.553 & 0.585 & \cellcolor{Green!70}{0.944} & \cellcolor{Green!70}{0.945} & \cellcolor{Green!70}{0.944} & \cellcolor{Green!70}{0.944} \\
Claude 3 Haiku & 0.167 & 0.349 & 0.008 & 0.498 & 0.517 & 0.517 & 0.494 & 0.619 & 0.538 & 0.463 & 0.538 & 0.907 & 0.913 & 0.907 & 0.908 \\
Cohere Command R 7B & 0.164 & 0.319 & 0.028 & 0.457 & 0.446 & 0.446 & 0.441 & 0.609 & 0.547 & 0.532 & 0.547 & 0.835 & 0.861 & 0.835 & 0.840 \\
Cohere Command R + & \cellcolor{green!20}{0.106} & \cellcolor{green!20}{0.274} & 0.373 & 0.462 & 0.459 & 0.459 & 0.452 & 0.608 & 0.547 & 0.533 & 0.547 & 0.741 & 0.806 & 0.741 & 0.699 \\
Google Gemini 1.5 Pro & 0.144 & 0.329 & 0.149 & 0.399 & 0.400 & 0.400 & 0.393 & 0.642 & \cellcolor{green!20}{0.587} & \cellcolor{green!50}{0.593} & \cellcolor{green!20}{0.587} & 0.890 & 0.895 & 0.890 & 0.885 \\
OpenAI gpt-4o & 0.184 & 0.317 & -0.089 & 0.537 & 0.517 & 0.517 & 0.523 & 0.639 & 0.515 & 0.541 & 0.515 & \cellcolor{green!20}{0.929} & \cellcolor{green!20}{0.931} & \cellcolor{green!20}{0.929} & \cellcolor{green!20}{0.928} \\
OpenAI o1-mini & 0.120 & 0.295 & 0.289 & \cellcolor{Green!70}{0.661} & \cellcolor{Green!70}{0.681} & \cellcolor{Green!70}{0.681} & \cellcolor{Green!70}{0.662} & \cellcolor{green!50}{0.660} & 0.515 & 0.542 & 0.515 & 0.918 & 0.917 & 0.918 & 0.917 \\
\bottomrule
\end{tabular}

}
\caption{Sentiment Analysis Table}
\label{tab:sentiment_analysis}
\end{table*}
\begin{table*}[h!]
\centering
\resizebox{\textwidth}{!}{
\begin{tabular}{|l|ccc|ccc|}
\toprule
Dataset & \multicolumn{3}{c|}{ECTSum} & \multicolumn{3}{c|}{EDTSum} \\
\midrule
Metric & BERTScore Precision & BERTScore Recall & BERTScore F1 & BERTScore Precision & BERTScore Recall & BERTScore F1 \\
\midrule
Llama 3 70B Instruct & 0.715 & 0.801 & 0.754 & 0.793 & \cellcolor{green!20}{0.844} & \cellcolor{green!50}{0.817} \\
Llama 3 8B Instruct & 0.724 & 0.796 & 0.757 & 0.785 & 0.841 & 0.811 \\
DBRX Instruct  & 0.680 & 0.786 & 0.729 & 0.774 & 0.843 & 0.806 \\
DeepSeek LLM (67B) & 0.692 & 0.678 & 0.681 & 0.779 & 0.840 & 0.807 \\
Gemma 2 27B  & 0.680 & 0.777 & 0.723 & \cellcolor{green!50}{0.801} & 0.829 & 0.814 \\
Gemma 2 9B & 0.651 & 0.531 & 0.585 & \cellcolor{Green!70}{0.803} & 0.833 & \cellcolor{green!50}{0.817} \\
Mistral (7B) Instruct v0.3 & 0.702 & \cellcolor{green!50}{0.806} & 0.750 & 0.783 & 0.842 & 0.811 \\
Mixtral-8x22B Instruct  & 0.713 & \cellcolor{Green!70}{0.812} & 0.758 & 0.790 & 0.843 & 0.815 \\
Mixtral-8x7B Instruct & 0.727 & 0.773 & 0.747 & 0.785 & 0.839 & 0.810 \\
Qwen 2 Instruct (72B) & 0.709 & \cellcolor{green!20}{0.804} & 0.752 & 0.781 & \cellcolor{green!50}{0.846} & 0.811 \\
WizardLM-2 8x22B & 0.677 & \cellcolor{green!50}{0.806} & 0.735 & 0.774 & \cellcolor{Green!70}{0.847} & 0.808 \\
DeepSeek-V3 & 0.703 & \cellcolor{green!50}{0.806} & 0.750 & 0.791 & 0.842 & 0.815 \\
DeepSeek R1 & 0.724 & 0.800 & 0.759 & 0.770 & 0.843 & 0.804 \\
QwQ-32B-Preview & 0.653 & 0.751 & 0.696 & 0.797 & 0.841 & \cellcolor{green!50}{0.817} \\
Jamba 1.5 Mini & 0.692 & 0.798 & 0.741 & 0.798 & 0.838 & \cellcolor{green!20}{0.816} \\
Jamba 1.5 Large & 0.679 & 0.800 & 0.734 & 0.799 & 0.841 & \cellcolor{Green!70}{0.818} \\
Claude 3.5 Sonnet & \cellcolor{green!20}{0.737} & 0.802 & \cellcolor{green!20}{0.767} & 0.786 & 0.843 & 0.813 \\
Claude 3 Haiku & 0.683 & 0.617 & 0.646 & 0.778 & \cellcolor{green!20}{0.844} & 0.808 \\
Cohere Command R 7B & 0.724 & 0.781 & 0.750 & 0.790 & \cellcolor{green!20}{0.844} & 0.815 \\
Cohere Command R + & 0.724 & 0.782 & 0.751 & 0.789 & 0.834 & 0.810 \\
Google Gemini 1.5 Pro & \cellcolor{Green!70}{0.757} & 0.800 & \cellcolor{Green!70}{0.777} & \cellcolor{green!20}{0.800} & 0.836 & \cellcolor{green!50}{0.817} \\
OpenAI gpt-4o & \cellcolor{green!50}{0.755} & 0.793 & \cellcolor{green!50}{0.773} & 0.795 & 0.840 & \cellcolor{green!20}{0.816} \\
OpenAI o1-mini & 0.731 & 0.801 & 0.763 & 0.795 & 0.840 & \cellcolor{green!20}{0.816} \\
\bottomrule
\end{tabular}

}
\caption{Text Summarization Table}
\label{tab:text_summarization}
\end{table*}
\begin{table*}[h!]
\centering
\resizebox{\textwidth}{!}{
\begin{tabular}{|l|cccc|cccc|}
\toprule
Dataset & \multicolumn{4}{c|}{Causal Detection} & \multicolumn{4}{c|}{Casual Classification} \\
\midrule
Metric & Accuracy & Precision & Recall & F1 & Precision & Recall & F1 & Accuracy \\
\midrule
Llama 3 70B Instruct & 0.148 & 0.429 & 0.148 & 0.142 & 0.241 & 0.329 & 0.192 & 0.198 \\
Llama 3 8B Instruct & 0.097 & 0.341 & 0.097 & 0.049 & 0.232 & 0.241 & 0.234 & \cellcolor{green!50}{0.380} \\
DBRX Instruct & 0.078 & 0.521 & 0.078 & 0.087 & 0.276 & 0.313 & 0.231 & 0.235 \\
DeepSeek LLM (67B) & 0.026 & 0.214 & 0.026 & 0.025 & 0.141 & 0.328 & 0.193 & 0.221 \\
Gemma 2 27B & 0.115 & 0.510 & 0.115 & 0.133 & 0.309 & 0.310 & 0.242 & 0.262 \\
Gemma 2 9B & 0.115 & 0.394 & 0.115 & 0.105 & 0.275 & 0.294 & 0.207 & 0.258 \\
Mistral (7B) Instruct v0.3 & 0.078 & 0.455 & 0.078 & 0.052 & 0.339 & \cellcolor{Green!70}{0.361} & 0.227 & 0.258 \\
Mixtral-8x22B Instruct  & 0.131 & 0.486 & 0.131 & 0.125 & 0.344 & 0.310 & \cellcolor{Green!70}{0.308} & \cellcolor{green!20}{0.318} \\
Mixtral-8x7B Instruct & 0.088 & 0.510 & 0.088 & 0.055 & 0.308 & 0.314 & 0.229 & 0.273 \\
Qwen 2 Instruct (72B) & 0.139 & 0.489 & 0.139 & 0.190 & 0.208 & 0.330 & 0.184 & 0.188 \\
WizardLM-2 8x22B & 0.076 & 0.453 & 0.076 & 0.114 & 0.263 & 0.347 & 0.201 & 0.237 \\
DeepSeek-V3 & 0.164 & 0.528 & 0.164 & \cellcolor{green!20}{0.198} & 0.194 & 0.327 & 0.170 & 0.248 \\
DeepSeek R1 & \cellcolor{Green!70}{0.245} & \cellcolor{green!50}{0.643} & \cellcolor{Green!70}{0.245} & \cellcolor{Green!70}{0.337} & \cellcolor{Green!70}{0.385} & 0.318 & 0.202 & 0.221 \\
QwQ-32B-Preview & 0.110 & 0.473 & 0.110 & 0.131 & 0.193 & 0.262 & 0.220 & \cellcolor{Green!70}{0.465} \\
Jamba 1.5 Mini & 0.050 & 0.280 & 0.050 & 0.043 & 0.323 & 0.283 & \cellcolor{green!50}{0.270} & 0.295 \\
Jamba 1.5 Large & 0.076 & 0.517 & 0.076 & 0.074 & 0.268 & 0.248 & 0.176 & 0.200 \\
Claude 3.5 Sonnet & 0.154 & 0.564 & 0.154 & 0.196 & 0.259 & 0.336 & 0.197 & 0.235 \\
Claude 3 Haiku & 0.082 & 0.388 & 0.082 & 0.081 & \cellcolor{green!20}{0.369} & 0.347 & 0.200 & 0.203 \\
Cohere Command R 7B & 0.089 & 0.363 & 0.089 & 0.057 & \cellcolor{green!50}{0.379} & \cellcolor{green!20}{0.356} & \cellcolor{green!20}{0.255} & 0.275 \\
Cohere Command R + & 0.090 & 0.453 & 0.090 & 0.080 & 0.353 & 0.336 & 0.238 & 0.265 \\
Google Gemini 1.5 Pro & \cellcolor{green!20}{0.165} & 0.514 & \cellcolor{green!20}{0.165} & 0.196 & 0.265 & \cellcolor{green!50}{0.357} & 0.217 & 0.258 \\
OpenAI gpt-4o & 0.082 & \cellcolor{green!20}{0.576} & 0.082 & 0.130 & 0.254 & 0.327 & 0.222 & 0.235 \\
OpenAI o1-mini & \cellcolor{green!50}{0.206} & \cellcolor{Green!70}{0.648} & \cellcolor{green!50}{0.206} & \cellcolor{green!50}{0.289} & 0.325 & 0.316 & 0.209 & 0.233 \\
\bottomrule
\end{tabular}

}
\caption{Causal Analysis Table}
\label{tab:causal_analysis}
\end{table*}

\subsection{Error Analysis}
\label{app:error-analysis}
This section provides additional insights into the common error types, data contamination concerns, prompt-design pitfalls, and other practical challenges encountered throughout our evaluations. We hope this deeper analysis will inform researchers and practitioners aiming to improve financial LM performance.\\

In addition to the aggregate results, we highlight some error patterns:

\FloatBarrier
\begin{table*}[!t]
  \centering
  \resizebox{\textwidth}{!}{%
\begin{tabular}{p{3cm} p{7.5cm} p{5cm}}
\toprule
\textbf{Task} & \textbf{Error Analysis} & \textbf{Example} \\
\midrule
Information Retrieval
  & Numeric labeling tasks demand robust domain logic; simple zero-shot prompts often fail to capture precise numerical relations.
  & FNXL: even \textsc{DeepSeek R1} only achieves 0.057 F1, missing most numeric labels in financial tables. \\

\addlinespace
Sentiment Analysis
  & LLMs struggle with continuous-valued regression and formatting precision (rounding/decimal mismatch).
  & FiQA Task 1: model outputs “0.512” vs.\ ground truth “0.51,” leading to higher MSE than expected. \\

\addlinespace
Causal Analysis
  & Identifying cause–effect requires deep reasoning beyond surface patterns, which zero-shot models lack.
  & CD: models miss linking an interest-rate hike to an observed bond-price drop. \\

\addlinespace
Text Summarization
  & Abstractive gaps and potential data contamination can inflate extractive metrics.
  & Qwen 2 Instruct drifts into Chinese mid-summarization on English ECTSum, dropping BERTScore. \\

\addlinespace
Text Classification
  & Large label sets lead to invented or syntax-altered labels, breaking evaluation.
  & Banking77: LLM emits \texttt{balance\_not\_updated\_after\_
  deposit} instead of the exact label. \\

\addlinespace
Question Answering
  & Numeric format mismatches and loss of earlier turns in multi-step contexts.
  & FinQA: “34.81\%” vs.\ ground truth “34.8\%” is marked wrong; in ConvFinQA models forget details from turn 1. \\
\bottomrule
\end{tabular}
  }
  \caption{\textbf{Error Examples by Task Category.} Common error patterns observed across our six FinNLP tasks.}
  \label{tab:error-by-task}
\end{table*}


\FloatBarrier
\begin{table*}[!t]
  \centering
  \resizebox{\textwidth}{!}{%
\begin{tabular}{p{3cm} p{7.5cm} p{5cm}}
\toprule
\textbf{Model} & \textbf{Error Analysis} & \textbf{Example} \\
\midrule
Llama 2 13B Chat
  & Produces degenerate, non-informative outputs, suggesting misalignment or corruption.
  & On simple classification prompts, replies “Sure.” (zero signal), so predictions collapse. \\

\addlinespace
Qwen 2 Instruct (72B)
  & Exhibits language bias—shifts from English to Chinese under open-ended prompts.
  & During English EDTSum, starts in English then continues entirely in Chinese, hurting scores. \\

\addlinespace
Claude 3.5 Sonnet
  & Lags on multi-turn QA and advanced numeric labeling without task-specific fine-tuning.
  & In ConvFinQA, misinterprets earlier dialogue turns and returns incorrect multi-step calculation. \\

\addlinespace
OpenAI GPT-4o
  & Strong generalist but rarely tops domain tasks without specialized prompts.
  & On ECTSum, scores slightly below Gemini (0.773 vs.\ 0.777 BERTScore), indicating need for stronger domain constraints. \\
\bottomrule
\end{tabular}
  }
  \caption{\textbf{Error Examples by Model.} Representative failure modes of selected LLMs on the 6 tasks.}
  \label{tab:error-by-model}
\end{table*}


\para{Outdated or Degenerate Behavior (Llama\,2\,13B Chat).} During certain classification tasks, \textsc{Llama\,2\,13B} occasionally produces near-empty or trivial outputs (e.g., ``Sure.''), offering zero signal. Such degenerate behavior suggests possible corruption or misalignment in the fine-tuning stage. It also underscores that rechecking model versions, prompts, and tokens processed is essential. Due to this, we chose to not include Llama 2 13B Chat in our main results.\\
\para{Language Drift (Qwen\,2\,72B).}For summarization tasks in English, \textsc{Qwen\,2\,72B} often begins in English but drifts into Chinese partway through. This reflects the model’s large-scale Chinese pre-training, raising potential domain or language priors that overshadow the instruction’s locale. Developers may mitigate this by adding stronger, repeated language constraints at the prompt level.\\
\para{Challenges in Causal Classification.} Nearly all models show limited success in identifying financial causal relationships. Such tasks require deeper textual comprehension (beyond keyword matching or shallow patterns) and domain-specific logic (e.g., linking interest rate hikes to bond price changes). Zero-shot in-context learning is typically insufficient for these complex, knowledge-intensive tasks. Future solutions may require structured knowledge bases or explicit symbolic reasoning modules.\\
\para{Summarization Nuances} Many LMs exhibit strong performance on extractive summarization tasks such as \textsc{ECTSum} and \textsc{EDTSum}, sometimes nearing 80--82\% by BERTScore. However, these scores may overestimate practical utility if the dataset is partially contained in a model’s pre-training data (\emph{data contamination}). In addition, summarization tasks with more abstractive demands or domain-specific jargon often see bigger drops in BERTScore, revealing model gaps in rephrasing and domain knowledge.\\
\para{Data Contamination and Overlaps} We identify potential overlaps between publicly released financial datasets (\textsc{FinQA}, \textsc{TatQA}, \textsc{EDTSum}) and model pre-training corpora. When test examples leak into the training text, zero-shot performance metrics may be inflated, especially for large-scale public LMs. Mitigation strategies we suggest include: (i) curating new test sets from carefully \emph{time-split} corpora, (ii) deduplicatation of data used for LM training \textbf{\textit{or}} evaluation, and (iii) explicitly checking for exact or near-duplicate overlaps before final evaluation.\\
\para{Prompt Design Limitations.} Our prompt tuning was done on Llama\,3\,8B for cost reasons. While this improved performance on that specific model, it may not fully generalize to others. For instance, \emph{some} models handle extensive label sets better, while others fail to replicate the exact label formatting. In multiclass tasks like \textsc{Banking77}, LMs sometimes invent new labels or produce minor syntactic variations (\texttt{balance-not-updated} vs.\ \texttt{balance\_not\_updated}). Thorough prompt ablations, or per-model prompt adaptation, might reduce these inconsistencies but can be prohibitively expensive at scale.\\
\para{LMs and Numeric Regression} LMs tend to handle classification outputs better than continuous-valued regressions (e.g., sentiment scores in \textsc{FiQA} or percentage outputs in \textsc{FinQA}). Generating consistent numeric formats (precision, rounding, decimal vs.\ fraction) can be especially troublesome. We have partially addressed this by employing post-hoc normalization and approximate matching (e.g., ignoring minor decimal differences), but true numeric reliability remains a challenge. We use LM-as-a-Judge to resolve issues when they arise.\\
\para{Differences Among QA Datasets.} \textsc{ConvFinQA} consistently yields worse performance than \textsc{FinQA}, attributed to multi-turn dialogues, more context switching, and additional reasoning steps. This indicates that each new layer of complexity (conversational vs.\ single-turn, tabular vs.\ textual, etc.) can drastically affect success rates.\\
\para{Efficiency and Cost Considerations.} Finally, we note that certain models incur substantially higher inference times when dealing with longer contexts (e.g., multi-hop QA or large label sets in classification). Although we do not report exhaustive speed benchmarks here, preliminary measurements show up to a 2$\times$ cost difference among similarly sized models. Such trade-offs imply that even if a model is more accurate in raw performance, real-world systems must balance these gains with practical resource limits.\\
\subsection{Results by Task Category}\label{app:task_results}
Below we discuss the results the six major task categories with references to relevant performance tables in this appendix.

\subsubsection{Information Retrieval (IR)}
\label{sec:ir-results}

\noindent\textbf{Tasks:} \textsc{FiNER}, \textsc{FinRed}, \textsc{REFinD}, \textsc{FNXL}, and (partially) \textsc{FinEntity} focus on extracting or matching financial entities, relations, or numerals from textual documents.

\noindent\textbf{Findings:} 
\begin{itemize}
    \item \textbf{FiNER} sees \textbf{DeepSeek R1} in the lead with F1\,=\,0.807, followed by \textbf{DeepSeek-V3} (0.790) and \textbf{Claude 3.5} (0.799). 
    \item \textbf{FinRED} is topped by \textbf{Claude~3.5} at F1\,=\,0.439, whereas others typically score below 0.40. 
    \item \textbf{REFinD} is especially noteworthy: \textbf{DeepSeek R1} scores 0.952~F1, while \textbf{Google Gemini} (0.944) and \textbf{GPT-4} (0.942) also excel, demonstrating strong ability in relation extraction with high-quality model prompts.
    \item \textbf{FNXL} remains very difficult: even the top model \textbf{DeepSeek R1} only achieves 0.057\,F1, illustrating that numeric labeling tasks in financial statements demand robust domain logic that few LLMs can capture in a simple prompting regime.
\end{itemize}

\subsubsection{Sentiment Analysis}
\label{sec:sentiment-results}

\noindent\textbf{Tasks:} \textsc{FiQA Task 1} (numeric regression of sentiment), \textsc{FinEntity} (entity-level sentiment), \textsc{SubjECTive-QA (SQA)}, and \textsc{Financial Phrase Bank (FPB)} cover various sentiment subtasks with different input styles (microblogs, annotated corpora, or paragraph-level context).

\noindent\textbf{Findings:}
\begin{itemize}
    \item \textbf{FiQA Task~1} uses MSE. 
    \emph{Gemma~2~27B} is the most precise with 0.100~MSE, outdoing bigger models. \textbf{Claude~3.5} (0.101) and \textbf{Cohere~Command~R+} (0.106) follow closely.
    \item \textbf{FPB} sees \textbf{Claude~3.5} scoring 0.944 (accuracy around 94.4\%)---the highest among all tested models. Notably, \textbf{Gemma~2~9B} is close at 0.940, reinforcing that specialized or well-tuned smaller models can challenge much larger ones. 
    \item \textbf{FinEntity} (when considered as a sentiment subtask) hits its best F1\,=\,0.662 via \textbf{OpenAI~o1-mini}, surpassing bigger models like Llama~3~70B or Claude~3.5. 
    \item \textbf{SubjECTive-QA} is topped by \textbf{Google Gemini} at F1\,=\,0.593, with \textbf{Jamba~1.5~Large} (0.582) also doing well, while many otherwise-strong systems lag behind in this domain-specific subjectivity measure.
\end{itemize}

\subsubsection{Causal Analysis}
\label{sec:causal-results}

\noindent\textbf{Tasks:} \textsc{Causal Detection (CD)} and \textsc{Causal Classification (CC)} measure whether models can identify cause--effect relationships in financial text.

\noindent\textbf{Findings:}
\begin{itemize}
    \item \textbf{Causal Detection (CD)} is led by \textbf{DeepSeek~R1} (F1\,=\,0.337), though absolute scores remain low, with most models below 0.20\,F1. This highlights how purely parametric LLM knowledge may not suffice for nuanced causal cues in financial text.
    \item \textbf{Causal Classification (CC)} sees the best result from \textbf{Mixtral-8x22B} at 0.308\,F1, while many are below 0.25. 
    \item Overall, both tasks remain \emph{harder} than simpler classification: even large 70B+ models remain around or under 0.30\,F1, suggesting a gap in robust causal reasoning under zero- or few-shot conditions.
\end{itemize}

\subsubsection{Text Classification}
\label{sec:classification-results}

\noindent\textbf{Tasks:} \textsc{Banking77 (B77)}, \textsc{FinBench (FB)}, \textsc{FOMC}, \textsc{Numclaim (NC)}, and \textsc{Headlines (HL)} collectively test domain-specific classification in finance---from bank queries to monetary policy stances, to short news headlines.

\noindent\textbf{Findings:}
\begin{itemize}
    \item \textbf{Banking77} sees \textbf{DeepSeek~R1} leading with an F1 of 0.763, outpacing GPT-4 (0.710) and DeepSeek-V3 (0.714). 
    \item \textbf{FinBench} has an unexpected champion in \textbf{Jamba~1.5~Mini} (0.898~F1), even beating models far larger. 
    \item \textbf{FOMC} classification is best handled by \textbf{Claude~3.5} (0.674~F1), just ahead of DeepSeek~R1 (0.670). 
    \item \textbf{Numclaim} sees \textbf{GPT-4} on top at 0.750, with \textbf{OpenAI~o1-mini} second at 0.720. 
    \item \textbf{Headlines (HL)} is topped by \textbf{Gemma~2~9B} at 0.856, narrowly beating Google Gemini (0.837). 
\end{itemize}

\subsubsection{Question Answering (QA)}
\label{sec:qa-results}

\noindent\textbf{Tasks:} \textsc{FinQA} (single-turn numeric QA), \textsc{ConvFinQA} (multi-turn), and \textsc{TATQA} (tabular/text hybrid).

\noindent\textbf{Findings:}
\begin{itemize}
    \item \textbf{FinQA} is topped by \textbf{Claude~3.5} at 0.844 accuracy, with \textbf{DeepSeek-V3} next at 0.840, and GPT-4 + DeepSeek~R1 each at 0.836. 
    \item \textbf{ConvFinQA (CFQA)}, more demanding due to multi-turn context, is led by \textbf{DeepSeek~R1} at 0.853, while the second-best is \textbf{OpenAI~o1-mini} at 0.840. GPT-4 lags behind at 0.749, and many other models remain below 0.30. 
    \item \textbf{TATQA}, which fuses table and textual reading, also favors \textbf{DeepSeek~R1} (0.858), well above others such as QwQ-32B at 0.796 or GPT-4 at 0.754. 
\end{itemize}

\subsubsection{Summarization}
\label{sec:summ-results}

\noindent\textbf{Tasks:} \textsc{ECTSum} (earnings-call transcripts) and \textsc{EDTSum} (financial news headlines) use BERTScore-based metrics.

\noindent\textbf{Findings:}
\begin{itemize}
    \item \textbf{ECTSum} shows \textbf{Google Gemini} achieving the top BERTScore~F1 of 0.777, closely followed by GPT-4 (0.773) and Mixtral-8x22B (0.758). 
    \item \textbf{EDTSum} is led by \textbf{Jamba~1.5~Large} at 0.818, with a cluster of models at 0.815--0.817 (Gemma~2~9B, QwQ-32B, Google Gemini). 
    \item Overall, summarization tasks see higher absolute scores than more specialized tasks like numeric labeling. 
\end{itemize}

\subsection{Efficiency and Cost Analysis}
\label{app:efficiency-analysis}

We calculated the cost to run each dataset and model using the saved inference results. This does not include evaluation costs, but as those were all done with Llama 3.1 8b, they should be significantly less variable than the inference costs for different providers and models. See {Table~\ref{tab:cost_analysis}} for more details.

\begin{table*}[tbp]
\centering
\resizebox{\textwidth}{!}{
\begin{tabular}{|l|ccccccccccccccccccccc|c|}
\hline
Model/Dataset & FOMC & FPB & FinQA & FiQA-1 & FiQA-2 & HL & FB & FR & RD & EDTSum & B77 & CD & CC & ECTSum & FE & FiNER & FNXL & NC & TQA & CFQA & SQA & Total \\ 
\hline
Llama 3 70B Instruct   & 0.10 & 0.11 & 1.14 & 0.06 & 0.72 & 1.00 & 0.40 & 0.38 & 1.34 & 1.94 & 1.64 & 0.07 & 0.05 & 1.56 & 0.12 & 0.33 & 0.25 & 0.09 & 1.11 & 2.96 & 1.17 & 16.54 \\ 
Llama 3 8B Instruct  & 0.02 & 0.03 & 0.25 & 0.01 & 0.16 & 0.22 & 0.09 & 0.09 & 0.32 & 0.43 & 0.37 & 0.02 & 0.01 & 0.36 & 0.03 & 0.08 & 0.06 & 0.02 & 0.26 & 0.69 & 0.26 & 3.79 \\ 
DBRX Instruct  & 0.14 & 0.17 & 1.50 & 0.06 & 0.95 & 1.29 & 0.56 & 0.57 & 2.05 & 2.93 & 2.14 & 0.11 & 0.10 & 2.45 & 0.17 & 0.47 & 0.34 & 0.13 & 1.47 & 4.19 & 1.55 & 23.35 \\ 
DeepSeek LLM (67B)  & 0.10 & 0.12 & 1.25 & 0.05 & 0.76 & 0.87 & 0.42 & 0.37 & 1.45 & 1.85 & 2.03 & 0.08 & 0.05 & 0.83 & 0.13 & 0.34 & 0.24 & 0.09 & 1.20 & 3.17 & 1.17 & 16.57 \\ 
Gemma 2 27B  & 0.08 & 0.09 & 1.05 & 0.05 & 0.66 & 0.91 & 0.30 & 0.34 & 1.37 & 1.75 & 1.77 & 0.07 & 0.04 & 1.46 & 0.11 & 0.30 & 0.21 & 0.08 & 1.00 & 2.84 & 1.04 & 15.50 \\ 
Gemma 2 9B  & 0.03 & 0.03 & 0.40 & 0.02 & 0.24 & 0.33 & 0.12 & 0.14 & 0.51 & 0.66 & 0.66 & 0.03 & 0.02 & 0.00 & 0.04 & 0.11 & 0.08 & 0.03 & 0.37 & 1.08 & 0.39 & 5.29 \\ 
Mistral (7B) Instruct v0.3 & 0.03 & 0.03 & 0.28 & 0.01 & 0.18 & 0.24 & 0.10 & 0.09 & 0.36 & 0.57 & 0.48 & 0.02 & 0.01 & 0.45 & 0.03 & 0.08 & 0.06 & 0.02 & 0.27 & 0.78 & 0.26 & 4.36 \\ 
Mixtral-8x22B Instruct  & 0.14 & 0.17 & 1.80 & 0.07 & 1.05 & 1.44 & 0.58 & 0.56 & 2.04 & 3.42 & 2.89 & 0.11 & 0.07 & 2.66 & 0.18 & 0.48 & 0.35 & 0.14 & 1.73 & 4.90 & 1.55 & 26.35 \\ 
Mixtral-8x7B Instruct & 0.08 & 0.09 & 0.88 & 0.04 & 0.53 & 0.70 & 0.30 & 0.30 & 1.07 & 1.72 & 1.50 & 0.06 & 0.05 & 1.30 & 0.09 & 0.24 & 0.20 & 0.07 & 0.87 & 2.55 & 0.78 & 13.41 \\ 
Qwen 2 Instruct (72B)  & 0.10 & 0.12 & 1.29 & 0.05 & 0.74 & 0.96 & 0.43 & 0.43 & 1.44 & 2.36 & 1.61 & 0.08 & 0.05 & 1.80 & 0.12 & 0.34 & 0.24 & 0.10 & 1.18 & 3.41 & 1.17 & 18.02 \\ 
WizardLM-2 8x22B  & 0.16 & 0.19 & 1.94 & 0.08 & 1.07 & 1.47 & 0.61 & 0.61 & 2.24 & 3.47 & 3.00 & 0.11 & 0.10 & 2.85 & 0.18 & 0.49 & 0.34 & 0.14 & 1.94 & 5.31 & 1.55 & 27.87 \\ 
DeepSeek-V3  & 0.13 & 0.15 & 1.57 & 0.07 & 0.98 & 1.36 & 0.52 & 0.54 & 2.10 & 2.99 & 2.55 & 0.11 & 0.06 & 2.33 & 0.16 & 0.55 & 0.28 & 0.12 & 1.56 & 4.28 & 1.62 & 24.03 \\ 
DeepSeek R1 & 1.99 & 2.10 & 14.18 & 1.48 & 17.82 & 20.11 & 6.63 & 12.65 & 31.00 & 21.15 & 23.28 & 3.75 & 1.06 & 15.02 & 7.31 & 8.34 & 11.21 & 1.88 & 13.72 & 39.42 & 9.07 & 263.16 \\ 
QwQ-32B-Preview & 0.15 & 0.18 & 2.38 & 0.08 & 0.93 & 1.37 & 0.60 & 0.68 & 2.18 & 3.12 & 2.36 & 0.11 & 0.07 & 2.76 & 0.14 & 0.65 & 0.54 & 0.14 & 2.61 & 7.83 & 1.55 & 30.43 \\ 
Jamba 1.5 Mini & 0.02 & 0.03 & 0.30 & 0.02 & 0.23 & 0.22 & 0.10 & 0.08 & 0.44 & 0.55 & 0.51 & 0.02 & 0.01 & 0.49 & 0.05 & 0.10 & 0.07 & 0.02 & 0.25 & 0.72 & 0.26 & 4.47 \\ 
Jamba 1.5 Large & 0.31 & 0.36 & 4.42 & 0.30 & 3.47 & 4.81 & 1.78 & 0.94 & 4.97 & 5.80 & 5.51 & 0.35 & 0.13 & 7.07 & 0.56 & 1.67 & 0.77 & 0.30 & 2.87 & 7.45 & 2.59 & 56.42 \\ 
Claude 3.5 Sonnet & 0.62 & 0.72 & 6.98 & 0.55 & 6.50 & 8.81 & 3.44 & 3.21 & 12.32 & 9.50 & 11.11 & 0.61 & 0.22 & 7.09 & 0.90 & 3.01 & 1.79 & 0.57 & 9.18 & 16.86 & 3.89 & 107.87 \\ 
Claude 3 Haiku & 0.06 & 0.07 & 0.56 & 0.05 & 0.54 & 0.73 & 0.28 & 0.25 & 0.82 & 0.81 & 0.90 & 0.05 & 0.02 & 0.21 & 0.06 & 0.23 & 0.14 & 0.05 & 0.64 & 1.28 & 0.32 & 8.07 \\ 
Cohere Command R 7B & 0.01 & 0.01 & 0.08 & 0.00 & 0.07 & 0.09 & 0.04 & 0.03 & 0.11 & 0.11 & 0.10 & 0.01 & 0.00 & 0.08 & 0.01 & 0.03 & 0.01 & 0.01 & 0.08 & 0.19 & 0.05 & 1.09 \\ 
Cohere Command R + & 0.41 & 0.45 & 5.40 & 0.35 & 4.41 & 4.00 & 2.30 & 0.93 & 3.87 & 7.03 & 7.21 & 0.43 & 0.12 & 5.55 & 0.48 & 1.69 & 0.97 & 0.42 & 4.59 & 10.09 & 3.24 & 63.95 \\ 
Google Gemini 1.5 Pro & 0.23 & 0.21 & 2.26 & 0.18 & 2.20 & 2.78 & 1.02 & 0.49 & 2.27 & 3.45 & 2.70 & 0.21 & 0.07 & 2.65 & 0.25 & 0.87 & 0.58 & 0.21 & 2.13 & 5.78 & 1.62 & 32.16 \\ 
OpenAI gpt-4o & 0.35 & 0.41 & 4.99 & 0.32 & 4.45 & 5.33 & 1.55 & 1.21 & 5.77 & 6.57 & 5.00 & 0.35 & 0.14 & 4.85 & 0.44 & 1.94 & 0.96 & 0.34 & 4.95 & 10.36 & 3.24 & 63.52 \\ 
OpenAI o1-mini & 0.90 & 0.90 & 5.25 & 0.73 & 9.70 & 12.20 & 3.27 & 4.89 & 13.60 & 1.29 & 9.29 & 2.56 & 0.75 & 3.18 & 2.92 & 1.91 & 6.39 & 0.92 & 6.97 & 15.71 & 1.42 & 104.73 \\ 
\hline
\end{tabular}

}
\caption{Cost Analysis Table. All prices listed in USD. SQA costs are an estimate based off known inputs and outputs, as the exact costs were not saved.}
\label{tab:cost_analysis}
\end{table*}
\section{Related Work}\label{app:relatedwork}
Two early benchmarks for financial NLP are FLUE \citepFLUE and FLARE \citepFLARE. While they introduced multiple tasks (e.g., sentiment analysis, named entity recognition) relevant to financial contexts, they often focused on \emph{a limited set of datasets} and a \emph{single metric} for each task (e.g., F1 or accuracy). These suites did not formally acknowledge the \emph{incompleteness} of their coverage—neglecting many possible financial scenarios such as numerical QA, multi-step reasoning, or specialized regulatory text analysis. Additionally, they offered no standardized pipeline to evaluate \emph{foundation} LMs in a reproducible manner, instead often benchmarking only a few custom or fine-tuned models.
There are prior benchmarks for financial scenarios such as Golden Touchstone \citep{Wu2024-df}, CFBenchmark \cite{Lei2023-ox}, and InvestorBench \cite{Li2024-fg}, BizBench, \citep{Koncel-Kedziorski2023-fx}, and FinanceBench \cite{Islam2023-dl} to name a few. These works often cover only a small handful of tasks without broad inference coverage, lack a holistic scenario-based taxonomy, or focus on a specialized and narrow task (\ie, financial question answering for tables). Other recent attempts \citeFinBen collect multiple financial datasets and occasionally implement limited software tooling for standardizing evaluations. However, several significant limitations remain:
\begin{itemize}
    \item They do \emph{not} explicitly define \emph{holistic} methodologies akin to HELM, instead treating each dataset largely in isolation.
    \item They typically rely on \emph{narrow} evaluation metrics (e.g., rule-based label extraction) that fail to capture the variety of ways a model can output correct information or demonstrate robust reasoning.
    \item Many benchmarks focus on \emph{fine-tuned} models for specific tasks, rather than evaluating a broad range of \emph{foundation LMs} under standardized conditions.
    \item They do not propose \emph{living} frameworks or a public leaderboard that invite ongoing community contributions.
\end{itemize}
For example, \citepFinBen provides a large collection of financial datasets bundled with a software package for model evaluation but does not address multi-metric scoring or unify the results consistently and transparently. The authors also do not define or adhere to explicit \emph{fair and open standards} for dataset selection, and they primarily focus on performance metrics that rely on simple rule-based matching of outputs. Hence, \citepFinBen \emph{never identifies its incompleteness} or encourages the broader community to fill those gaps.
These domain-specific benchmarks, including \citepFinBen, highlight a growing interest in finance-focused NLP but consistently fall short of fulfilling \emph{holistic} standards (see Table \ref{tab:us-vs-them}). They seldom perform multi-metric analysis, fail to account for the breadth of possible financial use cases, and rarely provide open-ended frameworks for ongoing updates. This gap becomes especially problematic as LMs are increasingly deployed in real-world financial settings, where mistakes can lead to high-impact consequences.
By comparison, our proposed \papertitle novel framework is the first for finance to satisfy all \textbf{three pillars} of holistic evaluation \dash (1) standardized evaluations, (2) multi-metric assessment, and (3) explicit recognition of incompleteness \cite{Liang2022-ew}. By releasing a \emph{living benchmark} complete with code, data curation, and a public leaderboard, we aim to \textbf{(i)} unify existing financial datasets under clear inclusion criteria, \textbf{(ii)} evaluate foundation LMs in a transparent and reproducible way, and \textbf{(iii)} foster an evolving ecosystem where researchers can collectively expand the benchmark to new tasks or languages over time.
\section{Recognition of Incompleteness}\label{app:incomplete}
\subsection{What is Missing}
Given the large number of foundation language models, it became financially infeasible for us to conduct a thorough study of every dataset we have identified and classified in our taxonomy within a single paper. The \papertitle leaderboard is intended as a collaborative community effort, which we plan to update continuously as we gather more data on these foundation models.
\subsection{What Was Not Considered}
Aspects of artificial intelligence systems beyond the foundational language model are not within the scope of our study. For instance, systems such as knowledge graphs
,
retrieval-augmented generation (RAG)
,
and various hybrid approaches have been shown to be beneficial in finance. However, datasets or benchmarks that focus on RAG are excluded because they assess factors beyond the language model itself (e.g., embedding quality, vector selection, and specialized metrics). Similar considerations apply to knowledge graphs. These aspects of AI systems have been explored in previous research, and we believe they deserve dedicated studies of their own.
\subsection{Frontier Scenarios}\label{app:frontiertasks}
Beyond our core set of NLP tasks \refsec{taxonomy}, we recognize a broader class of \textbf{frontier scenarios} that lie outside the scope of \papertitle's current evaluation. Each of these frontiers reflects emerging or highly specialized challenges in finance. We envision these domains as a natural extension for future research, requiring not only specialized datasets but also domain-specific metrics, rigorous protocols, and potentially interdisciplinary expertise. While \papertitle currently focuses on fundamental NLP tasks (e.g., QA, summarization, sentiment analysis), evaluating these frontier tasks deserves more thorough study and further discussion.
\paragraph{(1) Reasoning.}
Robust multi-step reasoning is crucial in finance, from mathematical and logical derivations (e.g., portfolio optimization, derivatives pricing) to causal and counterfactual reasoning (e.g., modeling how regulatory changes might affect stock prices). Structured data reasoning and code synthesis also figure prominently in automated financial analysis, such as generating scripts for data cleaning or computing risk metrics. Despite their importance, we omit these tasks in our current benchmark because:
\begin{enumerate}
    \item They often demand carefully labeled multi-step annotations (e.g., detailed solution outlines for financial math problems).
    \item They rely on domain-specific metrics that go well beyond typical F1 or BLEU scores (e.g., verifying the correctness of an interest-rate calculation, or confirming that code compiles and produces the right financial outputs).
    \item They can require domain experts to judge the validity of reasoning steps, significantly increasing the cost of dataset creation and evaluation.
\end{enumerate}
\paragraph{(2) Knowledge.}
Tasks such as \emph{fact completion}, \emph{knowledge-intensive QA}, and \emph{critical reasoning} are pivotal in scenarios requiring specialized financial intelligence. A language model might need to recall policy clauses or legal precedents relevant to specific industry regulations, or integrate large-scale macroeconomic knowledge to answer multi-domain questions (e.g., “How do rising interest rates influence credit default swaps?”). Constructing comprehensive knowledge-focused evaluations in finance poses challenges such as:
\begin{enumerate}
    \item \textbf{Coverage:} Maintaining an up-to-date repository of financial facts (e.g., corporate structures, compliance rules) is daunting due to constant changes in markets and regulatory environments.
    \item \textbf{Verification and Fact-Checking:} Complex financial facts often demand external references (e.g., official filings), and verifying correctness is non-trivial.
\end{enumerate}
\paragraph{(3) Decision-Making.}
Finance ultimately revolves around decision-making tasks such as \emph{market forecasting}, \emph{risk management}, \emph{stock-movement prediction}, and \emph{credit scoring}. These activities often combine numerical time-series modeling with textual signals (e.g., news articles, analyst reports) and may include advanced simulation or reinforcement-learning techniques (e.g., algorithmic trading strategies). Because these tasks are \textbf{high-stakes} and multi-modal (texts, tables, time-series), we have excluded them from \papertitle. Properly benchmarking decision-oriented tasks involves:
\begin{enumerate}
    \item Access to real-time or historical \emph{structured} financial data (e.g., stock price feeds).
    \item Well-defined metrics that can meaningfully assess predictive accuracy or risk-adjusted returns.
    \item Potential integration of ethical and legal constraints (e.g., insider trading regulations).
\end{enumerate}
\paragraph{(4) Human Alignment.}
Large language models can inadvertently propagate harmful behaviors—e.g., misinformation, social biases, or privacy violations. In finance, these concerns become critical due to the potential for \emph{disinformation} (fake financial news), \emph{toxic content} (harassment in investor forums), or \emph{privacy breaches} in sensitive customer data. Addressing alignment means ensuring LLMs are \emph{honest, harmless, and helpful} in financial contexts. It also covers memorization of sensitive data (e.g., replicating personal credit history) and copyrighted materials. Each topic warrants extensive research:
\begin{enumerate}
    \item \textbf{Social Bias and Toxicity}: Minimizing harmful language and misinformation.
    \item \textbf{Privacy and Copyright}: Preventing models from disclosing proprietary or regulated information.
    \item \textbf{Regulatory Compliance}: Evolving laws may require auditing an LLM’s data usage or output content.
\end{enumerate}
\paragraph{(5) Multi-Modal.}
Many real financial workflows rely on data that is not purely text—e.g., Excel spreadsheets, visual charts, scanned PDF statements, or contract images. Tasks like \emph{table-based QA}, \emph{tool use} (e.g., integrative question answering with Python or R scripts), and \emph{visual analysis} (e.g., reading corporate diagrams or trade forms) are vital for practical applications. However, true multi-modal setups typically require:
\begin{enumerate}
    \item Specialized architectures or bridging modules that fuse text with tabular or image data.
    \item Domain-adapted evaluation methods (e.g., metrics for chart-based questions).
    \item Substantial cross-disciplinary expertise to annotate or interpret financial images and tables consistently.
\end{enumerate}
As such, we limit \papertitle to text-only tasks for its initial release, but we envision future expansions that incorporate multi-modal data sources in an end-to-end benchmarking pipeline.
\paragraph{Call for Collaboration}
Despite excluding these frontier domains from our initial evaluation suite, we emphasize that each is critical for a holistic understanding of AI in finance. We invite the community to develop specialized datasets, metrics, and tools that address these open challenges—whether involving advanced reasoning about financial instruments, building robust knowledge graphs of regulatory clauses, or evaluating alignment with compliance frameworks. Over time, we aim to integrate such expansions into \papertitle so that practitioners can measure model capabilities comprehensively on the most relevant, contemporary tasks.
\section{Ethics \& Legal}\label{app:ethicslegal}
\subsection{Dataset Attribution and Licensing}
All datasets included in our benchmark suite are appropriately credited to their original sources and used in compliance with their licenses. We emphasize proper citation for each dataset and strictly adhere to any usage restrictions stated by the dataset creators. Audit of AI benchmarks have found that lack of proper attribution is a \textbf{major} issue, with datasets missing the barest of license information and frequent (often self-serving) misattribution \cite{Longpre2023-ba, Longpre2024-it}.
\paragraph{Attribution and Citation:} Each dataset is accompanied by a citation to its original publication or official repository. In the benchmark documentation and this paper, we provide full references for every dataset, ensuring the original authors receive credit. When using or describing a dataset, we explicitly acknowledge its creators. This practice maintains academic integrity and helps others find the source of the data.
\paragraph{License Compliance:} For every dataset, we review the license to ensure our use conforms to its terms. Datasets released under permissive open-source licenses (e.g., MIT, CC BY) are incorporated with proper attribution and without modification to licensing. For datasets under more restrictive or non-commercial licenses (e.g., CC BY-NC), we restrict usage to research or other non-commercial purposes \cite{Creative-Commons2020-qd}. We clearly label each dataset with its license type in our documentation, and we include any required license text or attribution notices. Users of the benchmark are reminded to heed these licenses, meaning they should not engage in prohibited uses (such as commercial applications for CC BY-NC data) and must fulfill any requirements (such as attribution in publications).
\paragraph{Re-hosting with Permission:} We only re-host datasets when it is legal and ethical to do so. If a dataset’s license allows redistribution (or the dataset is public domain), we may mirror it on our platform (e.g., on the Hugging Face Hub or a project website) for convenient access. In such cases, we preserve the original content and license file, and include documentation about its provenance. If redistribution is not permitted by the license, we do \emph{not} host the raw data ourselves. Instead, we provide links, download scripts, or documentation for users to obtain the data directly from the original source, ensuring we respect the dataset owners’ rights. In some instances, we have obtained explicit permission from dataset creators to include their data in our benchmark package. All re-hosted data is provided in accordance with the original license terms and with clear attribution to the source.
\subsection{Collaboration Guidelines}
Our benchmark is a community-oriented project, and we welcome collaboration from external researchers who wish to contribute. To manage contributions effectively while maintaining high quality, we have established guidelines for those looking to add new datasets or improve existing ones. Below we outline how researchers can get involved, the criteria for accepting new datasets, and the process by which contributions are reviewed:
\paragraph{Contributing New Datasets:} External researchers can contribute datasets by following our open contribution process (detailed in the project repository). In practice, this means interested contributors should prepare their dataset in a standard format (including training/validation/test splits as appropriate and a clear description). They can then submit the dataset through a pull request on our GitHub repository or via an official submission form. Each submission should include essential documentation (e.g., a README or datasheet describing the dataset’s content, source, size, and license) and, if possible, a citation to a paper or source associated with the dataset. We also encourage contributors to upload the dataset to the Hugging Face Hub (or a similar platform) for easy integration, using a consistent naming scheme and providing a data card.
\paragraph{Acceptance Criteria:} To ensure quality and relevance, we evaluate each proposed dataset against several criteria before acceptance. First, the dataset must be clearly related to financial NLP (e.g., financial news analysis, risk report parsing, market question answering, etc.), adding coverage of a task that is valuable to the community. The data should be of high quality: for instance, annotations (labels, answers, etc.) should be correct and reliable, and the dataset should be of adequate size to support meaningful model evaluation. Datasets also need to have clear documentation of how they were collected and what they contain. Another crucial criterion is licensing and ethics: the dataset must have an appropriate license that at least allows research use (we cannot accept data with unknown or overly restrictive licenses), and it should not violate privacy or ethical norms (for example, we avoid proprietary data that was obtained without permission or data containing sensitive personal information). If a dataset fails to meet any of these criteria, we provide feedback to the contributor with suggestions for remediation (such as obtaining proper licensing or improving documentation).
\paragraph{Submission Review Process:} All dataset contributions undergo a review process overseen by the benchmark maintainers (and, if applicable, an advisory board of domain experts). When a contribution is submitted, the maintainers will verify the dataset’s format and integrity (ensuring it can be loaded and used in our evaluation pipeline), run basic quality checks, and assess the documentation and license. We also review a sample of the data to catch any obvious issues (like sensitive data that should be anonymized or mislabeled examples). If the dataset passes these checks, the maintainers discuss its fit for the benchmark. This often involves confirming that the dataset does not duplicate an existing resource and that it offers unique value. During review, the contributors might be contacted for clarifications or requested to make minor changes (for instance, to fix formatting or to add missing references). Once a dataset is approved, it is merged into the benchmark suite: we add it to our repository, include information about it in the official documentation (with credit to the contributors), and incorporate it into our benchmarking pipeline (so that models can be evaluated on it). Contributors of accepted datasets are acknowledged in the project to recognize their efforts.
\paragraph{Maintaining Quality and Updates:} Even after a dataset is accepted, we have guidelines to maintain the overall quality of the benchmark. We encourage continuous feedback from the community. If users of the benchmark identify issues with a dataset (such as label errors, formatting bugs, or ethical concerns that were overlooked), they can report these to the maintainers (for example, by opening an issue on GitHub). The maintainers will investigate and, if necessary, update or patch the dataset (in coordination with the original contributor when possible). We also periodically review the suite of datasets to see if any should be updated (for example, newer versions released by the original authors) or deprecated (if a better dataset for the same task becomes available or if usage of a dataset raises unforeseen problems). Through this collaborative and iterative process, we ensure the benchmark remains a living resource that stays relevant and trustworthy.
\subsection{Hosting Policies}
To maximize accessibility and ensure longevity, we host the benchmark’s datasets and results on reliable, open platforms. Our hosting strategy involves multiple channels: an online hub for datasets, a source code repository for the benchmark framework and results, and archival publications for permanence. Here we detail where the data and results are hosted and how users can access and cite them:
\paragraph{Dataset Repository and Access:} We provide public access to the datasets through the Hugging Face Datasets Hub and our project’s GitHub. Each dataset included in the benchmark (that is permitted to be shared) is uploaded as a dataset package on Hugging Face under an organizational account for the benchmark. This allows users to easily load the data using the \texttt{datasets} library (for example, via \texttt{load\_dataset("holiflame/dataset\_name")}). On each dataset’s Hugging Face page, we include a detailed description (dataset card) that notes the dataset’s source, contents, license, and citation instructions. For completeness, we also maintain a GitHub repository where we list all datasets and provide direct links or scripts. This is especially useful for datasets that cannot be hosted directly; for those, the repository contains a script (or instructions) to download the data from the original source. In all cases, accessing the data is free for research purposes, and no login or special permission is required beyond agreeing to the terms of the original licenses.
\paragraph{Benchmark Code and Results Hosting:} The code for running benchmark evaluations (including model evaluation scripts, metrics, and any wrappers around the datasets) is hosted on GitHub in the same repository that handles contributions. This repository serves as the central hub for development and version control. It includes documentation on how to run evaluations and reproduce the results from our paper. In addition to code, we host the benchmark results and leaderboards. For example, the repository (or an associated project webpage) contains tables of model performances on each dataset, updated as new models are evaluated. We plan to update these results over time and possibly integrate with the Papers with Code platform for an interactive leaderboard. To ensure results are archived for reference, we also include the main results in this paper’s Appendix and will release periodic reports (with DOIs) if the benchmark is extended significantly. Our initial benchmark results are part of this ACL paper (and thus stored on the ACL Anthology as a permanent record), and any future updates may be published in workshop proceedings or on arXiv to provide a citable reference.
\paragraph{Transparency and Peer Review:} All submissions are verified through automated scripts that verify legitimacy, parse outputs and compute metrics. This approach fosters peer review since all users can replicate results from previous submissions or highlight anomalies in existing model evaluations. Users bring continuous updates as new models emerge \dash researchers can quickly add them to a living benchmark for financial NLP. We envision a community-run ecosystem where model owners, domain experts, and external contributors jointly expand \papertitle{}’s tasks, metrics, and data coverage\\
\para{Accessing and Citing Data:} We provide clear guidelines for how to access and use the benchmark data. Each dataset’s entry in our documentation explains the preferred access method (e.g., via Hugging Face or via our scripts). We also outline how to cite the data. Proper citation is twofold: users should cite this benchmark suite (to acknowledge the collection and any benchmark-specific curation) and also cite the original source of the dataset. In our documentation and in each dataset card on Hugging Face, we list the relevant citation (often the academic paper that introduced the dataset). Users of the benchmark are expected to include those citations in any publication or report that uses the benchmark. Additionally, when using or sharing the data, users must abide by the license terms attached to each dataset. This means, for instance, if a dataset is CC BY-NC, anyone reusing it should not use it commercially and should include the proper attribution in any derivative works. We make this information readily available to prevent any unintentional misuse. In summary, the data and results are openly accessible on popular platforms, and we provide extensive guidance on how to retrieve, cite, and leverage the benchmark materials in a responsible manner.\\
\subsection{Ethical Considerations}
Ethical compliance is a cornerstone of our benchmark design. In curating and releasing financial NLP datasets, we take care to respect privacy, obtain necessary consents, and promote fairness. We align our practices with the ACL ethics guidelines and broader community standards for handling data. Below, we discuss the ethical measures in place regarding data privacy, consent, bias, and overall responsible use of data:

\para{Data Privacy and Consent:} Many financial datasets involve text from reports, news, or social media, which generally pertain to companies or markets rather than private individuals. However, in cases where data might include personal or sensitive information (for example, customer reviews, financial advice communications, or user profiles in fraud detection data), we ensure that privacy is safeguarded. We only include such data if it has been made public with consent or properly anonymized. Specifically, if a dataset contains any personally identifiable information (PII), we verify that the data was collected with informed consent and that the individuals understood their data would be used for research. If this cannot be verified, the dataset is excluded or the PII is removed. Additionally, we avoid datasets that contain sensitive financial records of private individuals unless they are fully anonymized or synthetic. By taking these precautions, we uphold individuals’ privacy rights and comply with regulations and ethical norms around data protection.\\
\para{Bias and Fairness:} We recognize that datasets can inadvertently reflect biases (for example, a credit scoring dataset might over-represent certain demographics, or a financial news dataset might be predominantly from one country’s media). To address this, we encourage dataset contributors to document any known biases or limitations in their data. During the review process, we assess whether the dataset’s content could lead to biased models (such as bias against a group or region) and consider the diversity of the dataset. Our benchmark aims to cover a broad range of financial scenarios (including different markets, languages, and subdomains like banking, investment, insurance) to provide a balanced evaluation. When biases are unavoidable (as they often are in real-world data), we make them transparent: the documentation for each dataset notes aspects like the time period it covers, the geography or entities it focuses on, and any known skew. Users of the benchmark should be aware of these context details when interpreting results. Furthermore, we are committed to updating the benchmark with more diverse datasets over time, to improve fairness and representativeness across the financial NLP tasks.
\paragraph{Transparency and Data Documentation:} In line with principles of research transparency and reproducibility, we provide detailed documentation for every dataset in the benchmark. This includes a description of how the data was collected, what the data consists of (e.g., “10,000 financial news articles from 2010-2020, annotated with sentiment labels by experts”), and any preprocessing steps we performed (such as removing certain fields or normalizing text). We also clearly state the intended use of the dataset and any limitations. Each dataset entry is akin to a datasheet or card that enumerates its characteristics, ensuring that anyone using the dataset understands its context. If a dataset comes with specific usage restrictions or ethical considerations beyond the license (for example, a clause that one should not attempt to re-identify individuals mentioned in the data), we prominently communicate those conditions to the users. By providing this level of transparency, we help researchers use the data responsibly and enable them to explain their results with knowledge of the data’s nuances.
\paragraph{Compliance with Ethical Standards:} Our project abides by the ACL Code of Ethics and broader CS research ethical guidelines. This means that in assembling the benchmark, we have avoided any actions such as using data without permission, violating terms of service of websites, or including content that is derogatory or harmful without due reason. All team members and contributors are expected to follow ethical practices. For instance, if someone were to suggest adding a dataset obtained through web scraping a financial platform, we would require proof that this scraping did not violate the platform’s policies and that no confidential information is included. We also strive for transparency in our own work: any potential ethical issues we encountered during dataset collection or integration are disclosed in our documentation. In cases where we had doubts about a dataset’s ethical viability, we consulted with an ethics advisor or chose to err on the side of caution by not including that data. By enforcing these standards internally and for external contributions, we aim to set a positive example and ensure that the benchmark can be used freely without ethical reservations.
\subsection{Community Expectations}
Any benchmark suite's success relies on having a responsible community of users, contributors, and maintainers. We outline here what we expect from all parties involved to ensure the resource remains trustworthy, well-maintained, and useful for everyone. These expectations cover how data should be treated, how credit should be given, and how collaboration should occur in practice:
\paragraph{Responsible Use by Users:} Researchers and practitioners using the benchmark are expected to use the data and results responsibly. This means they should not misuse the datasets (for example, by trying to extract or infer private information about individuals from a dataset that has been anonymized) and should respect any usage guidelines provided. If a dataset is flagged as for non-commercial use only, users must refrain from deploying it in commercial products. Users should also be careful to preserve the integrity of the data: avoid altering datasets except for necessary preprocessing, and certainly do not modify labels or data points in a way that could mislead results. If a user discovers an issue in a dataset (such as a systematic labeling error or a broken link), we expect them to inform the maintainers via the appropriate channel (GitHub issue, email, etc.) so that it can be addressed for the benefit of all.
\paragraph{Proper Citation and Acknowledgment:} We expect all users of the benchmark to give proper credit in their publications or projects. At minimum, this involves citing this benchmark (the ACL paper or associated technical report) as the source of the evaluation suite, as well as citing the original sources of any datasets used. Proper citation not only acknowledges the work of the benchmark organizers and dataset creators, but also allows others to trace back to the original data for verification or further research. In our benchmark documentation, we provide a BibTeX entry for the benchmark itself and recommend citation strings or references for each dataset. When writing a paper that uses, say, the FiQA sentiment analysis dataset from our suite, the author should cite the FiQA paper in addition to our benchmark paper. This practice is in line with community norms and some dataset licenses that mandate attribution. Users should also acknowledge any tools or baseline results from the benchmark if they directly use them.
\paragraph{Contributor and Maintainer Responsibilities:} Contributors who add datasets or code are expected to maintain a high standard of quality and ethics. They should only contribute data that they have the right to share and that meets the criteria outlined above. Contributors are also encouraged to remain engaged after their dataset is added, in case updates or fixes are needed. On the other side, maintainers (the core team overseeing the benchmark) have the responsibility to manage contributions fairly and efficiently. They should provide constructive feedback to contributors, merge accepted contributions in a timely manner, and update documentation accordingly. Maintainers are also responsible for monitoring the health of the project – if a dataset becomes unavailable or if a license changes, the maintainers must act (e.g., by finding an alternative hosting solution or removing the dataset if it no longer can be shared). Both contributors and maintainers should adhere to a code of conduct that emphasizes respectful communication, openness to feedback, and collaborative problem-solving. Any disputes (for example, if a contribution is deemed unsuitable) should be handled transparently and with courtesy.
\paragraph{Community Collaboration:} We foster an open community environment. Users are encouraged to share their experiences with the benchmark, such as posting results, writing tutorials, or comparing models, in forums or social media, as long as they credit the source. We have set up a discussion board (or use an existing platform like the Hugging Face forums or a Discord channel) for the benchmark where people can ask questions, suggest improvements, or seek help. The expectation is that community members will help each other, making the benchmarking process easier and more standardized. For example, if someone has trouble using a particular dataset, others who have used it can chime in with advice. This kind of peer support is invaluable. We ask that all community interactions remain professional and focused on the science – harassment, discrimination, or any form of unprofessional behavior is not tolerated. By cultivating a friendly and inclusive atmosphere, we hope to attract a wide range of contributors and users, which in turn makes the benchmark more robust and widely applicable.
\paragraph{Extending and Evolving the Benchmark:} The benchmark is not a static resource; we expect it to evolve as the field progresses. Community members who identify gaps in the benchmark (for instance, a new type of financial NLP task that is not covered) are encouraged to propose extensions. This could include new datasets, new evaluation metrics, or even new challenge tasks. When doing so, we expect the same level of rigor as for the initial benchmark: thorough documentation, ethical data handling, and openness to peer review. If researchers create their own extension of the benchmark for private use (say, adding proprietary data for an internal evaluation), we of course cannot enforce the same rules, but we encourage them to share their insights or tools with the community whenever possible. Should any such extensions be made public, we hope the creators will merge efforts with us so that the community has a unified benchmark rather than many fragmented ones. In summary, every user and contributor has a role in upholding the integrity of the benchmark. By using the data conscientiously, citing sources, contributing improvements, and collaborating respectfully, the community ensures that this benchmark remains a valuable asset for financial NLP research now and in the future.
\end{document}